\pdfoutput=1
\documentclass{arxivstyle}

\usepackage{amsmath,amsfonts,bm}

\def\eqref#1{equation~\ref{#1}}

\def\1{\bm{1}}

\DeclareMathAlphabet{\mathsfit}{\encodingdefault}{\sfdefault}{m}{sl}
\SetMathAlphabet{\mathsfit}{bold}{\encodingdefault}{\sfdefault}{bx}{n}

\usepackage{url}
\usepackage{mathrsfs}
\let\paperdefaulttt\ttdefault
\usepackage{inconsolata}
\let\ttdefault\paperdefaulttt
\definecolor{darkgreen}{HTML}{005e19}
\definecolor{darkblue}{HTML}{240394}
\newcommand{\code}[1]{{\fontencoding{T1}\fontfamily{zi4}\selectfont #1}}
\newcommand{\exampleg}[1]{\textcolor{darkgreen}{\textbf{\small{\code{#1}}}}}
\newcommand{\exampleb}[1]{\textcolor{darkblue}{\textbf{\small{\code{#1}}}}}

\newtcolorbox{promptbox}{enhanced, colback=accentLight, colframe=accentLight,
  boxrule=0pt, arc=6pt, left=8pt, right=8pt, top=6pt, bottom=6pt,
  width=0.485\linewidth, box align=top, nobeforeafter,
  equal height group=prompts, before upper=\raggedright}

\newcommand{\papertitle}{Probe with Participation Trophies: Random-Reward RL as a Probe of LLM Capability}
\newcommand{\paperdate}{September 2026}
\newcommand{\codeurl}{https://github.com/mynotwo/rl-random-prob}

\begin{document}


\begin{herobox}
{\setlength{\parskip}{0pt}%

{\raggedright\titlefont\papertitle\par}

\vskip 0.2cm

{\raggedright\authorfont
Yu Mao,
Lei Yu\textsuperscript{1},
Zining Zhu\textsuperscript{2},
Yusheng Zheng\textsuperscript{3},
Haohang Li\textsuperscript{2},
Freda Shi\textsuperscript{4,5},
Yutong Yin\textsuperscript{6},
Zhaoran Wang\textsuperscript{6},
Jingcheng Niu\textsuperscript{1}\par}

\vskip 0.2cm

{\raggedright\normalsize
\textsuperscript{1}University of Toronto,
\textsuperscript{2}Stevens Institute of Technology,
\textsuperscript{3}University of California, Santa Cruz,
\textsuperscript{4}University of Waterloo,
\textsuperscript{5}Vector Institute,
\textsuperscript{6}Northwestern University\par}

\vskip 0.5cm

We connect the spurious-reward paradox to a model's \textbf{reachability} and propose random-reward reinforcement learning (RL) as a useful tool for the probing enterprise, addressing a decade-long debate over what probing performance actually reveals about a model.
There are two prevailing explanations for the surprising finding that even random rewards can improve the performance of large language models (LLMs): one attributes the gains to particular mechanisms within RL training \citep{shaoSpuriousRewardsRethinking2026, chenExplorationVsExploitation2025}; the other to data contamination \citep{yanSpuriousRewardsParadox2026}. Our results motivate a different view: spurious-reward RL can probe a model's reachability, or what further training can attain from its current state under specified constraints, beyond what is reflected in its current performance. Two OLMo checkpoints with the same accuracy on synthetic arithmetic (3.5\%), for example, reach 8.5\% and 55\% in their best runs under the same correctness-rewarded RL. Examining OLMo checkpoints across pre-training and mid-training reveals three distinct regimes of training response: early on, RL produces little improvement even when correct answers are rewarded; later in pre-training, rewarding correct answers becomes effective while random rewards remain weak; and, upon entering mid-training, even random rewards can produce large gains. A similar ordering appears in a number-masked supervised fine-tuning (SFT) analysis of these checkpoints, suggesting that the pattern is not specific to a particular RL mechanism. Moreover, RL with random rewards offers a distinctive perspective on what training can make an LLM do, since its reward signal supplies no information about which answers are correct. By asking what training can attain without correctness feedback, it addresses the label-leakage side of a central problem in decodability-based probing: whether a successful probe reveals the model's capabilities or learns the task itself.\par

\vskip 0.5cm

{\metadatafont
\begin{tabular}{@{}ll}
\textbf{Date:} & \paperdate \\
\textbf{Code:} & \href{\codeurl}{\codeurl}\\
\end{tabular}
}

}
\end{herobox}

\vspace{2.0em}


\section{Introduction}
\label{sec:intro}

Reinforcement learning with verifiable rewards (RLVR) has played a central role in improving the capabilities of large language models (LLMs) and ushering in an ``age of LLM agents.'' A natural intuition is that LLMs can learn from reward signals by favoring better responses over worse ones. Yet \citet{shaoSpuriousRewardsRethinking2026} recently reported that even spurious (random) rewards can improve model performance. Why can a model improve when its rewards are assigned at random, providing no more correctness information than a random coin toss? This counterintuitive finding prompted a number of investigations and led \citet{yanSpuriousRewardsParadox2026} to coin the term \emph{spurious-reward paradox}.
Explaining it requires understanding what shapes the model’s response to training beyond correctness feedback.

There are currently two main lines of explanation for the paradox, but neither provides a complete account in light of our new empirical results. The first attributes the gains to data contamination \citep{yanSpuriousRewardsParadox2026}. The gains we observe on synthetically generated problems suggest, however, that the phenomenon cannot be attributed entirely to data leakage. The second looks to particular mechanisms within RL training. \citet{shaoSpuriousRewardsRethinking2026} conjecture that clipping in GRPO \citep{shaoDeepSeekMathPushingLimits2024} favors behavior to which the model already assigns high probability. \citet{chenExplorationVsExploitation2025}, however, argue that clipping bias does not explain the gains. They explain the paradox through the model's current performance, as reflected in its rollouts: more accurate models generate fewer incorrect responses that can be rewarded by chance. Nevertheless, when two checkpoints are trained offline on the same rollouts with the same random rewards, the later, higher-accuracy checkpoint shows larger mean gains under both rollout sources. This suggests that understanding the gains requires looking beyond the model's current performance to properties of the learner that shape its response to the same training signal. In particular, capabilities the model already has but does not fully put to use, or its \emph{dormant capability}, may play an important role.

\begin{figure}
    \centering
    \includegraphics[width=\linewidth]{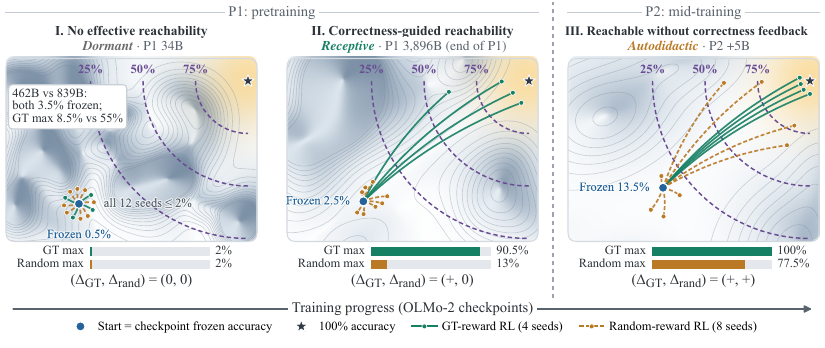}
    \caption{Development of LLM reachability across pre-training and mid-training (OLMo-2, synthetic GSM). Early checkpoints barely improve under either reward (dormant), later ones improve under GT rewards (receptive), and in mid-training some random-reward runs also reach high accuracy (autodidactic). Each line is a seed ending at its held-out accuracy after 500 RL steps, higher closer to the star; bars give the best across seeds, and $(\Delta_{\rm GT}, \Delta_{\rm rand})$ marks a gain (+) or none (0). The landscape is schematic; overlapping endpoints (panel I; GT in panel III) are spread slightly.}
\end{figure}

We instead consider the model's dormant capability together with the information provided in the question prompt. A model may not yet reliably generate the correct answer, yet the prompt can still elicit behavior that training can act on. Under random rewards, some useful responses may be reinforced by chance; which behaviors can be elicited and reinforced should therefore depend in part on the starting model and the information in the input. On this account, training can improve performance even when the reward does not tell the model which answers are correct.

We can therefore use a model's response to random-reward RL training as a probe. We introduce \emph{reachability} to describe how far, and how readily, a model can improve from its current state under specified constraints on the training procedure, the information supplied, and the training budget. Specifically, we use \emph{random-reward reachability} to probe dormant capability. We prove that, under random-reward RL training, the reward channel carries no correctness information, allowing us to study what training can attain without correctness feedback. This isolates the label-leakage side of a longstanding question in probing: whether a successful probe reveals the model's dormant capability or learns the task from its own supervision \citep{hewittDesigningInterpretingProbes2019}.

Applying random-reward probing in the wild across OLMo checkpoints reveals three stages of LLM capability development. Early in pre-training, \exampleb{OLMo-2-1124-7B} is \textbf{dormant}: even RL with ground-truth (GT) rewards yields little improvement. As pre-training progresses, the model becomes \textbf{receptive}: GT rewards become effective, suggesting that the model has acquired the basic task knowledge needed to benefit from further training. Random rewards remain weak, however, suggesting that the model has acquired relevant knowledge and skills but has yet to integrate them into capabilities it can readily draw on. When the model enters mid-training, it becomes \textbf{autodidactic}: training can now help it put this knowledge to use even without a reward signal that tells it which answers are correct. These results suggest that a model's receptiveness to post-training depends on its overall capability at that point in training, and that reachability provides a viable way to study it.

\paragraph{Contributions} We make three contributions: (1) we provide an alternative account of the spurious-reward paradox that incorporates the starting model and the information in the input (\S\ref{sec:mechanism}); (2) we formalize \emph{reachability} as a framework for probing what further training can attain and develop random-reward RL as a correctness-free probe (\S\ref{sub:reachability-probe}); and (3) we identify a three-stage pattern in training response that reveals changes not fully reflected in current task performance (\S\ref{sec:readings}).

\section{Background, Related Work, and Experimental Setup}
\label{sec:background}

Before discussing our experiments and results, we review RL, the spurious-reward paradox, and its two prevailing accounts and their limitations. We then revisit the probing debate: whether a probe reveals the model's knowledge or learns the task itself.\footnote{For the sake of space, we defer a more comprehensive literature review to Appendix~\ref{app:extensive_literature_review}.}

\paragraph{Reinforcement Learning with Spurious Rewards}

In a typical RLVR setting\footnote{Appendix~\ref{app:prelim} reviews core RL concepts and notation, following \citet{shaoSpuriousRewardsRethinking2026} and \citet{chenExplorationVsExploitation2025}.}, the model generates a response, and a verifier assigns a reward by checking the answer or executing tests. Methods such as GRPO \citep{shaoDeepSeekMathPushingLimits2024} can train models without worked solutions by sampling multiple responses to the same prompt and updating the model based on their relative rewards. Surprisingly, \citet{shaoSpuriousRewardsRethinking2026} found that even \textit{spurious rewards}---rewards with little, no, or even negative correlation with answer correctness---can elicit strong reasoning performance under certain conditions. Subsequent work has proposed several explanations for this phenomenon, including reduced reward-misalignment damage in more accurate models \citep{chenExplorationVsExploitation2025} and increased model confidence through entropy minimization \citep{agarwalUnreasonableEffectivenessEntropy2025,gaoOneshotEntropyMinimization2026}. In this work, we study spurious rewards from a different perspective. We show that random rewards---a special case of spurious rewards with zero correlation with correctness---can serve as an effective probe of dormant LLM capabilities in generation tasks such as reasoning and coding.

\paragraph{The Probing Debate: Measuring an LLM's Dormant Capability}
Probing has long been a central tool for studying what models know and what they are capable of. The usual recipe is straightforward: freeze the model, extract its representations, and train a classifier, or \emph{probe}, to predict a property of interest. Good probing performance is then taken as evidence that the model has learned something about that property. But whose knowledge have we found? The probe has, after all, been given the answers during training. As \citet{hewittDesigningInterpretingProbes2019} ask, ``can we conclude that the representation encodes linguistic structure, or has the probe just learned the task?''

This question has motivated different accounts of \emph{decodability}: what can be recovered from a representation, and under what conditions. \citet{pimentelInformationTheoreticProbingLinguistic2020} formulate probing as estimating mutual information between representations and task labels. But information can be preserved without being easy to extract. The Bayesian account of \citet{pimentelBayesianFrameworkInformationTheoretic2021} considers what a learner can extract given its prior knowledge and finite training data. Predictive $\mathcal V$-information makes the predictor family explicit \citep{xuTheoryUsableInformation2019}, while conditional probing asks what the representation contributes beyond a baseline \citep{hewittConditionalProbingMeasuring2021}. The question thus becomes not merely whether information is present, but under what conditions it can be used.

Nevertheless, probes for LLM capabilities still largely rely on supervised learning from task labels and therefore do not remove the concern raised by \citet{hewittDesigningInterpretingProbes2019}: the probe may learn the task from its own supervision. We show that random-reward RL can probe what training can attain while excluding correctness information from the reward signal, thereby isolating this label-leakage component of the broader probing problem.

While this debate may seem to have gone quiet, its central questions have only become more relevant and important, resurfacing under different names. Recent work such as sparse autoencoders \citep[SAE;][\textit{inter alia}]{brickenMonosemanticityDecomposingLanguage2023}, the linear representation hypothesis \citep[][\textit{inter alia}]{parkLinearRepresentationHypothesis2023,lampinenLinearRepresentationsLanguage2026}, and the J-space \citep{gurneeVerbalizableRepresentationsForm2026} again raises the question of what the structures recovered from a model's representations tell us about what the model knows and can do.

\section{Explaining the Spurious-Reward Paradox: Learner and Input}
\label{sec:mechanism}

We now examine why LLMs can improve under random-reward RL. Previous accounts \citep{shaoSpuriousRewardsRethinking2026, chenExplorationVsExploitation2025} do not account for two observations we present below: different learner checkpoints respond differently to identical rollouts and rewards, and the same checkpoint responds differently across the two random-token input conditions. These results point to the starting model and the information in the input as additional determinants of training response.

Our main experiments follow the publicly released checkpoints of \exampleb{OLMo-2-1124-7B} \citep{olmo2OLMo22025} through pre-training (P1) and mid-training (P2). P1 uses a broad corpus dominated by web data, while P2 introduces a curated mixture of high-quality and task-focused data. The checkpoints span both stages, with architecture and parameter count held fixed. We also examine training checkpoints of \exampleb{SmolLM2-1.7B} and \exampleb{OLMo-3-7B}.\footnote{See the checkpoint coverage in Appendix~\ref{app:tasks} and additional checkpoint comparisons in Appendix~\ref{app:crossmodel-checkpoints}.}

\begin{table}[t]
    \centering\small
    \begin{tabular}{ccccc} \toprule Response & \multirow{2}{*}{Learner} & Initial & Terminal & Mean\\ source & & score & max & gain\\ \midrule \multirow{2}{*}{P1} & P1 & 2.5 & 9.5 & +0.6\\  & P2 & 13.5 & 80.0 & +4.5\\ \midrule \multirow{2}{*}{P2} & P1 & 2.5 & 14.0 & +2.9\\  & P2 & 13.5 & 57.0 & +13.7\\ \bottomrule \end{tabular}
    \caption{Identical rollouts and random rewards yield different outcomes across OLMo-2 checkpoints. P1: end of pre-training; P2: +5B mid-training. Each row summarizes 16 runs, with terminal results at step 500. Scores are accuracies (\%); gains are changes from initial accuracy (percentage points).}
    \label{tab:shared-rollout-main}
\end{table}

\begin{figure}[t]
    \centering
    \includegraphics[width=0.72\linewidth]{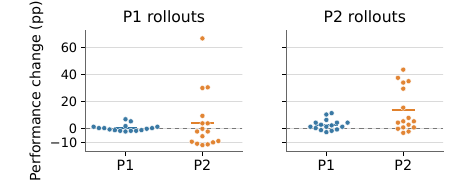}
    \caption{With the same rollouts and rewards, P2 gains more on average and varies more. Panels split runs by rollout source, the x-axis by learner. Dots are the 16 runs per group and lines their means; changes are step-500 minus initial accuracy (percentage points).}
    \label{fig:shared-rollout-change}
\end{figure}

\paragraph{Different Models Yield Different Learning Outcomes from the Same Rollouts}

First, we present an offline random-reward RL experiment in which two learner checkpoints are trained on exactly the same rollouts and rewards. We take two \exampleb{OLMo-2-1124-7B} checkpoints, one from the end of pre-training (P1: 3,896B tokens) and the first checkpoint in mid-training (P2: +5B tokens). We generate rollouts from each frozen checkpoint and train both learners on each set, giving four source--learner combinations. Within each source, both learners receive exactly the same cached rollouts and random rewards and start from identical LoRA weights. We run 16 paired replicates for 500 updates each. The rollouts remain fixed throughout training, so differences between the learners cannot be attributed to differences in the responses or rewards they receive.

The responses generated by P1 can take P2 from 13.5\% to 80.0\%, while P1 itself reaches at most 9.5\% at the endpoint (Table~\ref{tab:shared-rollout-main}). With P2 rollouts, the highest terminal scores are 57.0\% for P2 and 14.0\% for P1. Thus, under either shared rollout source, P2 shows the larger upward changes. Its mean gains are also larger: 4.5 versus 0.6 percentage points with P1 rollouts, and 13.7 versus 2.9 percentage points with P2 rollouts.

Recall that \citet{chenExplorationVsExploitation2025} explain the spurious-reward paradox through the model's current performance and rollout quality, arguing that training stronger models ``incurs less total misalignment damage ... and is therefore more likely to yield improvements under random rewards.'' That mechanism cannot by itself explain the learner difference here: within each source condition, P1 and P2 receive the same responses and the same random rewards, so the number and identity of accidentally rewarded responses are fixed. Their account may still explain variation across rollout sources, but the contrast between learners requires an additional property of the learner itself.

\paragraph{Random Inputs without Task Cues Yield Little Improvement}

\begin{figure}[t]
    \centering
    \begin{promptbox}
    \textbf{With math instructions}\par\smallskip
    \exampleg{\footnotesize Solve the math problem. Show your reasoning and put the final answer on a line like: \#\#\#\# <number>.}\par\smallskip
    \exampleg{\footnotesize Question: <random tokens>}\par
    \exampleg{\footnotesize Answer:}
    \end{promptbox}\hfill
    \begin{promptbox}
    \textbf{Random tokens only}\par\smallskip
    \exampleg{\footnotesize <random tokens>}
    \end{promptbox}
    \caption{Inputs used to separate task cues from correctness information.}
    \label{fig:randq-prompts}
\end{figure}

\begin{figure}
\centering
\includegraphics[width=\linewidth]{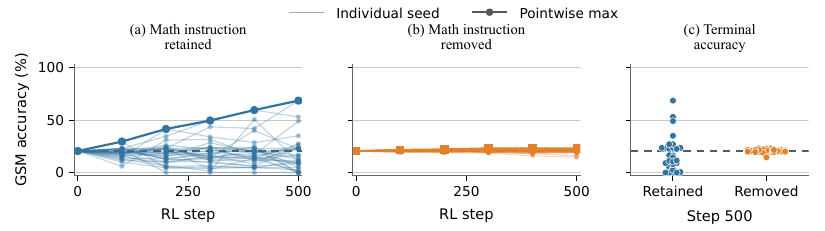}
\caption{Random-reward training of OLMo-2 P2 +50B on sequences of randomly sampled tokens: (a) accompanied by the math instruction and (b) presented alone. Panel (c) shows terminal accuracy after 500 training steps. Thin lines and dots show individual runs (32 seeds per condition); highlighted lines show the highest accuracy across seeds at each step. The dashed line marks the initial accuracy of 20.5\%. Further comparisons appear in Appendix~\ref{app:randq-results}.
}
\label{fig:randq-overview}
\end{figure}

Next, we train the same model checkpoint on two types of input constructed from random tokens. In one condition, we replace the original questions while retaining the instruction and template markers. In the other, the entire prompt consists of random tokens (Figure~\ref{fig:randq-prompts}).

The P2 +50B \exampleb{OLMo-2-1124-7B} checkpoint responds differently to these two input conditions under random-reward RL. Because the question content in both consists of random tokens, neither presents a valid mathematical problem to solve; a current task-accuracy account on these inputs therefore does not by itself predict the difference between them. As shown in Figure~\ref{fig:randq-overview}, the checkpoint has the same initial held-out accuracy of 20.5\% in both conditions and reaches up to 68.5\% terminal accuracy when the math instruction is retained. Without the task instruction, however, all recorded evaluations remain between 14.5\% and 23\%. The contrast shows that training response depends on information in the input even when the reward supplies no correctness information.

Additional input controls in Appendix~\ref{app:randq-results} support the same conclusion. We also observe the same pattern in \exampleb{Qwen2.5-7B} and \exampleb{Llama-3.1-8B} (Appendix~\ref{app:crossmodel-inputs}).

\paragraph{Reachability and an Account of the Spurious-Reward Paradox}

Taken together, these findings motivate an account of the spurious-reward paradox in terms of the starting model, the information in the input, and the model's \emph{reachability}, which we formalize in the next section.

In short, the reward function determines how sampled answers are reinforced or suppressed, but the effect of those updates also depends on the starting model and the input. Even when rewards are assigned by random coin tosses and contain no correctness information, the prompt can still elicit useful behavior, and some random updates can reinforce it. If the starting model can already produce task-relevant behavior from the information in the input, some random-reward trajectories may therefore make that behavior more usable and improve task performance.

The two experiments examine two factors not captured by rollout quality alone. First, P2 changes more than P1 when trained on identical rollouts and random rewards, showing that the learner state matters. In the second, the two random-token input conditions produce sharply different outcomes. The retained instruction and template markers still specify the task domain, request a reasoning chain, define the final-answer format, and structure the prompt. They do not supply the missing mathematical problem or its answer, but they provide task information that the model can draw on.

This gives a different view of dormant capability: random-reward training lets us ask how much task performance further training can attain from a model without correctness feedback through the reward. We call this property \emph{reachability}.

\section{Reachability Probing with Random-Reward RL}
\label{sub:reachability-probe}

This new way of studying a model's capability offers a general-purpose approach to probing that provides no correctness feedback during adaptation. It thus directly addresses the longstanding question of whether a probe reveals the model's capability or learns the task from its own supervision. This probe also yields new findings about the development of models' dormant capability, which are particularly relevant today, when post-training plays an ever more important role in developing LLM capabilities. In this section, we provide a formal definition of our reachability probe.

We first discuss what information means when studying a model's dormant capability, and then formalize reachability. We next present a short proof that random reward signals do not introduce correctness information through the reward channel. Finally, we introduce random-reward RL as a reachability probe, which we use in the next section on real LLMs.

\paragraph{What Do We Mean by Information and Dormant Capability?}
To study a model's dormant capability, we need to distinguish the information available to it from its ability to use that information. For a mathematical problem $X$ with a uniquely determined answer $Y=f(X)$, the answer's information is already in the question in the Shannon sense:
\begin{equation}
    H(Y\mid X)=0.
\end{equation}
Once we condition on $X$, asking whether training adds new Shannon information about $Y$ may therefore miss the point. The question is whether the model has the computational ability to extract the answer. This is what we mean by \emph{computational accessibility}. Learned lossless compression shows the same distinction: the information in the data is fixed, yet the number of bits needed depends on how well a predictor can model it \citep{maoTRACEFastTransformerbased2022,maoFasterStrongerLossless2023}, and text generated by one LLM is highly predictable under another \citep{mao2025losslesscompressionlargelanguage}.

Predictive $\mathcal V$-information gives us a language for this distinction. It measures how much information a restricted family of predictors $\mathcal V$ can use, and can increase through computation \citep{xuTheoryUsableInformation2019,hewittConditionalProbingMeasuring2021}. A model state $\theta$ specifies computations that produce a representation $Z_\theta=\phi_\theta(X)$. For a fixed family $\mathcal V$, $I_{\mathcal V}(Z_\theta\to Y)$ measures the task information usable by that family. It is therefore possible for training to produce
\begin{equation}
    I(Y;Z_{\theta_1})\approx I(Y;Z_{\theta_0}),
    \qquad
    I_{\mathcal V}(Z_{\theta_1}\to Y)>I_{\mathcal V}(Z_{\theta_0}\to Y).
\end{equation}
The same information has therefore become easier to use. Thus, we can probe the model's dormant capability through the task performance it can attain with post-training.

\paragraph{Reachability}
We introduce \emph{reachability} to describe how far, and how readily, a model can improve from its current state under specified constraints on the training procedure, the information supplied, and the training budget. These constraints are part of the measurement: the same checkpoint may respond differently when the available supervision or the allowed updates change.

Fix an initial checkpoint $\theta_0$, a task performance measure $J$, and a training budget $B$. Let $\mathfrak A$ be a prespecified family of adaptation procedures, with constraints on inputs, supervision, parameter updates, selection, and stopping. Writing $\mathcal A_B(\theta_0;U)$ for the checkpoint obtained by $\mathcal A$ within budget $B$ under training randomness $U$, we define:
\begin{equation}
    \operatorname{Reach}_B(\theta_0)
    :=
    \sup_{\mathcal A\in\mathfrak A}
    \mathbb E_U\!\left[J\!\left(\mathcal A_B(\theta_0;U)\right)\right].
    \label{eq:reachability}
\end{equation}
In words, this is the highest expected task performance attainable from the checkpoint within the given budget using the allowed procedures. Its improvement over current performance is
\begin{equation}
    \Delta_{\mathrm{reach}}(\theta_0)
    :=
    \operatorname{Reach}_B(\theta_0)-J(\theta_0).
\end{equation}
Both quantities depend on the task, intervention family, and budget; we leave the first two implicit in the notation. The dependence on $B$ describes the training needed for a given expected performance. Comparing checkpoints under the same constraints can reveal differences in training potential, even when their current scores are similar.

\paragraph{No Correctness Information through Random Rewards}
The information supplied during training matters when interpreting reachability as a probe of dormant capability. If the training signal identifies correct responses, it can teach the model how to solve the task. Random rewards let us remove this source of supervision.

Let $X_t$ be a training input, $A_t$ a sampled response, and $C_t=v(X_t,A_t)$ its correctness under a verifier $v$. Let $H_t$ denote the preceding training history.

\begin{theorem}[No correctness information through the reward channel]
Suppose each reward $R_t\sim\operatorname{Bernoulli}(p)$ is sampled independently of the input, response, and preceding training history. Then
\begin{equation}
    I(R_t;C_t\mid H_t,X_t)=0.
    \label{eq:no-correctness}
\end{equation}
If the initial checkpoint, adaptation protocol, and joint law of its random draws are held fixed, and all training decisions are independent of the verifier, replacing $v$ with any other verifier leaves the distribution of the training trajectory unchanged.
\end{theorem}

The proof couples the runs under different verifiers using the same random draws. Their inputs, responses, rewards, and updates agree at every step because none uses the verifier; the full argument appears in Appendix~\ref{app:reachability-theory}. Any performance gain under this protocol therefore occurs without the reward oracle telling the model which responses are correct. Correct answers are used for evaluation only. The model can still learn to use the task information in its inputs more effectively.

\paragraph{Random-Reward RL as a Reachability Probe}
For probing without correctness feedback, we restrict the allowed procedures to a family $\mathfrak A_0$ whose training signals, selection decisions, and stopping rules satisfy the verifier independence above. In what follows, $\operatorname{Reach}_B$ and $\Delta_{\mathrm{reach}}$ refer to this family. Random-reward RL gives us a concrete procedure $\mathcal A_{\mathrm{rand}}\in\mathfrak A_0$ with which to probe reachability. We call reachability under this fixed procedure \emph{random-reward reachability}, and report its expected improvement over the starting checkpoint:
\begin{equation}
    \Delta_{\mathrm{rand}}(\theta_0)
    :=
    \mathbb E_U\!\left[J\!\left(\mathcal A_{\mathrm{rand},B}(\theta_0;U)\right)\right]-J(\theta_0)
    \leq
    \Delta_{\mathrm{reach}}(\theta_0).
    \label{eq:random-reward-gain}
\end{equation}
The inequality follows because random-reward RL is one of the allowed procedures: its expected gain gives a lower bound on the improvement attainable within the family. Repeated runs estimate this gain through their mean terminal performance change. This gives us a concrete measure of how much the starting model can benefit from the specified training without requiring us to optimize over every allowed procedure. We use this \emph{correctness-free probe} in the next section to examine how training response changes across checkpoints.

\section{LLM Capability Emergence through the Lens of Reachability}
\label{sec:readings}

\begin{figure}[t!]
\centering
\includegraphics[width=\linewidth]{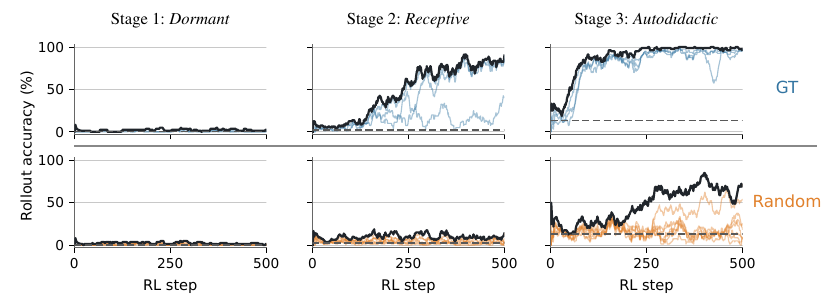}
\par\noindent\makebox[\linewidth][l]{\hspace*{0.06\linewidth}\begin{minipage}{0.87\linewidth}
\captionsetup[subfigure]{font=scriptsize,justification=centering,singlelinecheck=false}
\begin{subfigure}[t]{0.333\linewidth}
\caption{\mbox{P1: 34B; Frozen GSM: 0.5\%}}
\end{subfigure}\hfill
\begin{subfigure}[t]{0.333\linewidth}
\caption{\mbox{P1: 3,896B; Frozen GSM: 2.5\%}}
\end{subfigure}\hfill
\begin{subfigure}[t]{0.333\linewidth}
\caption{\mbox{P2: +5B; Frozen GSM: 13.5\%}}
\end{subfigure}
\end{minipage}}
\caption{Three representative OLMo-2 checkpoints illustrate the dormant, receptive, and autodidactic stages of training response on synthetic GSM. Top: GT, four seeds per checkpoint; bottom: Random, eight. Thin lines show sampled rollout accuracy (trailing 20-step mean); black lines show pointwise maxima across seeds; dashed lines show frozen greedy held-out accuracy.}
\label{fig:trajectories}
\end{figure}
\begin{figure}[t!]
\centering
\begin{subfigure}{\linewidth}
\centering
\includegraphics[width=\linewidth]{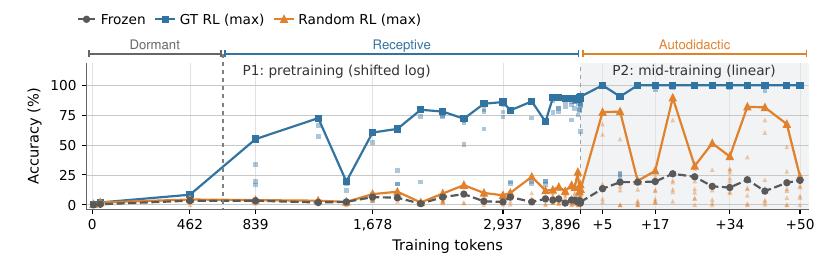}
\caption{Response to RL on synthetic GSM}
\label{fig:six-a}
\end{subfigure}
\begin{subfigure}{\linewidth}
\centering
\includegraphics[width=\linewidth]{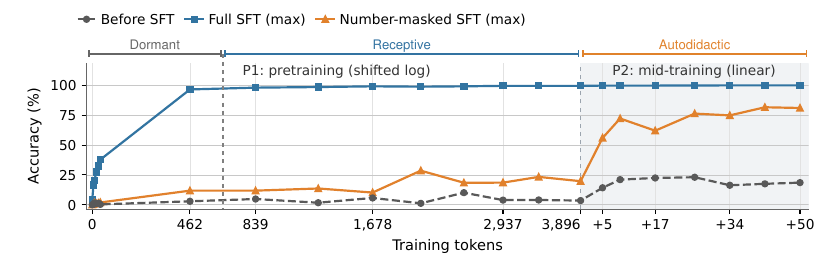}
\caption{Response to full and number-masked SFT on synthetic GSM}
\label{fig:six-b}
\end{subfigure}
\caption{Both RL and SFT show the same three-stage developmental pattern across OLMo-2 checkpoints (synthetic GSM). (a) Frozen and maximum terminal accuracy after 500 RL steps; faint points show individual seeds. (b) Initial and best reported SFT accuracy. Stage bands in both panels follow the RL results in (a). See Appendix~\ref{app:sft} for the SFT comparison.}
\label{fig:checkpoint-scan}
\end{figure}

We now deploy our reachability probe in the wild. For brevity and consistency, we focus on \exampleb{OLMo-2-1124-7B} \citep{olmo2OLMo22025} checkpoints. Experiments with \exampleb{SmolLM2-1.7B} and \exampleb{OLMo-3-7B} show similar patterns supporting the same conclusion (Appendix~\ref{app:crossmodel-checkpoints}).

We study how responsiveness to different training signals changes over the course of model training. For OLMo-2, this means examining checkpoints from pre-training and mid-training. We do not include post-training checkpoints, because prior post-training signals would make the results harder to interpret. In addition to measuring frozen performance, we train each checkpoint with two reward signals: ground truth (GT) and random rewards. We focus on synthetic GSM-style arithmetic problems \citep{cobbeTrainingVerifiersSolve2021}, for which training response continues to change throughout OLMo-2 pre-training and mid-training. We also experiment with MBPP Python programming problems \citep{odenaProgramSynthesisLarge2021} and synthetic one- and two-hop questions over invented facts \citep[cf.][]{westonAICompleteQuestionAnswering2015}. Appendix~\ref{app:tasks} gives further details.

\paragraph{Dormant, Receptive, and Autodidactic --- Three Stages of Training Response Revealed by Reachability Probing}

Figure~\ref{fig:trajectories} illustrates three stages of training response under our reachability probe. Early in pre-training (P1: 34B), the checkpoint is \emph{dormant}: it shows little performance improvement under either reward signal on synthetic GSM (mean gains of 0.4 and 0.5 percentage points under GT and Random rewards). Later in pre-training (P1: 3,896B), it is \emph{receptive}: GT rewards produce large gains (a mean gain of 79.5 points from a 2.5\% start, with runs ending at 61.5--90.5\%), while random-reward RL remains weak (a mean gain of 4.3 points). After the model enters mid-training (P2: +5B), it is \emph{autodidactic}\footnote{Autodidactic means ``Self-taught.'' Here, we mean improvement without correctness feedback from rewards.}: random rewards now yield a mean gain of 26.8 points from a 13.5\% start, driven by four of eight runs that reach 59.0--77.5\%. This response is not stable across mid-training: over the twelve P2 checkpoints, the mean Random gain ranges from $-$11.8 to +26.8 points and exceeds 10 points at three of them (+5B, +21B, and +42B; Appendix~\ref{app:scan-results}).

Figure~\ref{fig:six-a} presents the full checkpoint scan, where we observe the same three-stage pattern. Early checkpoints show little improvement under either reward signal; correct rewards become effective as pre-training progresses, followed by the abrupt development of large gains under random rewards in mid-training.
A similar ordering appears under supervised fine-tuning (SFT): full SFT becomes effective earlier, whereas number-masked SFT shows its strongest response later (Figure~\ref{fig:six-b}).

These three stages give an empirical interpretation of our reachability framework (\S\ref{sub:reachability-probe}). We use the expected gain $\Delta_{\mathrm{rand}}$ (Eq.~\ref{eq:random-reward-gain}) and define $\Delta_{\mathrm{GT}}$ analogously for GT reward training. For a fixed task, training budget, and pair of protocols, the observed ordering can be summarized schematically as:
\begin{equation}
    \begin{aligned}
        \textit{Dormant:}      &\quad \Delta_{\mathrm{GT}}\approx0, &\quad \Delta_{\mathrm{rand}}\approx0, \\
        \textit{Receptive:}    &\quad \Delta_{\mathrm{GT}}>0,       &\quad \Delta_{\mathrm{rand}}\approx0, \\
        \textit{Autodidactic:} &\quad \Delta_{\mathrm{GT}}>0,       &\quad \Delta_{\mathrm{rand}}>0.
    \end{aligned}
    \label{eq:three-stage-gains}
\end{equation}
Here, $\approx0$ and $>0$ indicate little improvement and meaningful gains, respectively.

We can derive the three stages from the ordering  $\Delta_{\mathrm{GT}}\geq\Delta_{\mathrm{rand}}$ proved in Section~\ref{sub:reachability-probe}. Interestingly, OLMo-2 exhibits these regimes in sequence as training progresses, with relatively distinct transitions; a similar developmental pattern also appears under SFT. Do these transitions constitute ``emergence''? Both the notion of ``emergence'' itself and whether transitions such as these constitute emergence remain debated \citep[\textit{inter alia}]{weiEmergentAbilitiesLarge2022, schaefferAreEmergentAbilities2023,niuIllusionAlgorithmInvestigating2025}. The agreement between theory and observation nevertheless suggests that capability development follows regularities we have yet to understand. Identifying them could help explain why capabilities develop when they do and what training conditions make that possible.

\begin{figure}
\centering
\includegraphics[width=\linewidth]{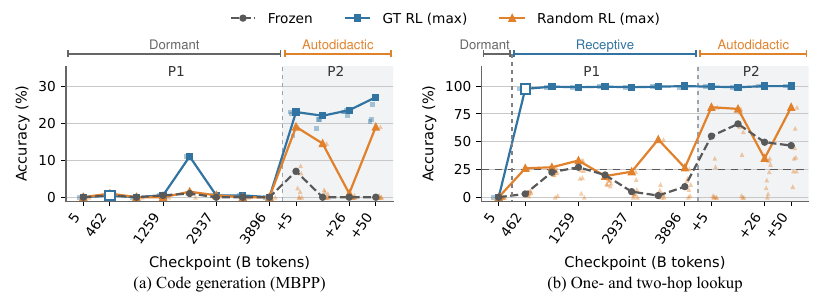}
\caption{OLMo-2 results on code (a) and lookup (b). The overall trend is similar, although the timing of the transitions is idiosyncratic to each task. Lines show frozen accuracy and maximum accuracy across seeds after 500 RL steps; faint points show individual runs. On code, GT and Random gains first appear together in P2. The dashed line marks chance accuracy on lookup (25\%).}
\label{fig:code-lookup-overview}
\end{figure}

\paragraph{Similar Patterns across Tasks and Models}
We observe similar developmental patterns beyond OLMo-2 and arithmetic. On code generation and compositional lookup, random rewards again produce larger gains at later checkpoints (Figure~\ref{fig:code-lookup-overview}). The timing is idiosyncratic to each task, however: at 462B tokens, GT training already brings OLMo-2 to 97.5\% on lookup but only 8.5\% on GSM. Differences in timing also appear across models. In \exampleb{OLMo-3-7B}, correct rewards become effective before random rewards, with large random-reward gains appearing later in mid-training than in OLMo-2. Within the checkpoints tested, \exampleb{SmolLM2-1.7B} reaches only the receptive stage: correct rewards yield substantial gains while random rewards remain weak. Additional results appear in Appendices~\ref{app:beyond-arithmetic} and~\ref{app:crossmodel-checkpoints}.

\section{Discussion and Conclusion}
\label{sec:discussion}
We offer a new explanation for the counterintuitive spurious-reward paradox and find that random-reward RL can be used as a way to probe the development process of an LLM under different training regimes. We believe we have taken a step forward toward a better understanding of LLMs and how their capabilities develop; nevertheless, more questions arise as we edge toward the truth.

Our theoretical account and empirical results suggest that there is usable information in input questions that LLMs can use even without correctness feedback from rewards. Nevertheless, studying the exact mechanism by which this information is utilized is a natural extension of this work. It could be a sign that, after a period of training, LLMs begin to bring their knowledge and skills together and show early signs of self-improvement: telling the model what task to attempt may be enough for it to improve through further training. The mechanism could also be superficial. The model may simply learn a better decoding strategy that generates correct responses more reliably. While shallow, such a strategy is certainly a capability; it does not, however, by itself establish the consistency across equivalent expressions that we would expect of ``knowledge'' \citep{niuWhatDoesKnowledge2024}. Future work can build on these findings by studying how random-reward training affects pass@$k$ across sampling budgets \citep{yueDoesReinforcementLearning2025} and the consistency of responses across equivalent questions.

Moreover, to keep this paper within a reasonable scope, we focus on RL as the vehicle for reachability-based probing. Yet other modes of post-training can also be used to develop a whole suite of reachability probes that help us better understand LLMs.

\clearpage

\bibliographystyle{acl_natbib}
\bibliography{references}
\clearpage
\beginappendix
\clearpage
\section{Preliminaries: RLVR, GRPO, and Random Rewards}
\label{app:prelim}
\label{app:preliminaries}

RL training generates responses, assigns rewards, and updates the model to favor responses receiving higher rewards. Random-reward training follows the same procedure, except that the reward no longer indicates whether a response is correct. This section introduces the notation used to describe that procedure, following \citet{shaoDeepSeekMathPushingLimits2024,shaoSpuriousRewardsRethinking2026,chenExplorationVsExploitation2025}.

\subsection{Language Models as Policies}

For a language model, a \emph{policy} is its distribution over the next token. Let $x\sim\rho$ be a training prompt, $\theta$ the model parameters, and $y=(y_1,\ldots,y_{|y|})$ a generated response, also called a \emph{rollout}. With history $h_t=(x,y_{<t})$, the probability of the response is
\[
\pi_\theta(y\mid x)=\prod_{t=1}^{|y|}\pi_\theta(y_t\mid h_t).
\]
Here $t$ indexes tokens within a response. The response length is determined by the stopping rule and generation cap.

Given a reward $r(x,y)$, RL seeks to increase its expectation,
\[
J_{\rm RL}(\theta)=\mathbb E_{x\sim\rho,\,y\sim\pi_\theta(\cdot\mid x)}[r(x,y)].
\]
When the reward has no explicit dependence on $\theta$, the policy-gradient identity gives
\[
\nabla_\theta J_{\rm RL}(\theta)
=\mathbb E_{x,y}\!\left[r(x,y)\nabla_\theta\log\pi_\theta(y\mid x)\right]
=\mathbb E_{x,y}\!\left[r(x,y)\sum_{t=1}^{|y|}\nabla_\theta\log\pi_\theta(y_t\mid h_t)\right].
\]
The reward weights a direction that makes the sampled response more likely. A gradient-ascent update takes the form $\theta\leftarrow\theta+\eta\widehat g$, where $\widehat g$ estimates this expectation and $\eta$ is the learning rate. GRPO changes how the responses are weighted and how their probabilities enter the objective.

\subsection{Verifiable and Random Rewards}

In RL with verifiable rewards (RLVR), a task-specific checker supplies the reward. Let $V(x,y)\in\{0,1\}$ indicate whether response $y$ answers question $x$ correctly. Our ground-truth (GT) condition uses $r(x,y)=V(x,y)$. On arithmetic, the checker evaluates the extracted final answer, so different solutions can receive the same reward. On code, it executes the generated program against tests.

The Random condition instead assigns an independent fair coin flip to every response. For $G$ responses to one prompt,
\[
r_1,\ldots,r_G\stackrel{\mathrm{iid}}{\sim}\operatorname{Bernoulli}(1/2),
\qquad (r_1,\ldots,r_G)\perp(x,y^{(1)},\ldots,y^{(G)})
\]
conditional on the preceding training history. Correct and incorrect responses therefore have the same chance of receiving a reward. This differs from rewarding incorrect answers, $r(x,y)=1-V(x,y)$, which still uses correctness information.

\subsection{Group-Relative Advantages}

Group Relative Policy Optimization (GRPO) compares responses to the same question. If $\pi_{\rm old}$ generated the group, its reward mean, population standard deviation, and response advantages are
\[
\bar r=\frac1G\sum_{i=1}^G r_i,
\qquad S_r=\sqrt{\frac1G\sum_{i=1}^G(r_i-\bar r)^2},
\qquad
A_i=\begin{cases}(r_i-\bar r)/S_r,&S_r>0,\\0,&S_r=0.\end{cases}
\]
A positive advantage favors a response relative to the others in the group; a negative advantage discourages it. For example, rewards $(1,0,1,0)$ give advantages $(1,-1,1,-1)$. Under GT, these signs distinguish correct from incorrect responses. Under Random, they depend only on the coin flips.

When every reward is equal, every advantage is zero. Otherwise, the response's advantage is applied to all of its generated tokens. It does not assign correctness to individual reasoning steps. Appendix~\ref{app:audit} gives the numerical stabilization used in the experiments.

\subsection{The GRPO Objective and Clipping}

GRPO uses the clipping rule of Proximal Policy Optimization (PPO) to limit the incentive to move away from the policy that generated the responses. Define the token probability ratio
\[
\rho_{it}(\theta)=\frac{\pi_\theta(y_t^{(i)}\mid h_t^{(i)})}
{\pi_{\rm old}(y_t^{(i)}\mid h_t^{(i)})},
\qquad h_t^{(i)}=(x,y_{<t}^{(i)}).
\]
The ratio exceeds one when the current policy assigns the token a higher probability than the old policy. We use $\rho_{it}$ for this ratio and $\rho$ without subscripts for the prompt distribution.

For clipping threshold $\varepsilon\in(0,1)$, the policy part of the GRPO objective is
\[
\begin{aligned}
u_t^{(i)}(\theta)
&=\min\!\left\{\rho_{it}(\theta)A_i,
\operatorname{clip}\!\left(\rho_{it}(\theta),1-\varepsilon,1+\varepsilon\right)A_i\right\},\\
J_{\rm GRPO}(\theta)
&=\mathbb E_{x\sim\rho,\,\{y^{(i)}\}_{i=1}^G\sim\pi_{\rm old}(\cdot\mid x)}
\left[\frac1G\sum_{i=1}^G\sum_{t=1}^{|y^{(i)}|}u_t^{(i)}(\theta)\right].
\end{aligned}
\]
Here $\operatorname{clip}(z,a,b)=\min\{\max\{z,a\},b\}$. The sampled responses, old-policy probabilities, and advantages remain fixed during differentiation. This is a surrogate objective evaluated on collected responses. Our implementation averages over response tokens, as described in Appendix~\ref{app:tokenmean}.

For positive advantages, clipping removes the incentive to increase the ratio past $1+\varepsilon$; for negative advantages, it removes the incentive to decrease it below $1-\varepsilon$. It does not impose a hard bound on the resulting parameter update. If old and new probabilities are computed identically at the same parameters, the ratios equal one and clipping leaves the gradient unchanged. Clipping can therefore be enabled without taking effect. Appendix~\ref{app:clip} examines this condition in our experiments.

\subsection{Reference Policies and Entropy}

The reference policy $\pi_{\rm ref}$ is usually a frozen copy of the starting model, whereas $\pi_{\rm old}$ is refreshed as training proceeds. A KL penalty can discourage departure from the reference. At a history $h$, the forward KL is
\[
D_{\rm KL}\!\left(\pi_\theta(\cdot\mid h)\,\|\,\pi_{\rm ref}(\cdot\mid h)\right)
=\sum_{a\in\mathcal V}\pi_\theta(a\mid h)
\log\frac{\pi_\theta(a\mid h)}{\pi_{\rm ref}(a\mid h)},
\]
where $\mathcal V$ is the vocabulary. This quantity is zero when the distributions agree. Differentiating a log-ratio on fixed sampled tokens does not generally give the gradient of this KL; Appendix~\ref{app:reference} gives the reference term used in our training.

Entropy measures how concentrated a policy's own probabilities are. Response entropy and next-token entropy are, respectively,
\[
\mathcal H(\pi_\theta;x)=-\sum_y\pi_\theta(y\mid x)\log\pi_\theta(y\mid x),
\qquad
\mathcal H_{\rm tok}(\pi_\theta;h)=-\sum_{a\in\mathcal V}\pi_\theta(a\mid h)\log\pi_\theta(a\mid h).
\]
Under the same generation and stopping rules, the chain rule gives
\[
\mathcal H(\pi_\theta;x)=\mathbb E_{y\sim\pi_\theta(\cdot\mid x)}
\left[\sum_{t=1}^{|y|}\mathcal H_{\rm tok}(\pi_\theta;h_t)\right].
\]
Our entropy diagnostic averages token entropies. It differs from the response-level sum above and from entropy over extracted answers. None of these quantities alone identifies whether the likely responses are correct. Token entropy is also the cost an entropy coder pays when a language model is used as a lossless compressor \citep{maoAcceleratingGeneralpurposeLossless2022,maoLiteByteEfficientFastAdapting2025}: low entropy means the model finds its own continuations predictable, not that they are right. Logarithms here are natural logarithms.

\subsection{Training Rewards and Evaluation Scores}

Evaluation measures correctness independently of the training reward. For an evaluation distribution $D$ and response sampler $q_\theta$, define
\[
p_\theta(x)=\mathbb E_{y\sim q_\theta(\cdot\mid x)}V(x,y),
\qquad F_{\rm sample}(\theta)=\mathbb E_{x\sim D}p_\theta(x).
\]
The sampler equals $\pi_\theta$ under ordinary policy sampling, but can differ under temperature scaling or nucleus sampling. Greedy accuracy evaluates the response obtained by selecting the most likely token at each step. With $k$ independent samples per question,
\[
\operatorname{pass@}k(\theta)=\mathbb E_{x\sim D}\left[1-(1-p_\theta(x))^k\right]
\]
measures the probability that at least one response is correct.

These scores describe what a fixed checkpoint produces. Reachability concerns how far training can improve that checkpoint under specified constraints (Definition~\ref{def:reach}). Repeated training runs let us estimate their expected outcome and variability. The best observed run shows an attained score; how often training attains such a score requires a separate estimate.

\section{Extended Related Work}
\label{app:extensive_literature_review}

The spurious-reward paradox connects two questions: where an improved answer comes from, and what the improvement tells us about the model. Work on contamination examines whether the answer was memorized; work on RL examines how training changes its probability. Probing raises the related question of whether a successful measurement reveals the model's knowledge or supplies what was missing. We discuss these connections below.

\subsection{Data Contamination and Memorized Answers}
\label{app:lit_contamination}

\citet{wuReasoningMemorizationUnreliable2026} investigate whether gains under weak or incorrect rewards depend on benchmark contamination. They find evidence of contamination in Qwen2.5 and introduce RandomCalculation, a generator of fresh arithmetic problems. Correct rewards produce sustained gains on these problems in their experiments, whereas random and incorrect rewards do not. Their result shows why gains on familiar benchmarks require further examination: a model can become better at returning an answer without becoming better at solving a new problem.

\citet{yanSpuriousRewardsParadox2026} approach this question through a memorization circuit. Earlier work interprets feed-forward layers as key--value memories \citep{gevaTransformerFeedForwardLayers2021} and uses interventions in middle-layer MLPs to study factual associations \citep{mengLocatingEditingFactual2022}. Yan et al. combine partial-prompt evaluations, activation patching, and neuron interventions to argue that middle layers retrieve memorized answers while later layers adapt to the retrieved signal. They distinguish these results from smaller gains on LiveMathBench, which they tentatively attribute to format alignment or dormant capability. Their account thus concerns the source of particular benchmark gains.

Related mechanistic work further separates whether information is stored by a model from whether
it is successfully expressed in its output. \citet{yuMechanisticUnderstandingMitigation2024} identify two distinct
failure modes behind non-factual hallucinations: insufficient factual information in lower-layer
MLPs and failures in upper-layer attention to select the appropriate factual attribute. Thus, the same
incorrect behavior can arise from substantially different internal states.
\citet{hongIntrinsicTestUnlearning2025} find a related dissociation in machine unlearning: methods can
suppress a concept behaviorally while leaving recoverable traces of the corresponding parametric
knowledge. \citet{hongReasoningMemorizationInterplayLanguage2025} further identify residual-stream features
that distinguish reasoning from memory recall and show that intervening on these features can
causally shift model performance between the two. Together, these results motivate distinguishing
whether information has been stored, whether it is accessible to the computation being performed,
and whether it is successfully used in the final response.

\citet{niuIllusionAlgorithmInvestigating2025} examine a related distinction in in-context learning using tasks built from random token sequences. Models can follow novel patterns, although performance remains sensitive to token frequency and task configuration. Their findings distinguish generalization beyond memorized instances from independence of pre-training statistics.

In our view, remembering an answer and learning how to answer should be distinguished. A model that applies a familiar arithmetic procedure to a new problem is generalizing, even if it has seen many similar problems. Synthetic instances reduce the opportunity to retrieve a specific question--answer pair, though template generation alone cannot certify absence from a large training corpus. Prior exposure to the underlying skill is part of what we want to measure. The question is why further training can help one checkpoint use that skill more readily than another.

\subsection{RL Training and Pre-existing Capabilities}
\label{app:lit_dynamics}

\citet{shaoSpuriousRewardsRethinking2026} show that format, incorrect, and random rewards can improve \exampleb{Qwen2.5-Math}, with less consistent benefits across other model families. They propose that clipping in GRPO \citep{shaoDeepSeekMathPushingLimits2024} creates asymmetric updates favoring behavior to which the model already assigns high probability. Their Appendix B.2 also reports occasional high-performing runs without clipping. This leaves room for clipping to affect stability without being necessary for every gain. The reward conditions themselves also differ: incorrect-answer rewards consult the verifier, format rewards inspect the response, and independent random rewards do neither.

\citet{chenExplorationVsExploitation2025} argue that clipping bias alone does not explain the gains. They relate clipping to entropy reduction and stability, but find that performance can improve with either increasing or decreasing entropy. Their reward-misalignment analysis compares random assignments with correctness-based assignments within a response group. A model that produces more correct responses has fewer incorrect trajectories available to reward by chance. This gives current task performance a direct role in explaining which models benefit.

More broadly, the information and granularity supplied by the reward can substantially change the
outcome of RL training. \citet{caoEnhancingReinforcementLearning2024} use LLM-generated critiques to turn
sparse sequence-level feedback into token- or span-level intermediate rewards, improving learning
efficiency and final performance across text-generation tasks. Their setting enriches the reward
channel, whereas our random-reward intervention deliberately removes correctness information from
it. Both make explicit that what further training can attain depends not only on the starting policy,
but also on what information the adaptation procedure is allowed to provide.

Our shared-rollout experiment asks about a further part of the process. Two learners receive identical responses and identical random rewards, so each pair has the same rewarded-incorrect and unrewarded-correct responses. We then compare how the learners change (Appendix~\ref{app:shared-rollouts}). The experiment holds the training material fixed to examine the learner's contribution. It does not contradict Chen et al.'s on-policy analysis, whose assumptions and object of study differ from this offline comparison.

Related work asks whether correctness-guided RL mainly improves the use of capabilities acquired during pre-training. \citet{yueDoesReinforcementLearning2025} find that RL gains at small sampling budgets can coexist with broader problem coverage by the base model at large pass@$k$. \citet{zhaoEchoChamberRL2025} vary pre-training data and model scale, finding that post-training amplifies patterns established during pre-training. These results connect post-training behavior to the model's history. They do not establish that a finite sampling budget recovers everything a model could learn to do, or that correct rewards supply no task information. At the parameter level, quantized LLM weights remain substantially compressible \citep{mao2024compressibilityquantizedlargelanguage,wangWhenCompressionMeets2024,wangECMEnhancingCompressibility2026}. This redundancy is consistent with small updates changing which existing behaviors a model uses, although we do not test that connection.

\citet{clayDemystifyingReinforcementLearning2026} study the interaction between the starting model, reward, and training questions. Rewards for intermediate progress can enable learning where final-answer-only rewards fail. Under random rewards, mathematical prompts can produce different outcomes from broader prompt collections: \exampleb{Qwen2.5-Math} improves under narrow mathematics training but shows little improvement on MATH and declines on AMC under WildChat prompts. They also observe different patterns of degradation under broad and mathematics-only training in OLMo-3. Their experiments retain meaningful questions, whereas our input controls also replace question content with random tokens.

Prompting provides a related way to change task performance without updating model parameters. \citet{jinImpactReasoningStep2024} find that expanding the reasoning steps in chain-of-thought demonstrations can improve accuracy, while compressing them can reduce it. They also find that some arithmetic tasks are relatively insensitive to an incorrect intermediate step. Their illustrated error control retains the correct final answer, however, so an incorrect rationale is not equivalent to removing correctness information. These results highlight how the form of the input can affect performance, complementing our analysis of training inputs. Input tokens can shape predictions without carrying task content: \citet{niuLlamaSeeLlama2025} find that LLMs assign higher logits and probabilities to tokens that appear in the context, even when those tokens are sampled at random. Random-token inputs therefore remove the mathematical problem without making the prompt inert.

The development of the starting model is also visible through in-context learning. Across Pythia checkpoints, \citet{niuIllusionAlgorithmInvestigating2025} find that pattern following with frequent tokens develops earlier than with rare tokens, while tasks requiring integration of linguistic and factual information develop later and more gradually. Reachability probing complements this analysis by measuring how the response to further parameter updates changes across checkpoints.

Complementary work traces changes in a model's internal organization during training and across
network depth. Tracking Pythia through its pre-training checkpoints,
\citet{leeGeometricSignaturesCompositionality2025} find that the relationship between linguistic compositionality and
representation geometry develops as linguistic features are learned. At the layer level,
\citet{chengEmergenceHighDimensionalAbstraction2024} identify a high-dimensional representational phase associated
with linguistic abstraction, transfer to downstream tasks, and language-modeling performance.
\citet{jinFartherShiftSparser2026} find that last-hidden-state representations become sparser as
task difficulty increases and, by pretraining small transformers from scratch, that
representational density develops as a learned property of data familiarity.
These studies do not measure reachability, but they provide complementary evidence that a model's
observable task performance need not fully characterize the internal state from which subsequent
computation or learning proceeds.

Taken together, these studies make the conditions of further training part of the question. Current accuracy describes the responses a model already produces; sampling examines what else it can produce without changing its parameters; adaptation examines what becomes attainable after updates. We use reachability for this last property, specifying the training procedure, available information, and budget. Random rewards remove correctness feedback from the reward channel while leaving the model and its training inputs able to affect what happens next.

\subsection{Probing, Information, and the Use of Knowledge}
\label{app:lit_probing}

Probes test what can be recovered from a model's representations. Linear classifiers at intermediate layers measure their usefulness for prediction \citep{alainUnderstandingIntermediateLayers2017}; diagnostic NLP tasks examine linguistic properties \citep{conneauWhatYouCan2018}; structural probes recover syntactic distances and depths through a transformation of word representations \citep{hewittStructuralProbeFinding2019}. The interpretation depends on the representation, predictor, training data, and evaluation criterion \citep{belinkovProbingClassifiersPromises2022}. The evaluation criterion matters for behavioral assessments as well: \citet{niuGrammaticalityLanguageModelling2020} show that conclusions about language models' grammatical knowledge depend on how model probabilities are scored against acceptability judgments, and argue for the point-biserial correlation.

The predictor's own learning creates an attribution problem. \citet{hewittDesigningInterpretingProbes2019} introduce control tasks with random labels assigned to word types, measuring whether a probe can memorize a mapping unrelated to the linguistic property of interest. Their selectivity measure compares performance on the linguistic and control tasks. \citet{voitaInformationTheoreticProbingMinimum2020} instead use minimum description length to account for the model complexity or training data needed to obtain a good predictor. Both go beyond final probe accuracy when deciding what credit belongs to the representation.

\citet{niuUsingRoarkHollingsheadDistance2022} address attribution through controlled comparisons of information sources. Their RH Probe combines BERT representations with part-of-speech tags and Roark--Hollingshead syntactic distances, then removes sources during training or perturbs them during evaluation. These experiments show that BERT contributes useful syntactic information while explicit syntactic encodings can still improve parsing. They also note that a parameter-free extraction algorithm can supply structural assumptions of its own.

The preference for a simple probe is itself disputed. \citet{pimentelInformationTheoreticProbingLinguistic2020} formulate probing as estimating mutual information between representations and annotations, for which a more expressive predictor can improve the estimate. Restricting the predictor may conceal information that is present but difficult to extract. Information content and ease of extraction are therefore different objects of study. Methods that order representation dimensions by importance, such as variational nested dropout \citep{cuiVariationalNestedDropout2021}, make a related point: how much of a representation is useful depends on how much of it the reader is allowed to use. Our use of training response concerns the latter distinction without requiring this debate to be resolved in favor of one probe architecture.

\citet{zhuInformationTheoreticView2020} connect these accounts by decomposing probing cross-entropy into label entropy, representation--label mutual information, and the probe's error in approximating the conditional label distribution. They also show that cross-entropy versions of selectivity and information gain have closely related estimation errors, with an equivalence under idealized randomization assumptions.

Subsequent formulations make the reader of a representation explicit. \citet{pimentelBayesianFrameworkInformationTheoretic2021} account for an agent's background knowledge and finite data through Bayesian mutual information. \citet{hewittConditionalProbingMeasuring2021}, using $\mathcal V$-information \citep{xuTheoryUsableInformation2019}, measure what a representation adds beyond a baseline for a specified predictive family. Both distinguish preserved information from information useful to a particular predictor. \citet{zhuPredictingFinetuningPerformance2022} connect these measurements to subsequent learning by predicting supervised fine-tuning performance from linguistic probing scores. For each of six GLUE classification tasks, three selected task--layer scores predict fine-tuning performance with lower error than random-feature baselines. Reachability asks a related question about the model's own adaptation: what performance can further training attain under the allowed procedures and budget?

Successful decoding also leaves open whether the model uses the decoded property. \citet{ravichanderProbingProbingParadigm2021} show that representations can encode properties unnecessary for the original task. Amnesic probing instead removes information associated with a property and measures the behavioral effect \citep{elazarAmnesicProbingBehavioral2021}. \citet{tibliasHypothesisDrivenFeatureManifold2025} combine geometric analysis with intervention: temporal feature manifolds change with the task prompt, and perturbing their identified subspaces impairs reasoning. Comparisons with random subspaces help establish the functional relevance of the recovered structure.

Recent work similarly combines representational readout with causal intervention.
\citet{jiCalibratingVerbalUncertainty2025} identify verbal uncertainty as a linear feature in LLM
representations and show that manipulating the feature changes expressed uncertainty and can reduce
confident hallucinations. \citet{yuRobustLLMSafeguarding2024, shayeganiMisalignedRolesMisplaced2025} study a refusal feature in the residual stream,
show that suppressing it reproduces a mechanism shared by adversarial jailbreaks, and use this
mechanistic account to construct an efficient adversarial-training procedure.
\citet{yangHallucinationReductionCASAL2025} take a further step by amortizing an activation-steering intervention
into model weights, so that a representational control signal discovered at inference time becomes
a persistent behavioral change after training. These examples illustrate a progression from decoding
a property, to establishing a functional role through intervention, to altering what a model reliably
produces after adaptation. Reachability is complementary: rather than prespecifying an internal
feature, it asks what task behavior training can make attainable from the checkpoint under specified
information and optimization constraints.

Similar care is needed when interpreting where knowledge resides: \citet{tenneyBERTRediscoversClassical2019} interpret layerwise results as an NLP-like processing sequence, while \citet{niuDoesBERTRediscover2022} find that token position, probe training stage, and random seed complicate that account. Extending layerwise probing to LLMs, \citet{jinExploringConceptDepth2025} train linear classifiers on factual, emotional, and reasoning tasks and compare the depths at which probe performance improves or stabilizes. Their concept depth characterizes supervised readout across network layers, complementing our comparison of training responses across pre-training checkpoints.

\citet{niuWhatDoesKnowledge2024} find that edits to factual associations often fail to generalize across inverse relations and synonymous expressions. Their results distinguish the ability to alter a token prediction from stronger claims about knowledge storage. Related ambiguity arises in circuit and sheaf discovery \citet{yuSheafDiscoveryJoint2025}. \citet{chenAllCircuitsLead2026} hold model weights fixed and recover structurally distinct, low-overlap mechanisms that support the same task. Their results separate functional sufficiency from claims of a unique internal mechanism. A probe can recover real structure without establishing the larger explanation proposed for it.

For generative models, \citet{orgadLLMsKnowMore2024} find internal information about correctness that is not always expressed in the generated answer. Tuned lenses decode intermediate states into vocabulary distributions \citep{belroseElicitingLatentPredictions2025}, and sparse autoencoders learn features from activations \citep{brickenMonosemanticityDecomposingLanguage2023}. The predictive distribution itself can also be read out as a code length, as in learned lossless compressors for images and gigapixel pathology slides \citep{maoEaszAgileTransformerbased2025,maoWISEFrameworkGigapixel2025}. These methods do not all use task labels, so the absence of label supervision alone does not distinguish our approach from every internal readout. Our measurement instead requires the model itself to perform the task after adaptation. Random rewards allow us to study this response without supplying correct-answer feedback during training. The resulting measurement still depends on the adaptation procedure, and does not establish that a complete successful procedure existed unchanged in the starting checkpoint.

\section{Reachability: Definitions and Theoretical Foundations}
\label{app:reachability-theory}

We first give the definition used in \S\ref{sub:reachability-probe} and relate it to the reported training outcomes. We then prove that changing the answer verifier leaves pure random-reward training unchanged, and bound the information introduced by selecting a trained candidate. The final calculation shows how random rewards can produce updates whose distribution depends on the model and its inputs.

\subsection{Reachability under Specified Training Constraints}
\label{app:scope}

Fix an initial checkpoint $\theta_0$, a task performance measure $J$, a training budget $B$, and a prespecified family $\mathfrak A$ of adaptation procedures. This family specifies the training inputs and supervision, permissible updates, and selection and stopping rules. Let $\mathcal A_B(\theta_0;U)$ be the checkpoint returned by procedure $\mathcal A$ within budget $B$, with randomness $U$.

\begin{definition}[Reachability]\label{def:reach}
Under the specified task, family, and budget, define
\[
\operatorname{Reach}_B(\theta_0)
=\sup_{\mathcal A\in\mathfrak A}\mathbb E_U[J(\mathcal A_B(\theta_0;U))],
\qquad
\Delta_{\rm reach}(\theta_0)=\operatorname{Reach}_B(\theta_0)-J(\theta_0).
\]
\end{definition}

The constraints determine what the model is allowed to learn from and how it may be trained. Varying $B$ describes how much training is needed to attain a given expected score. Random-reward probing restricts the family to correctness-free procedures $\mathfrak A_0$. For a fixed random-reward procedure in this family,
\[
\Delta_{\mathrm{rand}}(\theta_0)
=\mathbb E_U[J(\mathcal A_{\mathrm{rand},B}(\theta_0;U))]-J(\theta_0)
\leq\Delta_{\rm reach}(\theta_0),
\]
where reachability on the right is defined over $\mathfrak A_0$. The inequality concerns the population expectation. The mean over a finite number of runs estimates that expectation and is subject to sampling error.

A fixed procedure also produces a distribution of trajectories. Let $H_t$ include the parameters, optimizer memory, and mutable sampler state, and let $F(H_t)$ evaluate the parameters contained in that state. For a protocol $P$ and $T$ updates, define
\[
\mathcal L_{P,T}(\theta_0)=\operatorname{Law}_U\bigl((F(H_t))_{t=0}^{T}\bigr),
\]
and, for a target score $\tau$,
\[
R_P(\theta_0;T,\tau)=\Pr_P\!\left(\max_{0\le t\le T}F(H_t)\ge\tau\right),
\qquad
R_P^{\rm end}(\theta_0;T,\tau)=\Pr_P(F(H_T)\ge\tau).
\]
These are additional summaries of training response. The first records reaching a target within the budget; the second records retaining it at the endpoint, so $R_P^{\rm end}\leq R_P$. They complement the expected-performance definition above. Our experiments observe trajectories at scheduled evaluations, with the corresponding estimates given in Appendix~\ref{app:scan-readout}.

\subsection{Verifier-Intervention Invariance}
\label{app:t1}

Consider running the same training procedure twice with the same random draws, but different answer checkers. If the checker supplies neither training inputs nor decisions, the two runs follow the same path. Correctness can still be recorded for evaluation, provided that calculation does not affect subsequent training or its random state.

Let $Z=(H_0,X_{0:T-1},U_{0:T-1},\Xi_{0:T-1})$ have the same joint law under interventions on the verifier $V$. Here $X$ supplies problems, $U$ supplies response-sampling and other algorithmic randomness, and $\Xi$ supplies reward randomness. For measurable maps independent of $V$, write
\[
Y_t=S(H_t,X_t,U_t),\qquad R_t=h(\Xi_t),\qquad
H_{t+1}=\mathcal A_P(H_t,X_t,Y_t,R_t,U_t).
\]
Adaptive problem selection can be included in the state and algorithmic randomness, as long as it is also verifier-independent. With fresh independent reward draws and $C_t=V(X_t,Y_t)$, we additionally have $I(R_t;C_t\mid H_t,X_t)=0$. This conditional independence concerns the reward's correctness information; the following theorem concerns the entire training trajectory.

\begin{theorem}[Verifier-intervention invariance]\label{thm:t1}
Suppose the initial state and exogenous inputs have a joint law invariant under interventions on $V$, and the state update has no verifier dependence. Then changing $V$ leaves the law of $H_{0:T}$ unchanged. Under a coupling with identical initial state and exogenous inputs, the trajectories agree pathwise.
\end{theorem}

\begin{proof}
The coupled runs share $H_0$. If they share $H_t$, they have the same sampler inputs and therefore the same response $Y_t$. They also have the same reward inputs and reward $R_t$. Every argument of $\mathcal A_P$ then agrees, giving the same $H_{t+1}$. Induction establishes equality throughout training. The coupling is valid because the joint law of the initial state and exogenous inputs is unchanged by the verifier intervention.
\end{proof}

The protocol and checkpoint are fixed in this statement. They may previously have been chosen using labeled evaluations; the theorem does not erase that history. Joint invariance is also necessary: keeping each input's marginal distribution fixed is insufficient if their dependence changes. Finally, verifier invariance places no requirement on the mean update. A label-free objective can produce directed changes while satisfying the theorem.

\begin{corollary}[Equivalence to an unlabeled randomized algorithm]
\label{cor:unlabeled}
Removing verifier-dependent diagnostics from pure random-reward training leaves its trajectory law unchanged, provided those diagnostics do not affect updates, scheduling, or training randomness.
\end{corollary}

The model can still learn from prompts and self-generated responses. The result excludes correctness feedback through this training loop; it does not imply that the complete capability existed before adaptation. GT and steering controls fall outside its assumptions, and label-based evaluation is separate from the invariant training trajectory. We do not use $I(\theta_T;y^\star\mid x)=0$ to express the result: when the correct answer $y^\star$ is a deterministic function of $x$, that equality holds regardless of how the model was trained.

\subsection{Information Introduced by Selecting a Run}
\label{app:bits}

Training without correctness feedback and choosing a result using that feedback are separate operations. Choosing among $N$ fixed candidates can communicate an index of at most $\log_2N$ bits. This limits the information carried by the choice, though a small amount of selection information can correspond to a large score difference.

Let $V$ be a random verifier under a specified prior and $C=(C_1,\ldots,C_N)$ the candidate collection. Any other fixed context can be included in the conditioning below.

\begin{lemma}[Selection bits]\label{lem:bits}
If $K$ selects one of the $N$ candidates, then
\[
I(V;C_K\mid C)\le I(V;K\mid C)\le H(K\mid C)\le\log_2N.
\]
\end{lemma}

\begin{proof}
The selected candidate is a deterministic function of $(C,K)$, so conditional data processing gives the first inequality. The second follows from $I(V;K\mid C)=H(K\mid C)-H(K\mid C,V)$. An index taking at most $N$ values has entropy at most $\log_2N$.
\end{proof}

If the initial state and exogenous inputs are also jointly independent of the random verifier, Theorem~\ref{thm:t1} gives $I(V;C)=0$. The chain rule then yields the unconditional bound $I(V;C_K)\leq\log_2N$. Selecting among eight runs can therefore introduce at most three bits under these assumptions. This is not a bound on the selected run's improvement: one bit can choose between a perfect and a poor candidate. It also excludes information conveyed by reporting full scores or by earlier label-dependent protocol selection.

\subsection{Moments of the Random-Reward Gradient}
\label{app:t3}
\label{sub:gradient}

With fair-coin rewards, swapping zeros and ones preserves the probability of a reward group but reverses every advantage. For fixed responses, the two groups give opposite raw policy gradients. Their mean cancels, while their covariance can still depend on the responses and the model.

\begin{theorem}[Moments of the random-reward gradient]\label{thm:t3}
Fix $\theta,x$ and $G>1$. Let $Y_i$ be independent draws from $\pi_\theta(\cdot\mid x)$, with scores $S_i=\nabla_\theta\log\pi_\theta(Y_i\mid x)$ of zero mean and finite second moment. Let the reward-only advantages be exchangeable, independent of the responses, and satisfy $\sum_iA_i=0$ and $a_G=\mathbb E[A_i^2]<\infty$. For $\widehat g=G^{-1}\sum_iA_iS_i$,
\[
\mathbb E[\widehat g\mid\theta,x]=0,\qquad
\operatorname{Cov}(\widehat g\mid\theta,x)=\frac{a_G}{G}F(\theta;x),
\]
where $F(\theta;x)=\mathbb E[S_iS_i^\top]$.
\end{theorem}

We prove a more general form that also permits a sampler $q_\theta$ different from the model distribution. Let $Y_i$ be iid draws from $q_\theta(\cdot\mid x)$ and $S_i=s_\theta(Y_i,x)$ have finite second moment. Keep the same assumptions on the reward-only advantages. Exchangeability and centering give
\[
\mathbb E A_i=0,\qquad
\mathbb E[A_iA_j]=-\frac{a_G}{G-1}\quad(i\ne j),
\]
where the second identity follows by expanding $\mathbb E[A_i\sum_jA_j]=0$.

\begin{lemma}[Zero expected raw gradient]
\label{lem:zeromean}
$\mathbb E[\widehat g\mid\theta,x]=0$.
\end{lemma}
\begin{proof}
Condition on the responses. The score vectors are fixed and each reward-only advantage has mean zero, giving $\mathbb E[\widehat g\mid Y_{1:G},\theta,x]=0$. Taking the remaining expectation proves the claim.
\end{proof}

\begin{lemma}[Score covariance]
\label{lem:cov}
$\operatorname{Cov}(\widehat g\mid\theta,x)=(a_G/G)\operatorname{Cov}_{q_\theta}(S)$.
\end{lemma}
\begin{proof}
Write $m=\mathbb E_qS$ and $M=\mathbb E_q[SS^\top]$. Independence of the responses and advantages gives
\[
\mathbb E[\widehat g\widehat g^\top]
=\frac1{G^2}\left(Ga_GM+G(G-1)\frac{-a_G}{G-1}mm^\top\right)
=\frac{a_G}{G}(M-mm^\top).
\]
The mean is zero by Lemma~\ref{lem:zeromean}, so this second moment is the covariance. Under exact policy sampling and the score identity, $m=0$ and $M=F$, proving Theorem~\ref{thm:t3}.
\end{proof}

For Bernoulli$(1/2)$ rewards with exact population-standard-deviation normalization, $a_G=1-2^{1-G}$: every nonconstant group has mean squared advantage one, and constant groups have probability $2^{1-G}$. Holding the score covariance fixed, the multiplier $a_G/G$ is about 1.764 times larger at $G$ = 4 than at $G$ = 8. Related group-size factors appear in \citet{chenExplorationVsExploitation2025}. In actual training, nucleus sampling changes the score distribution, token averaging introduces response lengths, and AdamW transforms the gradient. The factor alone does not predict a universal group-size ordering of performance. Appendix~\ref{app:impl} gives these implementation details.

\paragraph{From gradients to behavior.}
A zero-mean gradient need not leave expected behavior unchanged. For a smooth statistic $f$, a second-order expansion under suitable regularity gives
\[
\mathbb E[f(\theta+\eta\widehat g)-f(\theta)\mid\theta]
=\frac{\eta^2}{2}\operatorname{tr}\!\left(\nabla^2f(\theta)C(\theta)\right)+o(\eta^2),
\qquad C(\theta)=\operatorname{Cov}(\widehat g\mid\theta).
\]
The covariance and local curvature jointly determine this term, whose sign need not be positive. Thus a zero raw mean does not make behavioral probabilities martingales or guarantee improved performance. Greedy accuracy is not generally smooth, so this expansion is an intuition for smooth behavioral quantities rather than a formula for the accuracy changes in our experiments.

\paragraph{The contribution of training inputs.}
Let $\rho$ be the prompt distribution. Since the conditional gradient mean is zero at every prompt, the law of total covariance gives
\[
\operatorname{Cov}(\widehat g\mid\theta)
=\frac{a_G}{G}\mathbb E_{x\sim\rho}
\left[\operatorname{Cov}_{Y\sim q_\theta(\cdot\mid x)}\bigl(s_\theta(Y,x)\bigr)\right].
\]
Changing the prompts changes which responses are sampled and which parameter directions receive updates, even when the reward law stays fixed. This is why training inputs belong among the conditions of reachability. The calculation establishes this dependence without predicting which prompt condition will improve GSM performance.

\section{Experimental Setup}
\label{app:exp}
\label{app:detailed_exp_setup}
\label{app:scan}

We compare checkpoints under fixed training conditions, then vary the training inputs and the source of the responses to examine what contributes to the observed gains. This section gives the data, protocols, and evaluation rules. Appendix~\ref{app:results} reports the outcomes for every seed.

\subsection{Models and Tasks}
\label{app:tasks}

\subsubsection{OLMo-2 training checkpoints}
\label{app:checkpoints}

Our main model is \exampleb{OLMo-2-1124-7B}. P1 denotes its first stage of pre-training, ending at 3,896B tokens; P2 denotes the ingredient-1 mid-training continuation. A label such as P2 +5B counts additional tokens after P1. The GSM scan includes 25 P1 and 12 P2 checkpoints. Code and lookup use eight P1 checkpoints, at 5B, 462B, 839B, 1,259B, 2,098B, 2,937B, 3,532B, and 3,896B, and four P2 checkpoints, at +5B, +13B, +26B, and +50B. The complete GSM coverage appears in Tables~\ref{tab:gsm-gt-full} and~\ref{tab:gsm-random-full}.

The additional OLMo-2 controls use the P1 endpoint, P2 +5B, and P2 +50B. The earlier exploratory study also includes P2 +21B. These checkpoints let us compare the end of pre-training with both the beginning and end of mid-training while keeping architecture and parameter count fixed.

\subsubsection{Synthetic arithmetic}

We generate 1,000 arithmetic word problems from templates and hold out 200 for evaluation. The split separates problem instances; training and evaluation can share template families. Each question follows the instruction
\begin{quote}
\exampleg{Solve the math problem. Show your reasoning and put the final answer on a line like: \#\#\#\# <number>.}
\end{quote}
The worked answer is not given as a target in random-reward training. Both training generation and greedy evaluation use a 96-token response cap.

Numerical correctness uses an absolute tolerance of $10^{-6}$. The primary, lenient extractor checks for a number following \exampleg{\#\#\#\#}, an annotated equation, an answer phrase, and finally the last number in the response, in that order. A separate strict score requires a standalone \exampleg{\#\#\#\# <number>} line. We report both because a correct answer can appear without the requested format.

OLMo-2 mid-training includes GSM-style material from TinyGSM-MIND and the GSM8K training split, alongside changes in the broader data mixture and learning-rate schedule. The checkpoint comparison therefore includes changing task exposure. Generated evaluation problems reduce dependence on particular benchmark answers without excluding familiarity with the underlying procedures.

\subsubsection{Code generation}

The MBPP dataset \citep{odenaProgramSynthesisLarge2021} used here contains 964 problems, with 764 for training and 200 for evaluation. Each prompt includes the first test:
\begin{quote}
\exampleg{Write a Python function to solve the following problem. Only output the function definition, no explanation.}\par\smallskip
\exampleg{Problem: <question>}\par
\exampleg{Your code should pass this test: <first test>}\par
\exampleg{Solution:}
\end{quote}
Responses are capped at 256 tokens.

The evaluator extracts the code, appends the test setup and tests, and executes it with a ten-second timeout. A problem is correct only if execution completes and all tests pass. Accuracy is the fraction of problems solved. The visible test supplies task information in both GT and Random conditions. Random rewards remove feedback about whether the generated program passed, while preserving the information already in its prompt.

\subsubsection{Compositional lookup}

We generate 1,000 problems using invented people and cities. Each problem contains eight people, four cities, and four pairs of people. Each pair contributes a relation and a city fact, giving eight facts presented in shuffled order. One-hop questions ask where a named person lives. Two-hop questions combine a relation with a location, for example \exampleg{A's teacher is B} and \exampleg{B lives in C}, followed by \exampleg{Where does A's teacher live?}. The instruction is
\begin{quote}
\exampleg{Answer the question using the facts. Put the final answer on a line like: \#\#\#\# <answer>.}
\end{quote}

The generator alternates one- and two-hop questions. The fixed evaluation set contains 104 one-hop and 96 two-hop items, which are pooled in the main score. The checker compares the extracted answer with the correct city. Uniform choice among the four cities would give 25\% accuracy, shown as a reference in Figure~\ref{fig:lookup-development}; it is not an assumption about the model's actual starting policy.

\subsection{Checkpoint-Scan Protocol}
\label{app:training}
\label{app:scan-protocol}

Table~\ref{tab:scan-protocol} gives the training budget and seed counts for each task. We use split seed 777 independently of the training seed, so every checkpoint and reward condition within a task sees the same evaluation problems. GT uses seeds 31001--31004 and Random uses 31001--31008 where four and eight seeds are available. Smaller cells use the corresponding prefix of these sequences.

\begin{table}[htbp]
\centering\small
\caption{Checkpoint-scan settings. Seed counts are per checkpoint. GT and Random use reference coefficients 0 and 0.01, respectively, in addition to different rewards. Appendix~\ref{app:reference} defines the reference term.}
\label{tab:scan-protocol}
\begin{tabular}{lccc}
\toprule
 & Synthetic GSM & Code & Invented-fact lookup\\
\midrule
P1 / P2 checkpoints & 25 / 12 & 8 / 4 & 8 / 4\\
Training / evaluation problems & 800 / 200 & 764 / 200 & 800 / 200\\
GT seeds, P1 / P2 & 4 / 4 & 1 / 4 & 2 / 2\\
Random seeds, P1 / P2 & 8 / 8 & 2 / 8 & 8 / 8\\
Response cap (tokens) & 96 & 256 & 96\\
Training steps & 500 & 500 & 500\\
Group size & 4 & 4 & 4\\
Learning rate & $3\cdot10^{-5}$ & $3\cdot10^{-5}$ & $3\cdot10^{-5}$\\
Reference coefficient, GT / Random & 0 / 0.01 & 0 / 0.01 & 0 / 0.01\\
Completed training runs & 444 & 72 & 120\\
\bottomrule
\end{tabular}
\end{table}

At each step, we sample four responses to one prompt at temperature 1 and top-$p$ = 0.95. Training follows a fixed shuffled list; 500 updates visit only the first 500 training problems. Each group receives one optimizer update, with one PPO epoch and no gradient accumulation across groups. We use AdamW, learning rate $3\times10^{-5}$, and gradient-norm clipping at 1. LoRA adapters have rank 8, scaling 16, and zero dropout on the attention and MLP projections. Base weights remain frozen. PPO clipping is enabled with threshold 0.2.

GT rewards accepted responses with one and rejected responses with zero. Random assigns independent Bernoulli$(1/2)$ rewards. Both standardize rewards within each group using the population standard deviation and a $10^{-6}$ stabilizer. The reference coefficient is zero for GT and 0.01 for Random, so the comparison changes this setting as well as the reward. Appendix~\ref{app:reference} specifies the implemented reference term.

\subsection{Training-Input Controls}
\label{app:randq-setup}

We vary the training inputs at OLMo-2 P2 +50B while keeping random rewards and the 200-question GSM evaluation fixed. Table~\ref{tab:randq-inputs} lists the five conditions. Original GSM and the two random-token conditions use 32 seeds each; shuffled GSM and unrelated MMLU questions use 16. The original-GSM condition reuses eight scan runs and adds 24, giving 128 OLMo-2 outcomes in total.

\begin{table}[htbp]
\centering\small
\caption{The five training-input conditions. All use random rewards and are evaluated on the same GSM questions. The math template contains the instruction and the \exampleg{Question:} and \exampleg{Answer:} markers.}
\label{tab:randq-inputs}
\begin{tabular}{clcc}
\toprule
Arm & Training content & Math template & Seeds\\
\midrule
B & Original GSM question & Yes & 32 (8 reused)\\
C & Word-shuffled GSM question & Yes & 16\\
A & Unrelated MMLU question stem & Yes & 16\\
D & Random vocabulary tokens & Yes & 32\\
E & Random-token complete prompt & No & 32\\
\bottomrule
\end{tabular}
\end{table}

Conditions B, C, A, and D use the same template:
\begin{quote}
\exampleg{Solve the math problem. Show your reasoning and put the final answer on a line like: \#\#\#\# <number>.}\par\smallskip
\exampleg{Question: <training content>}\par
\exampleg{Answer:}
\end{quote}
Only the question content changes, and no worked example is supplied. Condition E replaces the complete prompt with a random string, removing the instruction and both markers. Evaluation always uses real GSM questions with the original template.

Condition C shuffles whitespace-separated words within each question, preserving its words, numbers, and attached punctuation. We try up to 60 permutations and choose the closest token-length match. Condition A draws digit-free MMLU question stems \citep{hendrycksMeasuringMassiveMultitask2020} without replacement from subjects outside a quantitative-subject filter, matching lengths as closely as possible. It supplies neither answer choices nor labels. The subject filter does not exclude every possible reference to quantitative concepts.

Condition D samples vocabulary tokens, excluding special tokens and tokens that individually decode to empty, non-printable, or replacement-character text. Digits and task-related words remain eligible. The decoded strings are retokenized and matched to the original question lengths. Condition E instead matches complete-prompt lengths and additionally excludes digits, the hash character, and tokens containing \exampleg{answer}, \exampleg{question}, \exampleg{solve}, \exampleg{math}, \exampleg{problem}, \exampleg{reason}, \exampleg{number}, or \exampleg{final}, ignoring case. Its eligible pool has 94,309 token IDs. The completed strings are checked for these exclusions. Input construction uses seed 20260920 for A, C, and D and 20260921 for E; each input set is then shared across training seeds.

Using the OLMo-2 P2 +50B tokenizer, C matches 799 of 800 question lengths exactly, with maximum deviation one token. A matches 525 exactly, with mean absolute deviation 0.565 and maximum three; D matches 564, with mean deviation 0.547 and maximum seven. E matches 799 complete-prompt lengths exactly, with maximum deviation seven. D and E thus differ in random strings, vocabulary filters, and length targets as well as the template. The comparison tests the two input conditions together.

All conditions use the scan's Random settings and 500-update budget. Correctness diagnostics do not enter the rewards or updates. New OLMo-2 runs start at 20.5\%; the eight reused B runs record 21.0\% despite sharing the evaluation questions. Gains use each run's own starting score, while the figures mark the separately evaluated frozen score of 20.5\%. The mean terminal Frobenius norms of the LoRA $B$ matrices are 2.099 in D and 2.298 in E, from zero initialization. Thus E does change the adapters, although these norms do not equate the two conditions' effective weight changes or update directions.

\subsection{Additional Model Families}
\label{app:crossmodel-setup}

We select four checkpoints each from \exampleb{SmolLM2-1.7B} and \exampleb{OLMo-3-7B} to examine development beyond OLMo-2 (Table~\ref{tab:crossmodel-revisions}). S1--S4 follow SmolLM2 training. O1 and O2 are an early OLMo-3 pre-training checkpoint and the P1 endpoint; O3 and O4 are early and late P2 checkpoints. They use the same 800 training and 200 evaluation GSM problems as OLMo-2.

\begin{table}[htbp]
\centering\small
\caption{SmolLM2 and OLMo-3 checkpoints used in the developmental comparison. P2 token counts are additional to P1.}
\label{tab:crossmodel-revisions}
\begin{tabular}{lll}
\toprule
ID & Training position & Training step\\
\midrule
\multicolumn{3}{l}{\exampleb{SmolLM2-1.7B}}\\
S1 & 0.26T & \texttt{step-125000}\\
S2 & 5.77T & \texttt{step-2750000}\\
S3 & 9.96T & \texttt{step-4750000}\\
S4 & 10.75T & \texttt{step-5125000}\\
\midrule
\multicolumn{3}{l}{\exampleb{OLMo-3-7B}}\\
O1 & P1 34B & \texttt{stage1-step8000}\\
O2 & P1 end, 5.93T & \texttt{stage1-step1413814}\\
O3 & P2 +4.2B & \texttt{stage2-step2000}\\
O4 & P2 +100B & \texttt{stage2-step47684}\\
\bottomrule
\end{tabular}
\end{table}

Each checkpoint has two GT seeds and eight Random seeds, giving 80 runs of 500 updates. Training follows the GSM settings in Table~\ref{tab:scan-protocol} and uses NVIDIA H200 GPUs. Frozen evaluation includes greedy decoding and 32 sampled responses per question, at temperature 1 and top-$p$ = 0.95, in four batches of eight with seed 777. The developmental comparison uses greedy scores throughout; the frozen sampling layout differs from the earlier OLMo-2 evaluation. GT and Random maxima also use different seed budgets, which the individual points make visible.

The input controls additionally use the released base models \exampleb{Qwen2.5-7B} and \exampleb{Llama-3.1-8B}. Both receive the same D and E strings constructed for OLMo-2, with eight seeds per condition and the same Random settings. Evaluation retains the math template. The separately evaluated frozen scores are 43.5\% for Qwen and 25.5\% for Llama. Qwen's pre-update scores range from 44.0\% to 44.5\%, while Llama's remain 25.5\%; gains again use each run's own baseline.

\subsection{Group-Size and Early-Steering Controls}
\label{app:protocol}
\label{app:historical}

An earlier 55-run study varies group size and the rewards used for the first 50 training steps. It uses 1,000 synthetic mathematics problems generated with seed 13, split into 800 training and 200 evaluation problems. In this study, the training seed also chooses the evaluation subset. Matching seeds therefore match evaluation questions across checkpoints and conditions, while different seeds can see different questions. We keep these outcomes separate from the later fixed-split scan.

All runs use 500 updates, evaluation every 100 steps, learning rate $3\times10^{-5}$, and a 96-token response cap. The five seeds, 777, 888, 999, 1001, and 1002, differ from the exploratory seeds 42, 123, and 456 used to select the protocol. We retain every completed run, including the low-scoring GT run at P2 +50B.

Table~\ref{tab:configs} gives the five conditions. At each of the three checkpoints, GT has five $G$ = 4 runs, and Random has five each at $G$ = 4 and $G$ = 8. The two steering conditions each have five $G$ = 8 runs at P2 +5B: they use GT or inverted-GT rewards for 50 steps, followed by Random for 450. These controls ask whether early direction affects later training. Since their first 50 steps use the verifier, their complete trajectories are not zero label-information probes.

\begin{table}[htbp]
\centering\small
\caption{Group-size and steering conditions in the earlier 55-run study. All use learning rate $3\times10^{-5}$, 500 updates, and evaluation every 100 steps. Steering supplies correctness-dependent rewards for the first 50 steps.}
\label{tab:configs}
\begin{tabular}{llcc}
\toprule
configuration & reward & reference coefficient & $G$\\
\midrule
\texttt{r\_g4} / \texttt{r\_g8} & independent random & 0.01 & 4 / 8\\
\texttt{gt\_g4} & ground truth & 0 & 4\\
\texttt{gt\_K50} & ground truth for 50 steps, then random & 0.01 & 8\\
\texttt{anti\_K50} & inverted ground truth for 50 steps, then random & 0.01 & 8\\
\bottomrule
\end{tabular}
\end{table}

\subsection{Protocol Selection}
\label{app:provenance}
\label{app:grid}

Before the 55-run study, an exploratory sweep covered 1,224 runs across four P2 checkpoints, three annealing mixtures, 34 configurations, and three seeds. It informed the choice of learning rate $3\times10^{-5}$ and reference coefficient 0.01. Its marginal gain rates were 11.4\% at $G$ = 4 and 10.9\% at $G$ = 8, unlike the sharper group-size contrast in the subsequent controls. We therefore treat the later result as a property of the tested conditions rather than a general advantage of smaller groups. Twelve code/lookup runs in the completed scan also come from the task-level hyperparameter search. These reused outcomes are included once and do not constitute independent validation of the chosen settings.

\subsection{Evaluation and Training Outcomes}
\label{app:evaluation}

\subsubsection{Terminal scores and threshold crossings}
\label{app:scan-readout}

We evaluate greedily at steps 0, 100, 200, 300, 400, and 500. Developmental plots show the maximum terminal score across the available seeds, together with individual outcomes. For checkpoint $c$, protocol $P$, and $N$ seeds, the maximum is
\[
M_{P,N}(c)=\max_{1\leq s\leq N}F(H_{500}^{(c,s)}).
\]
This gives a score attained by at least one run under the budget. It depends on the number of seeds and estimates neither mean performance nor a capability ceiling.

For threshold $\tau$ and evaluation schedule $\mathcal E=\{0,100,200,300,400,500\}$, we separately count terminal successes and observed visits:
\[
\widehat R^{\rm end}_{P,N}=\frac1N\sum_s\mathbf1\{F(H_{500}^{(s)})\geq\tau\},
\qquad
\widehat R^{\mathcal E}_{P,N}=\frac1N\sum_s\mathbf1\{\max_{t\in\mathcal E}F(H_t^{(s)})\geq\tau\}.
\]
Visits between evaluations can be missed, so their population probabilities satisfy $R_P^{\rm end}\leq R_P^{\mathcal E}\leq R_P$. Reaching a high score and retaining it through step 500 are different outcomes.

\subsubsection{Gains relative to the starting checkpoint}
\label{app:jackpot}

The earlier study defines a \emph{jackpot} as a terminal improvement greater than ten percentage points:
\begin{equation}
J_T=\mathbf1\{\operatorname{acc}(\theta_T)-\operatorname{acc}(\theta_0)>0.10\},
\qquad q_P(\theta_0,T)=\Pr_P(J_T=1).
\label{eq:readout}
\end{equation}
We estimate this probability separately for each checkpoint and protocol. The absolute 50\% threshold used in the scan answers a different question. A high intermediate score does not meet the terminal criterion if it is lost before training ends.

GSM frozen scores come from separate evaluations on the shared 200 questions. Code and lookup use the step-zero evaluation of GT seed 31001. Small discrepancies between separate frozen evaluations and a run's pre-update evaluation remain in the results; a run-specific gain always subtracts that run's own score. In the earlier study, the probability also averages over seed-dependent evaluation splits. In the scan, the questions are fixed and the randomness is in training.

\paragraph{Plotting trajectories.}
In the code and lookup developmental figures, checkpoints are equally spaced in training order, with the P1/P2 boundary marked. Input-control plots connect the recorded evaluations without smoothing. In Figure~\ref{fig:trajectories}, each rollout-correctness trace instead uses a trailing 20-step average, and the thick curve is the pointwise maximum over those smoothed traces. That maximum can come from different seeds at different steps. The figure includes all four GT and eight Random seeds at each selected checkpoint.

\subsection{Update Rule and Its Relation to the Gradient Analysis}
\label{app:impl}

\subsubsection{Advantage normalization}
\label{app:audit}

The scan divides centered rewards by $s+\varepsilon$ when their population standard deviation $s$ exceeds $\varepsilon=10^{-6}$, and otherwise returns the centered rewards. For Bernoulli$(1/2)$ rewards at $G$ = 4 or $G$ = 8, only constant groups take the latter branch, giving zero advantages. If $m$ rewards equal one and $s_m=\sqrt{(m/G)(1-m/G)}$, then
\[
a_G(\varepsilon)=2^{-G}\sum_{m=1}^{G-1}\binom Gm\frac{s_m^2}{(s_m+\varepsilon)^2}.
\]
Without stabilization, this is $1-2^{1-G}$, giving $a_4=0.875$ and $a_8=0.9921875$. These are the factors in the sequence-mean gradient calculation of Appendix~\ref{app:t3}.

\subsubsection{Token averaging and response length}
\label{app:tokenmean}

Our loss averages over selected response tokens. Let $n_i$ be the number of selected tokens in response $i$, $S_i$ their summed score vector, and $\bar n=G^{-1}\sum_i n_i$. The raw policy gradient is
\[
\widehat g_{\rm tok}=\frac{\sum_iA_iS_i}{\sum_i n_i}=\frac{\widehat g}{\bar n},
\]
with zero gradient for an empty selected-token group. Conditional on the responses, the expected gradient over random rewards is zero. Writing $\bar S=G^{-1}\sum_iS_i$, the advantage cross moments give
\[
\mathbb E\!\left[\left(\sum_iA_iS_i\right)\left(\sum_iA_iS_i\right)^\top\middle|Y_{1:G}\right]
=\frac{a_GG}{G-1}\sum_i(S_i-\bar S)(S_i-\bar S)^\top.
\]
Dividing by the squared token count and averaging over responses yields
\begin{equation}
\operatorname{Cov}(\widehat g_{\rm tok}\mid\theta,x)
=\frac{a_G}{G(G-1)}\mathbb E\!\left[
\frac{\sum_i(S_i-\bar S)(S_i-\bar S)^\top}{\bar n^2}\middle|\theta,x\right],
\label{eq:tokcov}
\end{equation}
assuming finite moments and setting the integrand to zero for empty groups.

When every response has fixed length $n$ and the scores are iid, this reduces to $a_G\operatorname{Cov}(S)/(Gn^2)$. With random lengths, the denominator remains inside the expectation because length and score can depend on one another. Longer responses alone therefore do not determine the change in gradient covariance. Estimating it requires the responses used for training, rather than lengths from a separate evaluation.

\subsubsection{LoRA and the optimizer}
\label{app:lora}

Training changes the LoRA coordinates while keeping the base weights fixed. For one projection, write $\phi=(A,B)$ and $W(\phi)=W_0+cBA$. The map is nonlinear, with scores $S_\phi=D_\phi W^\top S_W$ and Fisher matrix $F_\phi=D_\phi W^\top F_WD_\phi W$. The raw-gradient results apply in adapter coordinates, whose relation to full-weight changes depends on the current adapter state.

AdamW and gradient-norm clipping further transform these gradients. A fixed linear preconditioner $H$ would send covariance $C$ to $HCH^\top$, but Adam's momentum and normalization depend on the gradient history and current draw. The zero-mean raw-gradient result therefore does not automatically give a zero-mean optimizer step. Weight decay and the reference term add further contributions.

\subsubsection{Clipping in the observed runs}
\label{app:clip}

PPO clipping is enabled with $\varepsilon_{\rm clip}=0.2$. The recorded \texttt{clip\_fraction} measures the fraction of selected tokens whose probability ratio lies outside [0.8, 1.2]. Table~\ref{tab:current-clipping} reports how often this diagnostic is nonzero. Across the three newer experiment groups, 1,057 of 434,000 steps are flagged, about 0.24\%. Five input-control steps have missing diagnostics. The earlier 55-run study has no flagged steps.

\begin{table}[htbp]
\centering\small
\caption{Clipping diagnostics across experiments. A flagged step contains at least one selected token with a probability ratio outside [0.8, 1.2]. The count does not measure the resulting gradient change.}
\label{tab:current-clipping}
\begin{tabular}{lrrr}
\toprule
Batch & Runs & Flagged steps / all steps & Runs with a flagged step\\
\midrule
Completed checkpoint scan & 636 & 519 / 318,000 & 155\\
Expanded input controls & 152 & 492 / 76,000 & 112\\
SmolLM2 / OLMo-3 checkpoints & 80 & 46 / 40,000 & 21\\
Earlier control study & 55 & 0 / 27,500 & 0\\
\bottomrule
\end{tabular}
\end{table}

For identical old and new scoring operations at unchanged parameters,
\[
\rho_{it}=\exp\!\left(\log\pi_{\rm new}(y_{it}\mid x,y_{i,<t})-
\log\pi_{\rm old}(y_{it}\mid x,y_{i,<t})\right)=1,
\]
so the clipped and unclipped objectives have the same gradient. One PPO epoch alone is insufficient if a rollout batch is divided into several optimizer minibatches. Our scan updates once per complete group, but computes old probabilities in evaluation mode and new probabilities in training mode. Zero LoRA dropout does not by itself establish identical numerical scoring, and the newer diagnostics do not support exact ratio equality throughout those runs.

The diagnostic also ignores the advantage sign, which determines whether an out-of-range ratio changes the surrogate gradient. It therefore measures neither the number of gradient-altering clipping events nor their magnitude. The earlier runs show that gains can occur with nonbinding clipping, consistent with the unclipped successes of \citet[][Appendix B.2]{shaoSpuriousRewardsRethinking2026}. The newer runs require the more limited description above.

\subsubsection{The reference term}
\label{app:reference}

The scan computes reference log-probabilities with the adapter disabled and gradients off. For selected response tokens $\mathcal M$, it adds the following term to the minimized loss:
\begin{equation}
\widehat L_{\rm ref}(\theta)=\frac{\beta}{N_{\rm tok}}\sum_{(i,j)\in\mathcal M}
\left[\log\pi_\theta(y_j^{(i)}\mid h_j^{(i)})-
\operatorname{stopgrad}\log\pi_{\rm ref}(y_j^{(i)}\mid h_j^{(i)})\right],
\label{eq:implemented-ref}
\end{equation}
where $N_{\rm tok}=|\mathcal M|$. Holding the sampled tokens and mask fixed gives
\[
\nabla_\theta\widehat L_{\rm ref}
=\frac{\beta}{N_{\rm tok}}\sum_{(i,j)\in\mathcal M}\nabla_\theta\log\pi_\theta(y_j^{(i)}\mid h_j^{(i)}).
\]
The reference values are constants in this gradient; changing them alone does not change the update.

This term differs from the gradient of a forward KL, which also accounts for the policy-dependent sampling distribution. We therefore do not interpret $\beta$ as establishing a force toward the reference policy. An expected score can cancel under exact policy sampling with suitable fixed-length averaging, but that cancellation does not transfer automatically to nucleus sampling, masks, and random-length normalization. The experiments retain $\beta=0$ for GT and 0.01 for Random. The auxiliary term is verifier-independent, so it is compatible with Theorem~\ref{thm:t1}, although the complete update need not share the random-reward gradient's zero mean.

\subsection{Statistical Analysis}
\label{app:stat-methods}

Training seeds are the units of replication. Comparisons pair shared seed IDs, keeping the earlier seed-dependent evaluation splits separate from the fixed-split scan. For P1/P2 comparisons, we first average over checkpoints within each phase and seed, then compare the seed-level averages. This accounts for seed reuse and limits the interpretation to the checkpoint grid studied.

For paired score differences $d_s$, we use a two-sided permutation test with statistic $|\sum_s d_s|$, enumerating sign assignments. It requires exchangeability within pairs, or independent paired differences symmetric about zero; equality of means alone is insufficient. Small samples are enumerated directly, and 32-seed comparisons use exact subset-sum counting on the 200-question score grid. Binary comparisons use exact McNemar tests on discordant pairs. Within a condition, a two-sided sign test instead asks whether nonzero changes are equally likely to be positive and negative. Ties remain in the reported counts but are omitted from that test.

Success-rate intervals are marginal, two-sided 95\% Clopper--Pearson intervals. They condition on the checkpoint, evaluation set, and protocol. With zero successes in eight runs, the upper endpoint is $1-0.025^{1/8}\approx0.369$; eight failures leave considerable uncertainty about the population rate. Pooling outcomes across checkpoints does not turn them into draws from one common success probability, and the 200 evaluation questions do not provide 200 independent training seeds.

We apply Holm correction within seven families: D/E comparisons (15 tests), other OLMo-2 input pairs (18), developmental contrasts (12), additional-model checkpoint and protocol contrasts (32), earlier controls (10), within-condition directional tests (150), and scan GT/Random comparisons (53). We also report a correction over all 290 evaluable tests. These analyses were chosen after inspecting results; correction accounts for the listed tests, not all prior protocol and hypothesis choices. The shared-rollout experiment has a separate family of six tests specified before its formal runs (Appendix~\ref{app:shared-rollouts}).

\section{Additional Results}
\label{app:results}

The checkpoint scans, input controls, and shared-rollout comparison comprise 932 distinct training runs. We report their individual outcomes below, followed by the earlier 55-run group-size and steering study and the statistical comparisons. Reused runs appear in the relevant comparisons but enter this total only once.

\subsection{OLMo-2 Checkpoint Development}
\label{app:scan-results}
\label{tab:scan}

Tables~\ref{tab:gsm-gt-full} and~\ref{tab:gsm-random-full} give all 444 GSM terminal scores. None of the 200 P1 Random runs ends at or above 50\%, compared with 15 of 96 P2 runs. Within P2, however, 35 runs reach this threshold at a scheduled evaluation. Mid-training therefore changes the range of outcomes observed under Random, while many improvements remain transient. The maxima also fluctuate across P2 checkpoints.

Current greedy accuracy does not fully describe this development. The 462B and 839B checkpoints both score 3.5\% before training, yet their best GT outcomes are 8.5\% and 55.0\%. The later checkpoint is a more useful starting point for this training procedure even though the frozen score is unchanged. We take this as evidence that changes relevant to post-training can precede their expression in the model's greedy answers. This comparison matches the greedy scores, rather than the full initial response distributions.

\paragraph{Numerical accuracy and answer format.}
All 15 Random runs ending above the 50\% lenient threshold fall below it under strict extraction (Table~\ref{tab:scan-format}). Strict extraction requires the requested answer-line structure as well as a correct number. The discrepancy shows that output conventions contribute to the measured behavior. Lenient gains demonstrate more correct extracted answers, but do not by themselves give a format-independent measure of mathematical capability.

\begin{table}[htbp]
\centering\small
\caption{Answer-format diagnostics for all GSM Random runs ending at or above 50\% lenient accuracy. Strict accuracy requires a standalone answer line; format rate records the presence of an answer marker. Values are percentages.}
\label{tab:scan-format}
\begin{tabular}{lrrrr}
\toprule
Checkpoint & Seed & Lenient & Strict & Format rate\\
\midrule
P2 +5B & 31003 & 67.5 & 0.0 & 100.0 \\
P2 +5B & 31004 & 59.0 & 0.0 & 0.0 \\
P2 +5B & 31006 & 77.5 & 0.0 & 0.0 \\
P2 +5B & 31007 & 77.0 & 0.0 & 50.5 \\
P2 +9B & 31004 & 78.0 & 43.0 & 54.0 \\
P2 +9B & 31008 & 55.0 & 4.5 & 9.5 \\
P2 +21B & 31001 & 87.0 & 0.0 & 0.0 \\
P2 +21B & 31002 & 89.5 & 0.0 & 98.0 \\
P2 +21B & 31008 & 71.0 & 0.0 & 0.0 \\
P2 +30B & 31003 & 51.5 & 1.0 & 45.0 \\
P2 +38B & 31007 & 82.0 & 0.0 & 41.5 \\
P2 +42B & 31001 & 60.5 & 0.0 & 0.0 \\
P2 +42B & 31002 & 81.5 & 0.0 & 0.0 \\
P2 +42B & 31006 & 71.0 & 0.0 & 40.0 \\
P2 +47B & 31008 & 67.5 & 2.5 & 62.5 \\
\bottomrule
\end{tabular}
\end{table}

\begin{table}[p]
\centering\small
\setlength{\tabcolsep}{7pt}
\caption{Synthetic GSM GT terminal accuracy at step 500 (\%). Seed headings omit 3100; Max is the row maximum. Dashes mark unrun configurations. Base follows the frozen-score convention in Appendix~\ref{app:scan-readout}.}
\label{tab:gsm-gt-full}
\begin{tabular}{lrrrrrrr}
\toprule
Checkpoint & Base & 1 & 2 & 3 & 4 & Max & $n$\\
\midrule
P1 5B & 0.0 & 0.0 & 0.0 & 0.0 & 0.5 & \textbf{0.5} & 4 \\
P1 34B & 0.5 & 2.0 & 0.5 & 0.0 & 1.0 & \textbf{2.0} & 4 \\
P1 462B & 3.5 & 4.5 & 3.5 & 4.5 & 8.5 & \textbf{8.5} & 4 \\
P1 839B & 3.5 & 55.0 & 33.5 & 16.5 & 19.5 & \textbf{55.0} & 4 \\
P1 1,259B & 2.0 & 66.5 & 72.5 & 1.5 & 57.0 & \textbf{72.5} & 4 \\
P1 1,469B & 2.5 & 18.5 & 18.0 & 12.0 & 19.5 & \textbf{19.5} & 4 \\
P1 1,678B & 6.5 & 60.5 & 52.0 & 8.5 & 3.5 & \textbf{60.5} & 4 \\
P1 1,888B & 6.0 & 17.5 & 63.5 & 29.0 & 17.5 & \textbf{63.5} & 4 \\
P1 2,098B & 1.0 & 79.5 & 79.5 & 74.0 & 19.0 & \textbf{79.5} & 4 \\
P1 2,307B & 6.5 & 76.5 & 73.5 & 78.0 & 74.5 & \textbf{78.0} & 4 \\
P1 2,517B & 9.0 & 68.5 & 72.0 & 70.0 & 50.5 & \textbf{72.0} & 4 \\
P1 2,727B & 3.0 & 79.5 & 78.0 & 63.5 & 84.5 & \textbf{84.5} & 4 \\
P1 2,937B & 2.5 & 77.5 & 84.0 & 86.0 & 73.5 & \textbf{86.0} & 4 \\
P1 3,020B & 6.5 & 9.5 & 19.0 & 79.0 & 18.0 & \textbf{79.0} & 4 \\
P1 3,272B & 2.5 & 86.5 & 72.5 & 19.0 & 18.0 & \textbf{86.5} & 4 \\
P1 3,440B & 5.0 & 69.5 & 20.0 & 17.0 & 19.5 & \textbf{69.5} & 4 \\
P1 3,532B & 4.0 & 89.0 & 89.5 & 77.0 & 87.5 & \textbf{89.5} & 4 \\
P1 3,608B & 5.0 & 89.5 & 80.5 & 77.5 & 81.0 & \textbf{89.5} & 4 \\
P1 3,691B & 1.5 & 82.5 & 88.5 & 86.5 & 89.0 & \textbf{89.0} & 4 \\
P1 3,775B & 4.0 & 71.0 & 89.0 & 86.5 & 83.5 & \textbf{89.0} & 4 \\
P1 3,817B & 4.0 & 86.0 & 86.5 & 80.5 & 89.0 & \textbf{89.0} & 4 \\
P1 3,859B & 2.5 & 88.0 & 75.0 & 85.0 & 8.0 & \textbf{88.0} & 4 \\
P1 3,880B & 3.5 & 90.0 & 87.0 & 79.5 & 79.0 & \textbf{90.0} & 4 \\
P1 3,893B & 1.5 & 84.5 & 83.5 & 89.0 & 83.0 & \textbf{89.0} & 4 \\
P1 3,896B & 2.5 & 90.5 & 61.5 & 87.0 & 89.0 & \textbf{90.5} & 4 \\
\midrule
P2 +5B & 13.5 & 98.5 & 99.5 & 100.0 & 99.5 & \textbf{100.0} & 4 \\
P2 +9B & 19.0 & 22.5 & 25.5 & 90.5 & 26.5 & \textbf{90.5} & 4 \\
P2 +13B & 19.0 & 100.0 & 99.0 & 98.5 & 100.0 & \textbf{100.0} & 4 \\
P2 +17B & 19.5 & 100.0 & 96.0 & 100.0 & 99.5 & \textbf{100.0} & 4 \\
P2 +21B & 26.0 & 100.0 & 98.0 & 100.0 & 100.0 & \textbf{100.0} & 4 \\
P2 +26B & 23.5 & 100.0 & 99.5 & 100.0 & 100.0 & \textbf{100.0} & 4 \\
P2 +30B & 15.5 & 99.0 & 100.0 & 99.5 & 100.0 & \textbf{100.0} & 4 \\
P2 +34B & 14.5 & 98.5 & 100.0 & 99.5 & 98.0 & \textbf{100.0} & 4 \\
P2 +38B & 21.0 & 100.0 & 98.5 & 99.5 & 99.0 & \textbf{100.0} & 4 \\
P2 +42B & 11.5 & 95.5 & 100.0 & 100.0 & 100.0 & \textbf{100.0} & 4 \\
P2 +47B & 18.5 & 99.5 & 99.5 & 100.0 & 100.0 & \textbf{100.0} & 4 \\
P2 +50B & 20.5 & 99.0 & 99.5 & 99.0 & 100.0 & \textbf{100.0} & 4 \\
\bottomrule
\end{tabular}
\end{table}

\begin{table}[p]
\centering\small
\setlength{\tabcolsep}{4pt}
\caption{Synthetic GSM Random terminal accuracy at step 500 (\%). Seed headings omit 3100; Max is the row maximum. Dashes mark unrun configurations.}
\label{tab:gsm-random-full}
\begin{tabular}{lrrrrrrrrr}
\toprule
Checkpoint & 1 & 2 & 3 & 4 & 5 & 6 & 7 & 8 & Max\\
\midrule
P1 5B & 0.0 & 0.0 & 0.0 & 0.0 & 0.5 & 0.5 & 0.5 & 0.5 & \textbf{0.5} \\
P1 34B & 0.5 & 2.0 & 1.0 & 1.5 & 1.0 & 0.0 & 0.5 & 1.5 & \textbf{2.0} \\
P1 462B & 4.0 & 2.5 & 4.0 & 3.5 & 4.5 & 0.0 & 2.5 & 3.0 & \textbf{4.5} \\
P1 839B & 1.5 & 2.5 & 1.5 & 3.0 & 4.0 & 4.0 & 4.0 & 3.5 & \textbf{4.0} \\
P1 1,259B & 3.0 & 2.5 & 2.5 & 1.0 & 2.0 & 2.5 & 2.5 & 3.5 & \textbf{3.5} \\
P1 1,469B & 2.0 & 1.0 & 1.0 & 1.5 & 2.0 & 2.5 & 1.0 & 0.0 & \textbf{2.5} \\
P1 1,678B & 8.0 & 6.5 & 5.5 & 6.0 & 7.0 & 9.0 & 4.5 & 3.5 & \textbf{9.0} \\
P1 1,888B & 6.0 & 8.0 & 5.0 & 3.0 & 6.0 & 6.0 & 11.0 & 10.0 & \textbf{11.0} \\
P1 2,098B & 0.5 & 2.0 & 0.5 & 2.0 & 2.0 & 1.0 & 1.5 & 1.5 & \textbf{2.0} \\
P1 2,307B & 2.5 & 3.5 & 3.0 & 3.0 & 4.5 & 9.0 & 9.5 & 1.5 & \textbf{9.5} \\
P1 2,517B & 2.0 & 9.0 & 15.0 & 1.5 & 10.5 & 16.5 & 9.5 & 10.5 & \textbf{16.5} \\
P1 2,727B & 5.0 & 9.0 & 0.0 & 4.0 & 3.5 & 2.5 & 8.0 & 10.0 & \textbf{10.0} \\
P1 2,937B & 1.5 & 0.5 & 2.0 & 2.0 & 3.5 & 2.0 & 8.0 & 3.0 & \textbf{8.0} \\
P1 3,020B & 7.5 & 9.5 & 7.5 & 9.5 & 0.5 & 3.0 & 10.0 & 5.5 & \textbf{10.0} \\
P1 3,272B & 5.5 & 2.5 & 2.0 & 23.5 & 2.5 & 1.5 & 3.5 & 10.0 & \textbf{23.5} \\
P1 3,440B & 12.0 & 5.5 & 0.0 & 11.0 & 4.0 & 7.0 & 1.0 & 6.0 & \textbf{12.0} \\
P1 3,532B & 2.0 & 10.0 & 2.0 & 7.0 & 12.5 & 4.5 & 8.0 & 6.5 & \textbf{12.5} \\
P1 3,608B & 13.5 & 8.0 & 3.5 & 2.0 & 15.0 & 1.0 & 14.0 & 1.5 & \textbf{15.0} \\
P1 3,691B & 11.5 & 1.0 & 1.5 & 1.5 & 6.0 & 9.5 & 8.0 & 2.0 & \textbf{11.5} \\
P1 3,775B & 2.0 & 3.0 & 7.0 & 16.5 & 0.0 & 0.5 & 6.5 & 14.5 & \textbf{16.5} \\
P1 3,817B & 9.0 & 1.0 & 2.0 & 6.5 & 1.5 & 2.5 & 10.0 & 14.5 & \textbf{14.5} \\
P1 3,859B & 15.0 & 0.0 & 2.0 & 2.5 & 27.5 & 5.0 & 0.0 & 0.0 & \textbf{27.5} \\
P1 3,880B & 11.0 & 3.5 & 1.5 & 3.5 & 10.0 & 0.5 & 6.0 & 1.0 & \textbf{11.0} \\
P1 3,893B & 7.0 & 17.0 & 3.5 & 2.0 & 3.5 & 1.5 & 4.5 & 8.0 & \textbf{17.0} \\
P1 3,896B & 7.0 & 2.0 & 13.0 & 11.5 & 4.5 & 9.0 & 5.5 & 2.0 & \textbf{13.0} \\
\midrule
P2 +5B & 17.5 & 5.0 & 67.5 & 59.0 & 2.5 & 77.5 & 77.0 & 16.0 & \textbf{77.5} \\
P2 +9B & 0.0 & 12.5 & 15.5 & 78.0 & 0.0 & 20.0 & 0.0 & 55.0 & \textbf{78.0} \\
P2 +13B & 20.0 & 13.0 & 18.5 & 2.5 & 0.5 & 0.0 & 3.5 & 0.0 & \textbf{20.0} \\
P2 +17B & 3.0 & 28.5 & 12.5 & 12.0 & 3.0 & 5.0 & 2.0 & 24.0 & \textbf{28.5} \\
P2 +21B & 87.0 & 89.5 & 20.5 & 3.0 & 19.0 & 11.0 & 13.0 & 71.0 & \textbf{89.5} \\
P2 +26B & 5.5 & 20.0 & 32.5 & 30.5 & 5.5 & 13.5 & 0.0 & 10.5 & \textbf{32.5} \\
P2 +30B & 0.0 & 29.5 & 51.5 & 15.5 & 4.5 & 12.5 & 1.5 & 17.0 & \textbf{51.5} \\
P2 +34B & 30.0 & 27.5 & 11.0 & 40.5 & 10.0 & 20.0 & 22.5 & 30.0 & \textbf{40.5} \\
P2 +38B & 14.0 & 21.5 & 1.0 & 0.0 & 13.5 & 40.0 & 82.0 & 8.0 & \textbf{82.0} \\
P2 +42B & 60.5 & 81.5 & 14.0 & 0.5 & 4.0 & 71.0 & 21.0 & 3.0 & \textbf{81.5} \\
P2 +47B & 28.0 & 48.5 & 0.0 & 0.0 & 6.0 & 12.0 & 13.5 & 67.5 & \textbf{67.5} \\
P2 +50B & 23.5 & 7.5 & 15.0 & 6.0 & 0.0 & 21.5 & 16.0 & 21.0 & \textbf{23.5} \\
\bottomrule
\end{tabular}
\end{table}

\clearpage
\subsection{Development across Tasks: Code Generation and Compositional Lookup}
\label{experiment}
\label{app:beyond-arithmetic}

\begin{figure}[!t]
\centering
\begin{subfigure}[b]{\linewidth}
\centering
\includegraphics[width=\linewidth]{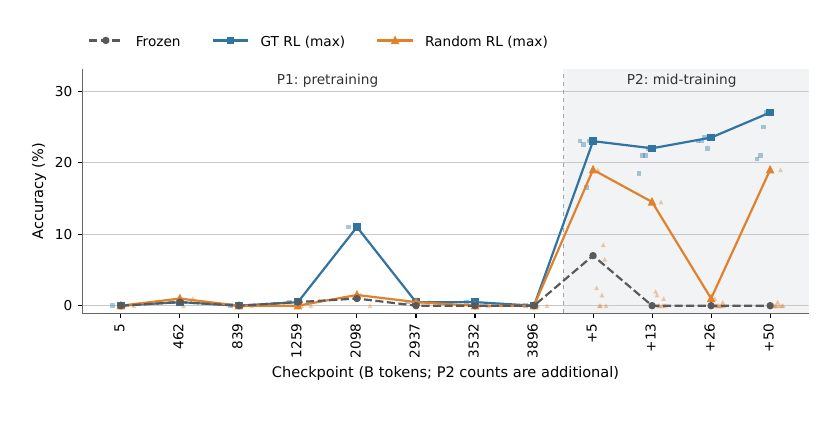}
\caption{Code generation (MBPP).}
\label{fig:code-development}
\end{subfigure}

\begin{subfigure}[b]{\linewidth}
\centering
\includegraphics[width=\linewidth]{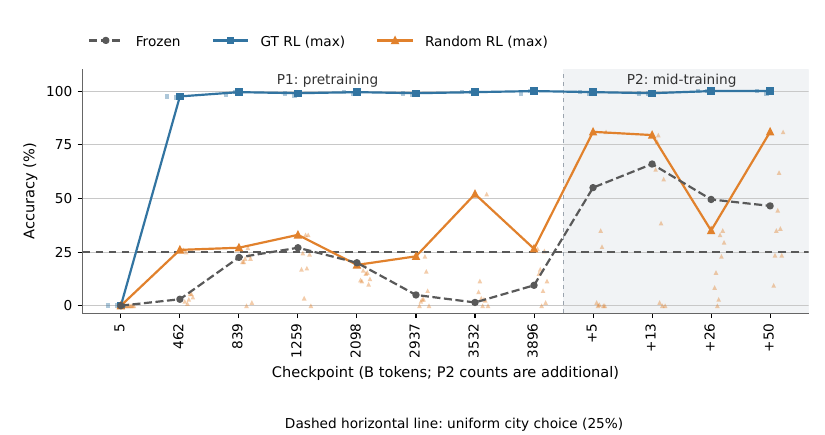}
\caption{One- and two-hop lookup.}
\label{fig:lookup-development}
\end{subfigure}
\caption{Frozen and maximum terminal accuracy after 500 updates on code and lookup. Faint points show individual seeds. Lookup pools 104 one-hop and 96 two-hop items, with uniform choice among four cities shown at 25\%. Panel scales differ. Checkpoints are equally spaced in training order; P2 tokens are additional to P1. Table~\ref{tab:scan-protocol} gives seed counts.}
\label{fig:code-lookup}
\end{figure}

The point at which a checkpoint becomes responsive to further training depends on the task. Figure~\ref{fig:code-lookup} compares code generation and invented-fact lookup at 12 OLMo-2 checkpoints. Lookup already supports successful GT training at a point where arithmetic remains difficult to improve, while consistently higher code scores appear later. These differences make the three-stage description a task-specific account of development.

The tasks and scoring rules are given in Appendix~\ref{app:tasks}. Code has one GT and two Random seeds per P1 checkpoint, increasing to four and eight in P2; lookup has two GT and eight Random seeds throughout. The lookup score pools 104 one-hop and 96 two-hop questions. Tables~\ref{tab:code-gt-full}--\ref{tab:lookup-random-full} give all terminal scores. MBPP uses an existing benchmark and is not an answer-contamination control.

\paragraph{Code generation becomes more responsive in mid-training.}
Most P1 code runs remain near zero. Random never exceeds 1.5\%, and GT remains at or below 0.5\% except at 2,098B, where it reaches 11.0\%. All 16 P2 GT runs instead score at least 16.5\%, with checkpoint maxima between 22.0\% and 27.0\%. The contrast persists across the P2 GT runs, rather than arising solely from taking a maximum over more seeds.

Random produces gains less consistently. Its P2 maxima are 19.0\%, 14.5\%, 1.0\%, and 19.0\% at +5B, +13B, +26B, and +50B. At +13B and +50B, frozen greedy accuracy is zero, yet both GT and Random reach test-passing performance. A model that fails under greedy decoding can therefore still improve through adaptation. The weak +26B result and failed seeds also show that this response does not increase monotonically with training tokens.

\paragraph{Lookup supports correctness-guided training earlier.}
At 462B, frozen lookup accuracy is 3.0\%, but the two GT runs reach 97.0\% and 97.5\%. The latter answers 94 of 96 two-hop questions correctly, so its high score is not due only to the easier subset. At the same checkpoint, the best GSM GT score is 8.5\%. The model has become able to learn to use the relations in the prompt well before it responds similarly on arithmetic.

Random remains weaker at this point, reaching at most 26.0\%, close to the uniform-city reference. Most subsequent P1 maxima also stay near this level, with an exception at 3,532B: one run reaches 52.0\% overall and 43.8\% on two-hop questions. P2 maxima rise to 81.0\%, 79.5\%, and 81.0\% at +5B, +13B, and +50B, while +26B ends at 35.0\%. These results again separate becoming responsive to GT from becoming responsive to Random, without placing a common boundary between the phases across tasks.

\begin{table}[p]
\centering\small
\setlength{\tabcolsep}{7pt}
\caption{Code GT terminal accuracy at step 500 (\%). Seed headings omit 3100; Max is the row maximum. Dashes mark unrun configurations. Base follows the frozen-score convention in Appendix~\ref{app:scan-readout}.}
\label{tab:code-gt-full}
\begin{tabular}{lrrrrrrr}
\toprule
Checkpoint & Base & 1 & 2 & 3 & 4 & Max & $n$\\
\midrule
P1 5B & 0.0 & 0.0 & --- & --- & --- & \textbf{0.0} & 1 \\
P1 462B & 0.5 & 0.5 & --- & --- & --- & \textbf{0.5} & 1 \\
P1 839B & 0.0 & 0.0 & --- & --- & --- & \textbf{0.0} & 1 \\
P1 1,259B & 0.5 & 0.5 & --- & --- & --- & \textbf{0.5} & 1 \\
P1 2,098B & 1.0 & 11.0 & --- & --- & --- & \textbf{11.0} & 1 \\
P1 2,937B & 0.0 & 0.5 & --- & --- & --- & \textbf{0.5} & 1 \\
P1 3,532B & 0.0 & 0.5 & --- & --- & --- & \textbf{0.5} & 1 \\
P1 3,896B & 0.0 & 0.0 & --- & --- & --- & \textbf{0.0} & 1 \\
\midrule
P2 +5B & 7.0 & 23.0 & 22.5 & 16.5 & 23.0 & \textbf{23.0} & 4 \\
P2 +13B & 0.0 & 18.5 & 21.0 & 21.0 & 22.0 & \textbf{22.0} & 4 \\
P2 +26B & 0.0 & 23.0 & 23.0 & 23.5 & 22.0 & \textbf{23.5} & 4 \\
P2 +50B & 0.0 & 20.5 & 21.0 & 25.0 & 27.0 & \textbf{27.0} & 4 \\
\bottomrule
\end{tabular}
\end{table}

\begin{table}[p]
\centering\small
\setlength{\tabcolsep}{4pt}
\caption{Code Random terminal accuracy at step 500 (\%). Seed headings omit 3100; Max is the row maximum. Dashes mark unrun configurations.}
\label{tab:code-random-full}
\begin{tabular}{lrrrrrrrrr}
\toprule
Checkpoint & 1 & 2 & 3 & 4 & 5 & 6 & 7 & 8 & Max\\
\midrule
P1 5B & 0.0 & 0.0 & --- & --- & --- & --- & --- & --- & \textbf{0.0} \\
P1 462B & 0.0 & 1.0 & --- & --- & --- & --- & --- & --- & \textbf{1.0} \\
P1 839B & 0.0 & 0.0 & --- & --- & --- & --- & --- & --- & \textbf{0.0} \\
P1 1,259B & 0.0 & 0.0 & --- & --- & --- & --- & --- & --- & \textbf{0.0} \\
P1 2,098B & 1.5 & 0.0 & --- & --- & --- & --- & --- & --- & \textbf{1.5} \\
P1 2,937B & 0.5 & 0.5 & --- & --- & --- & --- & --- & --- & \textbf{0.5} \\
P1 3,532B & 0.0 & 0.0 & --- & --- & --- & --- & --- & --- & \textbf{0.0} \\
P1 3,896B & 0.0 & 0.0 & --- & --- & --- & --- & --- & --- & \textbf{0.0} \\
\midrule
P2 +5B & 2.5 & 19.0 & 0.0 & 0.0 & 1.5 & 8.5 & 6.5 & 0.0 & \textbf{19.0} \\
P2 +13B & 2.0 & 1.5 & 0.0 & 0.0 & 14.5 & 0.0 & 1.0 & 0.0 & \textbf{14.5} \\
P2 +26B & 1.0 & 0.0 & 0.0 & 0.0 & 0.0 & 0.0 & 0.5 & 0.0 & \textbf{1.0} \\
P2 +50B & 0.0 & 0.0 & 0.0 & 0.5 & 0.0 & 19.0 & 0.0 & 0.0 & \textbf{19.0} \\
\bottomrule
\end{tabular}
\end{table}

\begin{table}[p]
\centering\small
\setlength{\tabcolsep}{7pt}
\caption{Compositional lookup GT terminal accuracy at step 500 (\%). Seed headings omit 3100; Max is the row maximum. Dashes mark unrun configurations. Base follows the frozen-score convention in Appendix~\ref{app:scan-readout}.}
\label{tab:lookup-gt-full}
\begin{tabular}{lrrrrrrr}
\toprule
Checkpoint & Base & 1 & 2 & 3 & 4 & Max & $n$\\
\midrule
P1 5B & 0.0 & 0.0 & 0.0 & --- & --- & \textbf{0.0} & 2 \\
P1 462B & 3.0 & 97.5 & 97.0 & --- & --- & \textbf{97.5} & 2 \\
P1 839B & 22.5 & 98.5 & 99.5 & --- & --- & \textbf{99.5} & 2 \\
P1 1,259B & 27.0 & 99.0 & 98.0 & --- & --- & \textbf{99.0} & 2 \\
P1 2,098B & 20.0 & 99.5 & 99.0 & --- & --- & \textbf{99.5} & 2 \\
P1 2,937B & 5.0 & 99.0 & 98.5 & --- & --- & \textbf{99.0} & 2 \\
P1 3,532B & 1.5 & 99.5 & 99.5 & --- & --- & \textbf{99.5} & 2 \\
P1 3,896B & 9.5 & 99.0 & 100.0 & --- & --- & \textbf{100.0} & 2 \\
\midrule
P2 +5B & 55.0 & 99.5 & 99.0 & --- & --- & \textbf{99.5} & 2 \\
P2 +13B & 66.0 & 99.0 & 99.0 & --- & --- & \textbf{99.0} & 2 \\
P2 +26B & 49.5 & 100.0 & 100.0 & --- & --- & \textbf{100.0} & 2 \\
P2 +50B & 46.5 & 100.0 & 99.0 & --- & --- & \textbf{100.0} & 2 \\
\bottomrule
\end{tabular}
\end{table}

\begin{table}[p]
\centering\small
\setlength{\tabcolsep}{4pt}
\caption{Compositional lookup Random terminal accuracy at step 500 (\%). Seed headings omit 3100; Max is the row maximum. Dashes mark unrun configurations.}
\label{tab:lookup-random-full}
\begin{tabular}{lrrrrrrrrr}
\toprule
Checkpoint & 1 & 2 & 3 & 4 & 5 & 6 & 7 & 8 & Max\\
\midrule
P1 5B & 0.0 & 0.0 & 0.0 & 0.0 & 0.0 & 0.0 & 0.0 & 0.0 & \textbf{0.0} \\
P1 462B & 26.0 & 2.0 & 25.0 & 1.0 & 3.0 & 5.5 & 5.5 & 4.0 & \textbf{26.0} \\
P1 839B & 20.5 & 20.5 & 22.0 & 0.0 & 27.0 & 23.5 & 22.0 & 1.5 & \textbf{27.0} \\
P1 1,259B & 17.0 & 24.5 & 3.5 & 33.0 & 17.5 & 33.0 & 24.0 & 0.0 & \textbf{33.0} \\
P1 2,098B & 12.0 & 11.5 & 16.5 & 19.0 & 15.0 & 15.5 & 10.0 & 12.5 & \textbf{19.0} \\
P1 2,937B & 0.0 & 2.0 & 3.0 & 3.0 & 23.0 & 16.0 & 7.0 & 0.0 & \textbf{23.0} \\
P1 3,532B & 6.5 & 11.5 & 3.5 & 0.0 & 3.0 & 2.5 & 52.0 & 0.0 & \textbf{52.0} \\
P1 3,896B & 26.5 & 15.0 & 17.0 & 0.0 & 7.0 & 25.5 & 1.5 & 11.5 & \textbf{26.5} \\
\midrule
P2 +5B & 1.5 & 0.0 & 0.5 & 35.0 & 27.5 & 0.0 & 0.0 & 81.0 & \textbf{81.0} \\
P2 +13B & 63.5 & 76.5 & 79.5 & 1.5 & 38.5 & 0.0 & 59.0 & 0.0 & \textbf{79.5} \\
P2 +26B & 8.5 & 15.5 & 0.0 & 3.0 & 33.0 & 23.0 & 35.0 & 29.5 & \textbf{35.0} \\
P2 +50B & 9.5 & 23.5 & 35.0 & 44.5 & 62.0 & 36.0 & 23.5 & 81.0 & \textbf{81.0} \\
\bottomrule
\end{tabular}
\end{table}

\clearpage
\subsection{Checkpoint Development in SmolLM2 and OLMo-3}
\label{app:crossmodel-checkpoints}

\begin{figure}[!htbp]
\centering
\begin{subfigure}[b]{0.485\linewidth}
\centering
\includegraphics[width=\linewidth]{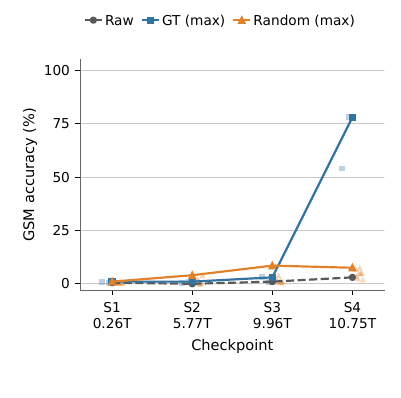}
\caption{\exampleb{SmolLM2-1.7B}.}
\end{subfigure}\hfill
\begin{subfigure}[b]{0.485\linewidth}
\centering
\includegraphics[width=\linewidth]{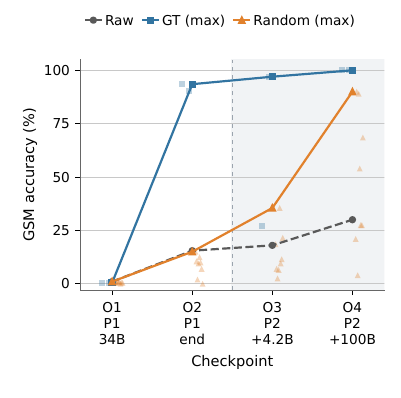}
\caption{\exampleb{OLMo-3-7B}.}
\end{subfigure}
\caption{Checkpoint development in SmolLM2 and OLMo-3. Curves show frozen greedy accuracy and maximum terminal accuracy after 500 updates; faint points show the two GT and eight Random seeds at each checkpoint. All use the same 200 GSM questions. Horizontal spacing follows checkpoint order; shading marks OLMo-3 mid-training.}
\label{fig:crossmodel-checkpoints}
\end{figure}

Figure~\ref{fig:crossmodel-checkpoints} compares frozen scores with the best GT and Random terminal outcomes at eight additional checkpoints. The two forms of training again become effective at different points. The selected checkpoints do not, however, cover all three phases in every model (Tables~\ref{tab:crossmodel-summary} and~\ref{tab:crossmodel-seeds}).

\begin{table}[htbp]
\centering\small
\caption{Additional-model checkpoint results (\%). Maxima use terminal scores, except Random peak, which also includes intermediate evaluations. Gain counts terminal improvements greater than ten points.}
\label{tab:crossmodel-summary}
\begin{tabular}{lrrrrrr}
\toprule
Checkpoint & Raw & GT max & Random max & Random mean & Gain & Random peak\\
\midrule
S1 & 0.5 & 1.0 & 1.0 & 0.56 & 0/8 & 1.0\\
S2 & 0.0 & 1.0 & 4.0 & 1.12 & 0/8 & 4.0\\
S3 & 1.0 & 3.0 & 8.5 & 2.44 & 0/8 & 8.5\\
S4 & 3.0 & 78.0 & 7.5 & 4.69 & 0/8 & 7.5\\
\midrule
O1 & 1.0 & 0.5 & 1.0 & 0.63 & 0/8 & 1.0\\
O2 & 15.5 & 93.5 & 15.0 & 8.31 & 0/8 & 31.5\\
O3 & 18.0 & 97.0 & 35.5 & 14.06 & 1/8 & 55.0\\
O4 & 30.0 & 100.0 & 90.0 & 47.69 & 4/8 & 90.0\\
\bottomrule
\end{tabular}
\end{table}

\paragraph{SmolLM2 becomes receptive within the sampled checkpoints.}
At S1--S3, neither protocol reaches 10\% terminal accuracy. At S4, the two GT runs reach 54\% and 78\%, while all eight Random runs remain between 2\% and 7.5\%. This resembles the receptive phase in OLMo-2: the model has acquired enough relevant knowledge for GT training to become effective, but that knowledge does not yet support comparable gains under Random. No Random run gains more than ten points in this four-checkpoint comparison.

\paragraph{OLMo-3 becomes responsive to Random later in training.}
Both protocols remain near zero at O1. At the P1 endpoint O2, GT reaches 90.5\% and 93.5\%, while Random ends between zero and 15\%. Early in P2, O3 has one Random gain to 35.5\%, though its mean terminal score remains below the starting score. At O4, both GT runs reach 100\% and four of eight Random runs gain more than ten points, with a maximum of 90\%. The ordering resembles OLMo-2, with the response to Random developing over mid-training rather than appearing immediately at its start.

The maxima conceal differences in stability. O2 and O3 reach Random peaks of 31.5\% and 55\%, above their terminal maxima, while the two O3 GT runs end at 27\% and 97\%. A checkpoint can therefore admit successful training without making that outcome reliable across runs. The individual scores in Table~\ref{tab:crossmodel-seeds} show both aspects of its response.

\begin{table}[htbp]
\centering\small
\setlength{\tabcolsep}{3pt}
\caption{All 80 terminal scores in the additional checkpoint comparison (\%). GT columns use seeds 31001--31002 and Random columns use seeds 31001--31008.}
\label{tab:crossmodel-seeds}
\begin{tabular}{lrrrrrrrrrr}
\toprule
& \multicolumn{2}{c}{GT} & \multicolumn{8}{c}{Random}\\
\cmidrule(lr){2-3}\cmidrule(lr){4-11}
ID & 1 & 2 & 1 & 2 & 3 & 4 & 5 & 6 & 7 & 8\\
\midrule
S1 & 1.0 & 0.5 & 0.5 & 0.5 & 0.5 & 0.5 & 0.5 & 1.0 & 0.5 & 0.5\\
S2 & 0.5 & 1.0 & 0.5 & 0.5 & 2.0 & 1.5 & 0.0 & 0.0 & 0.5 & 4.0\\
S3 & 3.0 & 1.0 & 8.5 & 2.0 & 1.0 & 4.0 & 1.0 & 1.5 & 0.5 & 1.0\\
S4 & 54.0 & 78.0 & 7.0 & 3.5 & 2.5 & 5.0 & 7.5 & 5.0 & 5.0 & 2.0\\
\midrule
O1 & 0.5 & 0.5 & 0.5 & 0.5 & 1.0 & 1.0 & 1.0 & 0.0 & 0.5 & 0.5\\
O2 & 93.5 & 90.5 & 15.0 & 10.5 & 2.0 & 9.5 & 12.5 & 10.0 & 7.0 & 0.0\\
O3 & 27.0 & 97.0 & 18.5 & 7.0 & 2.5 & 6.5 & 35.5 & 9.5 & 11.5 & 21.5\\
O4 & 100.0 & 100.0 & 21.0 & 90.0 & 4.0 & 89.0 & 54.0 & 27.5 & 27.5 & 68.5\\
\bottomrule
\end{tabular}
\end{table}

\paragraph{Frozen scores are sensitive to continued generation.}
Some frozen models answer a question and then generate another one before reaching the token limit. The lenient extractor can select a number from that continuation. Table~\ref{tab:crossmodel-scoring} applies a diagnostic that stops at the first blank line, \exampleg{Question:}, or \exampleg{Solve the} restart. For O2, the score changes from 15.5\% to 53\%. Some low frozen scores therefore reflect output conventions as well as problem-solving ability. This post-hoc rule can also truncate a legitimate solution, and complete trained outputs are unavailable for applying it throughout the scan. We retain the original scorer for the before/after comparison rather than replacing only its baseline.

\begin{table}[htbp]
\centering\small
\caption{Frozen accuracy with the primary full-output extractor and the first-block diagnostic (\%). The developmental figures retain the primary extractor before and after training.}
\label{tab:crossmodel-scoring}
\begin{tabular}{lrrrrrrrr}
\toprule
Checkpoint & S1 & S2 & S3 & S4 & O1 & O2 & O3 & O4\\
\midrule
Full output & 0.5 & 0.0 & 1.0 & 3.0 & 1.0 & 15.5 & 18.0 & 30.0\\
First block & 3.0 & 4.0 & 14.0 & 27.5 & 1.0 & 53.0 & 35.0 & 34.0\\
\bottomrule
\end{tabular}
\end{table}

\subsection{The Developmental Pattern under Supervised Fine-Tuning}
\label{app:sft}

Figure~\ref{fig:checkpoint-scan}(b) compares full-answer SFT with \emph{form-only} (number-masked) SFT. Full-answer SFT trains on the worked answer. Form-only SFT masks numbers in both questions and solutions while retaining the wording and solution structure. The latter still supplies information about how to approach the task, although it withholds the numerical content of the answer. The ordering resembles the RL results: checkpoints become responsive to fuller supervision before the restricted form becomes effective. Since SFT uses neither policy-ratio clipping nor group-relative advantages, this parallel suggests a developmental explanation beyond those RL-specific mechanisms.

The SFT comparison is qualitative. We recover 22 checkpoint values from the original figure's vector paths; per-seed SFT results are unavailable. That figure reports best SFT scores, whereas the RL curves use best terminal scores, and the budgets, checkpoint coverage, selection rules, and evaluation details have not been established as matched. The comparison supports examining the order of developmental changes, rather than numerical equivalence between the two training procedures.

\subsection{Training-Input Controls on OLMo-2}
\label{app:randq-results}

\begin{figure}[!htbp]
\centering
\includegraphics[width=\linewidth]{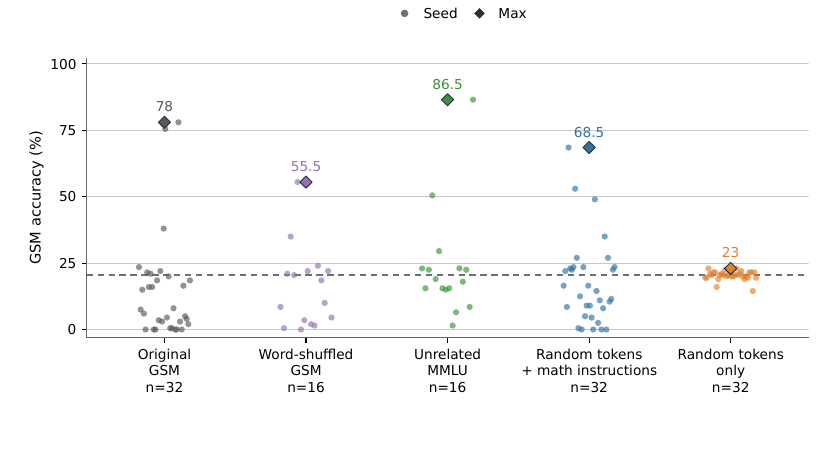}
\caption{Terminal GSM accuracy under the five OLMo-2 input conditions. Dots show every seed; diamonds mark maxima. Original GSM and the two random-token conditions have 32 seeds each, and shuffled GSM and MMLU have 16. The dashed line is the frozen score of 20.5\%.}
\label{fig:randq-terminal}
\end{figure}

Figure~\ref{fig:randq-terminal} and Table~\ref{tab:randq-summary} compare the five input conditions at P2 +50B. Large gains occur with original GSM, shuffled GSM, unrelated MMLU questions, and random tokens inside the math template. Random tokens without the template instead leave terminal scores close to the starting level. Ordinary mathematics questions are therefore unnecessary for the observed gains, while the remaining task cues appear relevant to how training affects the model.

\begin{table}[htbp]
\centering\small
\caption{OLMo-2 input-control outcomes at step 500. Scores are percentages. Gain means improvement greater than ten points from a run\textquotesingle s initial score; collapse means terminal accuracy at or below 5\%.}
\label{tab:randq-summary}
\begin{tabular}{lrrrrrr}
\toprule
Training input & $n$ & Max & Mean & Median & Gain & Collapse\\
\midrule
B: GSM & 32 & 78.0 & 13.98 & 6.75 & 3/32 & 15/32\\
C: shuffled GSM & 16 & 55.5 & 15.56 & 14.25 & 2/16 & 6/16\\
A: MMLU & 16 & 86.5 & 23.28 & 18.50 & 2/16 & 1/16\\
D: tokens + math template & 32 & 68.5 & 17.50 & 13.50 & 4/32 & 8/32\\
E: raw tokens & 32 & 23.0 & 20.36 & 20.50 & 0/32 & 0/32\\
\bottomrule
\end{tabular}
\end{table}

\paragraph{Task-formatted inputs produce a wider range of outcomes.}
With random tokens inside the template, D seed 31004 rises from 19.0\% at step 100 to 68.5\% at step 500; seed 31008 reaches 59.5\% at step 400 and ends at 53.0\%. Two further seeds end at 49.0\% and 35.0\%. Without the template, all 160 scheduled E scores lie between 14.5\% and 23\%. The increased response in D includes deterioration: its terminal mean is 17.50\%, below the 20.5\% frozen score. Figure~\ref{fig:randq-overview} shows all trajectories, and Table~\ref{tab:randq-seeds} gives every terminal result.

\begin{table}[htbp]
\centering\small
\caption{All OLMo-2 input-control terminal scores (\%). B uses original GSM, C shuffled GSM, A MMLU stems, D tokens within the math template, and E tokens without the template. Dashes indicate seeds not run.}
\label{tab:randq-seeds}
\begin{tabular}{rrrrrr}
\toprule
Seed & B & C & A & D & E\\
\midrule
31001 & 23.5 & 8.5 & 23.0 & 16.5 & 19.5\\
31002 & 7.5 & 0.5 & 15.5 & 22.0 & 19.5\\
31003 & 15.0 & 21.0 & 22.5 & 8.5 & 23.0\\
31004 & 6.0 & 35.0 & 50.5 & 68.5 & 21.0\\
31005 & 0.0 & 20.5 & 19.0 & 23.0 & 20.5\\
31006 & 21.5 & 55.5 & 29.5 & 22.5 & 21.5\\
31007 & 16.0 & 0.0 & 15.5 & 23.5 & 21.5\\
31008 & 21.0 & 3.5 & 15.0 & 53.0 & 16.0\\
31009 & 16.0 & 22.0 & 15.5 & 27.0 & 19.0\\
31010 & 0.0 & 2.0 & 1.5 & 0.5 & 20.5\\
31011 & 0.0 & 1.5 & 6.5 & 12.5 & 21.0\\
31012 & 18.5 & 24.0 & 23.0 & 0.0 & 20.5\\
31013 & 3.5 & 18.5 & 18.0 & 23.5 & 22.5\\
31014 & 22.0 & 10.0 & 22.5 & 5.0 & 20.0\\
31015 & 3.0 & 22.0 & 8.5 & 9.0 & 20.5\\
31016 & 38.0 & 4.5 & 86.5 & 16.5 & 21.0\\
31017 & 75.5 & -- & -- & 9.0 & 20.0\\
31018 & 4.5 & -- & -- & 4.5 & 20.0\\
31019 & 20.0 & -- & -- & 0.0 & 20.5\\
31020 & 0.5 & -- & -- & 49.0 & 23.0\\
31021 & 0.5 & -- & -- & 14.5 & 20.5\\
31022 & 8.0 & -- & -- & 2.5 & 21.0\\
31023 & 0.0 & -- & -- & 11.0 & 22.0\\
31024 & 0.0 & -- & -- & 0.0 & 20.0\\
31025 & 78.0 & -- & -- & 8.0 & 19.0\\
31026 & 3.0 & -- & -- & 35.0 & 20.0\\
31027 & 0.0 & -- & -- & 0.0 & 19.5\\
31028 & 16.5 & -- & -- & 27.0 & 21.5\\
31029 & 5.0 & -- & -- & 10.5 & 21.5\\
31030 & 4.0 & -- & -- & 11.5 & 14.5\\
31031 & 2.0 & -- & -- & 22.5 & 21.5\\
31032 & 18.5 & -- & -- & 23.5 & 19.5\\
\bottomrule
\end{tabular}
\end{table}

\begin{figure}[!htbp]
\centering
\begin{subfigure}[b]{0.485\linewidth}
\centering
\includegraphics[width=\linewidth]{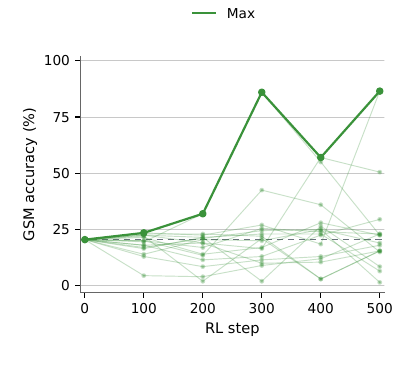}
\caption{Unrelated MMLU questions.}
\end{subfigure}\hfill
\begin{subfigure}[b]{0.485\linewidth}
\centering
\includegraphics[width=\linewidth]{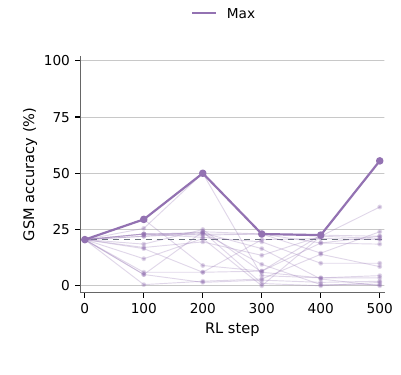}
\caption{Word-shuffled GSM questions.}
\end{subfigure}
\caption{Training trajectories with unrelated MMLU questions and shuffled GSM questions. Both retain the math template. Thin lines show all 16 seeds; thick lines show the pointwise maximum, which can switch seeds. The dashed line marks frozen accuracy.}
\label{fig:randq-ac}
\end{figure}

The MMLU and shuffled-GSM conditions also show gains that can either persist or disappear (Figure~\ref{fig:randq-ac}). MMLU seed 31016 rises from 18.5\% at step 400 to 86.5\% at step 500, while seed 31014 reaches 86.0\% at step 300 and ends at 22.5\%. Shuffled-GSM seed 31006 ends at 55.5\%, whereas seed 31016 reaches 50.0\% at step 200 and ends at 4.5\%. Continued random-reward updates can undo an earlier improvement.

\paragraph{A correct answer need not follow the requested format.}
The highest MMLU-trained run has 86.5\% lenient accuracy but zero strict accuracy and no \exampleg{\#\#\#\#} markers. For example,
\begin{quote}
\textit{Question:} \exampleg{Victor buys 3 pages at \$8 each and pays \$84. How much change does Victor get?}\par\smallskip
\textit{Model output:} Victor pays 3 * 8 = \$24 for the pages. Victor gets 84 - 24 = \$60 in change.
\end{quote}
The response supplies the correct answer without the prescribed marker. The two high-scoring D runs likewise have quite different format rates, 89.0\% and 14.5\%. Numerical correctness and compliance with the answer format need to be examined separately.

The conditions also differ in deterioration. Original GSM has three terminal gains above ten points and 15 collapses to at most 5\% out of 32 runs; D has four gains and eight collapses. MMLU has two gains and one collapse out of 16, and shuffled GSM has two gains and six collapses. These samples do not establish an ordering of input quality. They show that the same random-reward procedure can produce different distributions of outcomes depending on what the model trains on.

The clearest D/E statistical contrast concerns the magnitude of change. Sixteen of 32 D runs change by more than ten points in either direction, compared with none in E (exact McNemar $p=3.05\times10^{-5}$; Holm $p_H=4.27\times10^{-4}$). Positive gains alone, four versus zero, give $p$ = 0.125. The evidence for greater responsiveness is therefore stronger than the evidence for a higher success probability. Since D and E also differ in token construction (Appendix~\ref{app:randq-setup}), this supports a task-cue interpretation without isolating the effect of the instruction alone.

\subsection{Training-Input Controls on Qwen2.5 and Llama}
\label{app:crossmodel-inputs}

\begin{figure}[!htbp]
\centering
\begin{subfigure}[b]{0.485\linewidth}
\centering
\includegraphics[width=\linewidth]{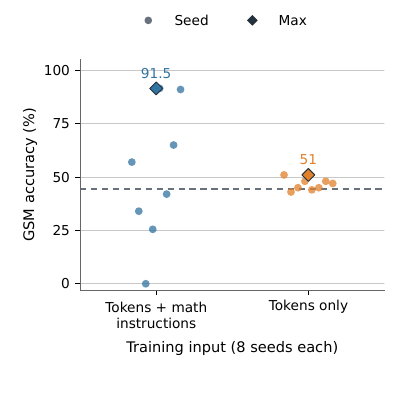}
\caption{\exampleb{Qwen2.5-7B}.}
\end{subfigure}\hfill
\begin{subfigure}[b]{0.485\linewidth}
\centering
\includegraphics[width=\linewidth]{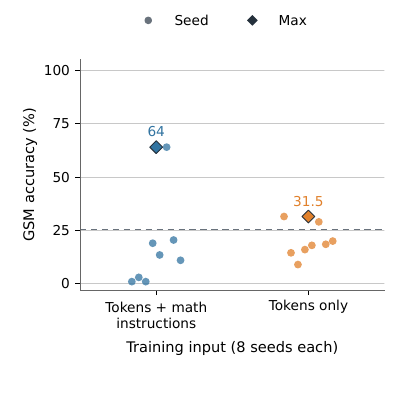}
\caption{\exampleb{Llama-3.1-8B}.}
\end{subfigure}
\caption{Random-token input controls on Qwen2.5 and Llama. Dots show the eight terminal scores in each condition; diamonds mark maxima. Dashed lines show mean pre-update scores, 44.3\% and 25.5\%. Input construction differs in both template and token filtering (Appendix~\ref{app:randq-setup}).}
\label{fig:crossmodel-inputs}
\end{figure}

The input dependence extends to other base models (Figure~\ref{fig:crossmodel-inputs}). With the template, Qwen's terminal scores range from zero to 91.5\%, including four gains above ten points. Without it, scores remain between 43\% and 51\%, with no change above ten points in either direction. The results resemble OLMo-2 in separating large changes under task-formatted inputs from relatively stable behavior without the template.

\begin{table}[htbp]
\centering\small
\caption{Qwen2.5 and Llama input-control results (\%), with eight seeds per condition. D retains the math template; E omits it. Gain means an improvement above ten points; collapse means terminal accuracy at or below 5\%.}
\label{tab:crossmodel-input-summary}
\begin{tabular}{llrrrrr}
\toprule
Model & Input & Max & Mean & SD & Gain & Collapse\\
\midrule
\exampleb{Qwen2.5-7B} & D & 91.5 & 50.75 & 29.78 & 4/8 & 1/8\\
& E & 51.0 & 46.38 & 2.45 & 0/8 & 0/8\\
\exampleb{Llama-3.1-8B} & D & 64.0 & 16.63 & 19.30 & 1/8 & 3/8\\
& E & 31.5 & 19.56 & 6.94 & 0/8 & 0/8\\
\bottomrule
\end{tabular}
\end{table}

Llama has one D gain to 64\%, but three D runs collapse and the mean falls below its starting score. Six of eight D runs lose more than ten points from their scheduled peak to the endpoint. E also changes Llama: two runs decline by more than ten points, though none gains that much. Thus the high-scoring runs coexist with model-specific patterns of deterioration. Tables~\ref{tab:crossmodel-input-summary} and~\ref{tab:crossmodel-input-seeds} show the full outcomes; neither model's D/E comparison of absolute change survives the 15-test correction reported in Appendix~\ref{app:statistics}.

\begin{table}[htbp]
\centering\small
\caption{All Qwen2.5 and Llama input-control terminal scores (\%). D uses tokens within the math template and E uses tokens alone.}
\label{tab:crossmodel-input-seeds}
\begin{tabular}{rrrrr}
\toprule
& \multicolumn{2}{c}{\exampleb{Qwen2.5-7B}} & \multicolumn{2}{c}{\exampleb{Llama-3.1-8B}}\\
\cmidrule(lr){2-3}\cmidrule(lr){4-5}
Seed & D & E & D & E\\
\midrule
31001 & 57.0 & 51.0 & 1.0 & 31.5\\
31002 & 34.0 & 43.0 & 3.0 & 14.5\\
31003 & 0.0 & 45.0 & 1.0 & 9.0\\
31004 & 25.5 & 48.0 & 19.0 & 16.0\\
31005 & 91.5 & 44.0 & 13.5 & 18.0\\
31006 & 42.0 & 45.0 & 64.0 & 29.0\\
31007 & 65.0 & 48.0 & 20.5 & 18.5\\
31008 & 91.0 & 47.0 & 11.0 & 20.0\\
\bottomrule
\end{tabular}
\end{table}

\subsection{Shared-Rollout Cross-Training}
\label{app:shared-rollouts}

This experiment separates the model that produces training responses from the model updated on them. We use the final P1 checkpoint of OLMo-2 and its first released P2 checkpoint, +5B. Each is both a frozen response source and a learner, giving four source--learner combinations. Sixteen replicate blocks, 51001--51016, contain all four combinations, for 64 runs of 500 updates. The main outcomes appear in Table~\ref{tab:shared-rollout-main}.

\paragraph{Shared training material.}
The two checkpoints share the same tokenizer and configuration, so responses can be passed directly as token IDs. We use the scan's 800/200 GSM split and the first 500 questions of the ordered training set, visiting one per step. Each frozen source samples four responses per question with the math template, temperature 1, top-$p$ = 0.95, and a 96-token cap, in batches of ten prompts. No response is filtered, resampled, or sorted by correctness. The 32 source streams contain 64,000 responses.

Within a replicate, all four learners share a $500\times4$ table of independent Bernoulli$(1/2)$ rewards and the same initial LoRA tensors. Within each source condition, the two learners also share all response tokens and loss masks. Advantages use the population standard deviation with $10^{-6}$ stabilization, and constant-reward groups have zero advantage. Correctness is used only for evaluation and diagnostics.

\paragraph{Offline updates.}
For four cached responses $y_i$ to prompt $x$, each learner minimizes
\begin{equation}
\mathcal L(\theta)=-\frac{\sum_{i=1}^{4}\sum_jm_{ij}\widehat A_i
\log\pi_\theta(y_{ij}\mid x,y_{i,<j})}{\sum_{i=1}^{4}\sum_jm_{ij}},
\label{eq:shared-rollout-loss}
\end{equation}
where $m_{ij}$ selects non-padding response tokens and includes EOS. The denominator counts valid tokens across the complete group. There is no importance weighting, policy-ratio clipping, reference penalty, entropy bonus, or SFT term. These are off-policy random-advantage updates; they do not estimate the on-policy GRPO gradient without bias.

The base weights remain frozen. LoRA uses rank 8, scaling 16, and zero dropout on attention query, key, value, and output projections and MLP gate, up, and down projections. AdamW has learning rate $3\times10^{-5}$, $(\beta_1,\beta_2)=(0.9,0.999)$, $\varepsilon=10^{-8}$, zero weight decay, and gradient-norm cap 1. Base computation is BF16; adapter weights and log-probabilities are FP32. The optimizer still steps on zero-advantage groups. Unlike the on-policy Random runs, these runs have neither the reference coefficient 0.01 nor weight decay 0.01, so even matching source and learner does not replicate the on-policy condition.

Greedy evaluation uses the same 200 questions and 96-token cap at steps 0, 100, 200, 300, 400, and 500. P1 starts at 2.5\% under the primary lenient scorer and P2 at 13.5\%. Gains subtract each learner's initial score. The experiment used approximately 17 NVIDIA H200 GPU-hours, including preliminary tests.

\paragraph{Paired comparisons.}
We specified six two-sided tests before the formal runs. For each response source, they compare P2 with P1 on signed gain, the indicator of a gain above ten points, and absolute gain. Signed and absolute differences use sign tests; improved-run indicators use exact McNemar tests. The four-run replicate is the unit of pairing, and Holm correction covers all six tests (Table~\ref{tab:shared-rollout-tests}). The absolute-change tests distinguish how strongly a model responds from whether it improves.

Only absolute change with P1 responses passes this correction. Mean absolute changes are 1.75 points for the P1 learner and 13.56 for P2; with P2 responses, they are 3.63 and 14.28. Table~\ref{tab:shared-rollout-seeds} gives all terminal scores. The improved-run counts in the four cells have marginal 95\% intervals of [0, 20.6]\%, [4.0, 45.6]\%, [1.6, 38.3]\%, and [15.2, 64.6]\%, in source--learner order P1/P1, P1/P2, P2/P1, and P2/P2. Their highest scheduled scores are 33.0\%, 80.0\%, 18.5\%, and 84.0\%, respectively; these include intermediate evaluations.

\begin{table}[htbp]
\centering\small
\begin{tabular}{llcrr}
\toprule
Source & Paired quantity (P2 minus P1) & Counts & $p$ & Holm $p$\\
\midrule
P1 & Gain & 6/9/1 & 0.6072 & 0.8405\\
P1 & Improved & 3/0 & 0.2500 & 0.8405\\
P1 & Absolute gain & 13/2/1 & 0.0074 & 0.0443\\
\midrule
P2 & Gain & 11/5/0 & 0.2101 & 0.8405\\
P2 & Improved & 5/1 & 0.2188 & 0.8405\\
P2 & Absolute gain & 12/4/0 & 0.0768 & 0.3841\\
\bottomrule
\end{tabular}
\caption{The six planned shared-rollout tests, with 16 paired blocks per source. Sign-test counts are positive/negative/tied differences. Improved-run counts are pairs where only P2 improves / only P1 improves. Holm correction covers all six tests.}
\label{tab:shared-rollout-tests}
\end{table}

The mean signed gain difference, P2 learner minus P1 learner, is +3.88 points with P1 responses and +10.72 with P2 responses. Block-bootstrap 95\% intervals are [$-$5.06, 14.78] and [3.28, 18.97] points, using 10,000 resamples and seed 20260921. These marginal intervals concern the mean, whereas the planned sign tests concern the balance of positive and negative paired differences. The second interval excludes zero even though the corresponding corrected sign test does not reject its null.

\paragraph{Response source and learner both matter.}
For the P2 learner, P2 responses yield six improved runs and no terminal scores at or below 5\%, while P1 responses yield three improved runs and six such low endpoints. Both improved P1 runs use P2 responses. An exploratory interaction gives a +6.84-point mean change in the learner gap across sources, with 13 positive and three negative block differences (sign-test $p$ = 0.0213). Exploratory source tests give $p$ = 0.3018 for P1 learners and $p$ = 0.0768 for P2 learners. These tests were outside the planned six-test family. We do not compare the learners' collapse rates at the 5\% cutoff because P1 already starts below it.

Sharing responses fixes every correct/incorrect count and every rewarded-incorrect or unrewarded-correct event within a pair. It leaves the learners' likelihoods and gradients free to differ. The result therefore concerns what different checkpoints do with the same training material, complementing an account based on the material they generate. It does not identify which internal property causes the difference or test Chen et al.'s on-policy theorem in its own setting. The experiment also lacks a matched on-policy control with both reference coefficient and weight decay set to zero.

\paragraph{Answer extraction changes the apparent starting point.}
\label{app:shared-rollout-scoring}
We specified a second scorer before training: apply the same lenient extractor only to text before the first newline followed by \texttt{Question:}. Unlike the diagnostic used for the other frozen checkpoints, this rule does not stop at blank lines. It addresses outputs that answer the current question and then invent another one, from which the full-output extractor can select an unrelated number.

Under this prefix rule, P1 starts at 41.5\% and P2 at 14.5\%, reversing their primary-score ordering. Mean prefix scores fall for P1 and rise for P2 under both sources (Table~\ref{tab:shared-rollout-diagnostic}). In each source condition, eight P2 runs improve by more than ten points and none declines by that amount. P1 instead has ten or eleven declines, depending on the source. Its small changes under the primary scorer therefore do not establish that the model is unresponsive to training.

Output conventions change as well. The fraction of P2 responses containing \texttt{\#\#\#\#} or \texttt{\textbackslash boxed} falls from 93.5\% to 51.8\% with P1 responses and 47.4\% with P2 responses. Mean strict accuracy falls from 7.0\% to 2.6\% and 2.8\%. Neither extraction rule alone separates mathematical capability from answer presentation. The prefix rule has not been validated as a scorer, and no confirmatory tests were planned for it; we report it as a diagnostic of this sensitivity.

\begin{table}[htbp]
\centering\small
\setlength{\tabcolsep}{5pt}
\begin{tabular}{llccrr}
\toprule
Source & Learner & Primary score & Prefix score & Prefix gain & Prefix gain\\
& & before $\to$ after & before $\to$ after & $>10$ pp & $<-10$ pp\\
\midrule
P1 & P1 & $2.5\to3.1$ & $41.5\to28.8$ & 2/16 & 10/16\\
P1 & P2 & $13.5\to18.0$ & $14.5\to38.4$ & 8/16 & 0/16\\
P2 & P1 & $2.5\to5.4$ & $41.5\to21.2$ & 1/16 & 11/16\\
P2 & P2 & $13.5\to27.2$ & $14.5\to39.3$ & 8/16 & 0/16\\
\bottomrule
\end{tabular}
\caption{Shared-rollout answer-extraction diagnostic. Scores are percentages; the terminal score is the mean of 16 runs. Prefix scoring stops at the first generated newline followed by \texttt{Question:}. These descriptive results use the same rule before and after training.}
\label{tab:shared-rollout-diagnostic}
\end{table}

\begin{table}[htbp]
\centering\small
\begin{tabular}{rrrrr}
\toprule
Replicate & P1 $\to$ P1 & P1 $\to$ P2 & P2 $\to$ P1 & P2 $\to$ P2\\
\midrule
51001 & 3.0 & 4.0 & 1.5 & 48.5\\
51002 & 4.0 & 44.0 & 7.0 & 14.5\\
51003 & 4.0 & 23.0 & 1.0 & 13.5\\
51004 & 8.0 & 2.0 & 4.0 & 10.5\\
51005 & 1.0 & 43.5 & 3.0 & 47.5\\
51006 & 0.5 & 11.5 & 7.0 & 19.0\\
51007 & 1.5 & 4.5 & 4.0 & 11.5\\
51008 & 1.0 & 13.5 & 5.5 & 19.0\\
51009 & 1.0 & 2.5 & 7.0 & 57.0\\
51010 & 2.0 & 8.0 & 2.0 & 51.0\\
51011 & 9.5 & 17.5 & 13.0 & 21.5\\
51012 & 3.0 & 1.5 & 5.0 & 29.0\\
51013 & 1.5 & 3.5 & 9.0 & 14.5\\
51014 & 2.5 & 11.5 & 14.0 & 43.0\\
51015 & 3.0 & 17.5 & 0.0 & 16.5\\
51016 & 4.5 & 80.0 & 4.0 & 18.0\\
\bottomrule
\end{tabular}
\caption{All shared-rollout terminal scores (\%) on the 200 held-out GSM questions. Arrows give response source $\to$ learner. Each row is one paired block of four runs.}
\label{tab:shared-rollout-seeds}
\end{table}

\subsection{Group Size and Early Steering}
\label{app:cfa}

The earlier controls examine two choices held fixed in the checkpoint scan. At $G$ = 4, one of five P1 runs and two of five runs at each P2 checkpoint gain more than ten points. None of the corresponding $G$ = 8 runs does so (Tables~\ref{tab:cfa} and~\ref{tab:cfa-agg}). This observed difference is consistent with sensitivity to the training procedure. The exploratory sweep did not show the same sharp ordering, however, so the small controlled study does not establish a general group-size effect.

\begin{table}[htbp]
\centering\small
\caption{All 30 random-reward runs in the earlier group-size comparison. Entries give initial / terminal accuracy in percent, with changes in parentheses. Stars mark gains above 10 points. Every run uses learning rate $3\times10^{-5}$, reference coefficient 0.01, and 500 updates.}
\label{tab:cfa}
\begin{tabular}{llll}
\toprule
seed & P1 (3,896B) & P2 +5B & P2 +50B \\
\midrule
\multicolumn{4}{l}{\texttt{r\_g4} ($G$ = 4)}\\
777  & 3.0 / 7.5 (+4.5) & 13.0 / 26.5 (+13.5)$^\star$ & 21.0 / 1.5 ($-$19.5) \\
888  & 1.0 / 27.0 (+26.0)$^\star$ & 9.0 / 2.5 ($-$6.5) & 17.5 / 0.5 ($-$17.0) \\
999  & 4.5 / 9.0 (+4.5) & 20.0 / 17.5 ($-$2.5) & 27.0 / 76.5 (+49.5)$^\star$ \\
1001 & 4.5 / 0.0 ($-$4.5) & 14.0 / 19.5 (+5.5) & 19.5 / 6.0 ($-$13.5) \\
1002 & 1.0 / 8.0 (+7.0) & 14.5 / 49.0 (+34.5)$^\star$ & 16.0 / 85.5 (+69.5)$^\star$ \\
\midrule
\multicolumn{4}{l}{\texttt{r\_g8} ($G$ = 8)}\\
777  & 3.0 / 1.0 ($-$2.0) & 13.0 / 16.0 (+3.0) & 21.0 / 1.0 ($-$20.0) \\
888  & 1.0 / 0.5 ($-$0.5) & 9.0 / 12.5 (+3.5) & 17.5 / 26.5 (+9.0) \\
999  & 4.5 / 10.5 (+6.0) & 20.0 / 9.0 ($-$11.0) & 27.0 / 16.0 ($-$11.0) \\
1001 & 4.5 / 5.0 (+0.5) & 14.0 / 7.5 ($-$6.5) & 19.5 / 21.5 (+2.0) \\
1002 & 1.0 / 1.0 (0.0) & 14.5 / 5.5 ($-$9.0) & 16.0 / 5.5 ($-$10.5) \\
\bottomrule
\end{tabular}
\end{table}

\begin{table}[htbp]
\centering\small
\caption{Group-size outcomes at each checkpoint, with five seeds per condition. Success means a terminal gain above 10 points; mean changes are in percentage points.}
\label{tab:cfa-agg}
\begin{tabular}{lcccc}
\toprule
Checkpoint & $G$ = 4 successes & $G$ = 4 mean $\Delta$ & $G$ = 8 successes & $G$ = 8 mean $\Delta$ \\
\midrule
P1 3,896B & 1/5 & +7.5 & 0/5 & +0.8 \\
P2 +5B & 2/5 & +8.9 & 0/5 & $-$4.0 \\
P2 +50B & 2/5 & +13.8 & 0/5 & $-$6.1 \\
\bottomrule
\end{tabular}
\end{table}

GT steering for the first 50 steps yields three terminal jackpots out of five, compared with none under inverted-GT steering (Table~\ref{tab:cfbc}). The other two GT-steered runs also improve early: seed 888 reaches 78.5\% at step 100 before ending at 1.0\%, and seed 1001 reaches 90.5\% before ending at 3.0\%. No inverted-GT run exceeds 27.5\% at a scheduled evaluation. Early direction can therefore precede a large gain that subsequent random-reward updates fail to preserve.

\begin{table}[htbp]
\centering\small
\caption{GT and early-steering outcomes in the earlier study. Terminal accuracies follow seed order 777, 888, 999, 1001, 1002 and are reported in percent; mean changes are in percentage points. Steering is applied at P2 +5B.}
\label{tab:cfbc}
\begin{tabular}{llc}
\toprule
arm & after accuracy (777 / 888 / 999 / 1001 / 1002) & mean $\Delta$ \\
\midrule
\texttt{gt\_g4} P1 & 90.5 / 92.5 / 81.5 / 81.0 / 84.0 & +83.1 \\
\texttt{gt\_g4} P2 +5B  & 99.0 / 97.0 / 98.0 / 99.5 / 100.0 & +84.6 \\
\texttt{gt\_g4} P2 +50B & 100.0 / 98.0 / 100.0 / 98.0 / \textbf{21.5} & +63.3 \\
\midrule
\texttt{gt\_K50} & 90.5 / 1.0 / 62.0 / 3.0 / 54.0 & +28.0 \\
\texttt{anti\_K50} & 1.5 / 1.5 / 16.5 / 3.5 / 9.0 & $-$7.7 \\
\bottomrule
\end{tabular}
\end{table}

The paired jackpot comparison has three discordant seeds favoring GT steering and none favoring inverted GT, giving exact two-sided $p$ = 0.25 and Holm $p_H$ = 1. Table~\ref{tab:stats-historical} gives the full tests. These five-seed samples leave substantial uncertainty about the rates, and their seed-dependent evaluation sets are kept separate from the fixed-split scan.

\clearpage
\subsection{Statistical Comparisons and Uncertainty}
\label{app:statistics}

The best observed score shows what one run attained. The tests below ask whether differences extend across seeds. They report all comparisons in the families defined in Appendix~\ref{app:stat-methods}, including nonsignificant results. Except for the separately planned shared-rollout tests, these analyses are exploratory.

\subsubsection{Training inputs}

On OLMo-2, mean absolute change is 12.59 points in D and 1.11 in E, a difference of 11.48 points ($p=4.66\times10^{-10}$; within-family Holm $p_H=6.98\times10^{-9}$). Every paired seed changes more in D. Changes exceeding ten points in either direction also differ, 16/32 versus 0/32 ($p_H=4.27\times10^{-4}$). Both comparisons survive correction across all 290 tests (Table~\ref{tab:stats-inputs}).

The same evidence does not establish more frequent improvement. Positive gains above ten points occur in 4/32 D runs and 0/32 E runs, giving $p$ = 0.125 and $p_H$ = 1; D's mean signed gain is lower, though not significantly so. Qwen and Llama also show larger observed absolute changes in D, but neither survives the D/E correction. The strongest statistical finding is that the OLMo-2 model changes more under task-formatted inputs, including both gains and losses.

\begin{table}[!htbp]
\centering\small
\setlength{\tabcolsep}{3.2pt}
\caption{Paired D/E comparisons on OLMo-2 (32 seeds), Qwen2.5 (eight), and Llama (eight). Effects are D minus E in percentage points; binary rows also report event counts. $p_H$ corrects the 15 tests in this table, and $p_{\rm all}$ corrects all 290 exploratory tests.}
\label{tab:stats-inputs}
\begin{tabular}{lrrrrrr}
\toprule
Quantity & D & E & Difference & $p$ & $p_H$ & $p_{\rm all}$\\
\midrule
\multicolumn{7}{l}{\textbf{OLMo-2}}\\
Mean signed gain (pp) & $-$3.00 & $-$0.14 & $-$2.86 & 0.3337 & 1 & 1\\
Mean absolute change (pp) & 12.59 & 1.11 & +11.48 & $4.66\!\times\!10^{-10}$ & $6.98\!\times\!10^{-9}$ & $1.35\!\times\!10^{-7}$\\
Gain $>10$ pp & 4/32 & 0/32 & +12.50 & 0.125 & 1 & 1\\
Absolute change $>10$ pp & 16/32 & 0/32 & +50.00 & $3.05\!\times\!10^{-5}$ & $4.27\!\times\!10^{-4}$ & 0.0088\\
Terminal score at most 5\% & 8/32 & 0/32 & +25.00 & 0.0078 & 0.1016 & 1\\
\addlinespace
\multicolumn{7}{l}{\textbf{Qwen2.5}}\\
Mean signed gain (pp) & 6.56 & 2.00 & +4.56 & 0.7188 & 1 & 1\\
Mean absolute change (pp) & 25.44 & 2.50 & +22.94 & 0.0078 & 0.1016 & 1\\
Gain $>10$ pp & 4/8 & 0/8 & +50.00 & 0.125 & 1 & 1\\
Absolute change $>10$ pp & 7/8 & 0/8 & +87.50 & 0.0156 & 0.1719 & 1\\
Terminal score at most 5\% & 1/8 & 0/8 & +12.50 & 1 & 1 & 1\\
\addlinespace
\multicolumn{7}{l}{\textbf{Llama-3.1}}\\
Mean signed gain (pp) & $-$8.88 & $-$5.94 & $-$2.94 & 0.6172 & 1 & 1\\
Mean absolute change (pp) & 18.50 & 8.31 & +10.19 & 0.0391 & 0.3906 & 1\\
Gain $>10$ pp & 1/8 & 0/8 & +12.50 & 1 & 1 & 1\\
Absolute change $>10$ pp & 6/8 & 2/8 & +50.00 & 0.125 & 1 & 1\\
Terminal score at most 5\% & 3/8 & 0/8 & +37.50 & 0.25 & 1 & 1\\
\addlinespace
\bottomrule
\end{tabular}
\end{table}

\begin{table}[!htbp]
\centering\small
\setlength{\tabcolsep}{3.2pt}
\caption{Input-control event counts and marginal 95\% Clopper--Pearson intervals (\%). Gain means $\Delta>10$ points; change means $|\Delta|>10$ points. Each interval describes one model, input condition, and protocol.}
\label{tab:stats-input-intervals}
\begin{tabular}{llrr}
\toprule
Model & Input & Gain: count [95\% CI] & Change: count [95\% CI]\\
\midrule
OLMo-2 & B & 3/32 [2.0, 25.0] & 21/32 [46.8, 81.4]\\
OLMo-2 & C & 2/16 [1.6, 38.3] & 10/16 [35.4, 84.8]\\
OLMo-2 & A & 2/16 [1.6, 38.3] & 5/16 [11.0, 58.7]\\
OLMo-2 & D & 4/32 [3.5, 29.0] & 16/32 [31.9, 68.1]\\
OLMo-2 & E & 0/32 [0.0, 10.9] & 0/32 [0.0, 10.9]\\
Qwen2.5 & D & 4/8 [15.7, 84.3] & 7/8 [47.3, 99.7]\\
Qwen2.5 & E & 0/8 [0.0, 36.9] & 0/8 [0.0, 36.9]\\
Llama-3.1 & D & 1/8 [0.3, 52.7] & 6/8 [34.9, 96.8]\\
Llama-3.1 & E & 0/8 [0.0, 36.9] & 2/8 [3.2, 65.1]\\
\bottomrule
\end{tabular}
\end{table}

The intervals show the remaining uncertainty in success rates (Table~\ref{tab:stats-input-intervals}). OLMo-2's 4/32 rate has a 95\% interval of 3.5--29.0\%, while 0/32 has an upper endpoint of 10.9\%. For Qwen and Llama, 0/8 leaves an upper endpoint of 36.9\%. Table~\ref{tab:stats-other-inputs} compares all other OLMo-2 input pairs using their shared seeds; the descriptive results retain the additional unmatched seeds.

\begin{table}[!htbp]
\centering\small
\setlength{\tabcolsep}{3.2pt}
\caption{Other paired OLMo-2 input comparisons, using shared seeds. The first triplet gives mean signed-gain differences; the second gives differences in the rate of gains above ten points. Effects are percentage points. $p_H$ corrects all 18 tests; D/E appears separately in Table~\ref{tab:stats-inputs}.}
\label{tab:stats-other-inputs}
\begin{tabular}{lrrrrrrr}
\toprule
Pair & $n$ & Mean diff. & $p$ & $p_H$ & Rate diff. & $p$ & $p_H$\\
\midrule
B $-$ C & 16 & $-$2.59 & 0.5857 & 1 & $-$6.25 & 1 & 1\\
B $-$ A & 16 & $-$10.31 & 0.0023 & 0.0417 & $-$6.25 & 1 & 1\\
B $-$ D & 32 & $-$3.64 & 0.4467 & 1 & $-$3.12 & 1 & 1\\
B $-$ E & 32 & $-$6.50 & 0.0672 & 1 & +9.38 & 0.25 & 1\\
C $-$ A & 16 & $-$7.72 & 0.1971 & 1 & 0.00 & 1 & 1\\
C $-$ D & 16 & $-$5.16 & 0.3455 & 1 & 0.00 & 1 & 1\\
C $-$ E & 16 & $-$4.91 & 0.2028 & 1 & +12.50 & 0.5 & 1\\
A $-$ D & 16 & +2.56 & 0.7046 & 1 & 0.00 & 1 & 1\\
A $-$ E & 16 & +2.81 & 0.6747 & 1 & +12.50 & 0.5 & 1\\
\bottomrule
\end{tabular}
\end{table}

\subsubsection{Checkpoint development}

For the P1/P2 comparison, we average scores within each phase and seed before taking the paired difference. Across the eight Random seed blocks, P2 exceeds P1 by 17.32 points ($p$ = 0.0078, $p_H$ = 0.0938 across 12 developmental tests). The difference in the frequency of terminal scores at least 50\% is 15.63 points ($p$ = 0.0156, $p_H$ = 0.1719). These contrasts summarize the selected checkpoints; they do not locate a universal developmental transition (Table~\ref{tab:stats-development}).

The 462B/839B comparison has a 25.88-point difference in mean GT terminal score, with all four paired differences positive. With four pairs, however, the smallest two-sided exact permutation $p$ is 0.125. The example demonstrates an observed separation between current score and training response while leaving its population magnitude uncertain. The additional-model GT comparisons have only two paired seeds, giving a minimum $p$ of 0.5. Code phase comparisons likewise use only one common GT seed and two common Random seeds. Tables~\ref{tab:stats-development} and~\ref{tab:stats-crossmodel} report these effect sizes alongside the tests their sample sizes permit.

\begin{table}[!htbp]
\centering\small
\setlength{\tabcolsep}{3.2pt}
\caption{OLMo-2 developmental contrasts. Effects are later minus earlier terminal scores in percentage points, except the last row, which compares high-score frequencies. P1/P2 contrasts average checkpoints within phase and seed before pairing. $p_H$ corrects 12 tests. Code GT has one common seed and no inferential test.}
\label{tab:stats-development}
\begin{tabular}{lrrrr}
\toprule
Comparison & Seeds & Difference & $p$ & $p_H$\\
\midrule
GSM GT: P1 3,896B $-$ P1 34B & 4 & +81.12 & 0.125 & 0.875\\
GSM GT: P2 +5B $-$ P1 3,896B & 4 & +17.38 & 0.125 & 0.875\\
GSM GT: P1 839B $-$ P1 462B & 4 & +25.88 & 0.125 & 0.875\\
GSM Random: P1 3,896B $-$ P1 34B & 8 & +5.81 & 0.0156 & 0.1719\\
GSM Random: P2 +5B $-$ P1 3,896B & 8 & +33.44 & 0.0156 & 0.1719\\
GSM Random: P1 839B $-$ P1 462B & 8 & 0.00 & 1 & 1\\
GSM GT: P2 $-$ P1 & 4 & +39.45 & 0.125 & 0.875\\
GSM Random: P2 $-$ P1 & 8 & +17.32 & 0.0078 & 0.0938\\
Code GT: P2 $-$ P1 & 1 & +19.62 & -- & --\\
Code Random: P2 $-$ P1 & 2 & +3.03 & 0.5 & 1\\
Lookup GT: P2 $-$ P1 & 2 & +12.97 & 0.5 & 1\\
Lookup Random: P2 $-$ P1 & 8 & +17.85 & 0.0156 & 0.1719\\
GSM Random: P2 $-$ P1 score 50\% or more & 8 & +15.62 & 0.0156 & 0.1719\\
\bottomrule
\end{tabular}
\end{table}

\begin{table}[!htbp]
\centering\small
\setlength{\tabcolsep}{3.2pt}
\caption{All checkpoint pairs within SmolLM2 and OLMo-3. Triplets give mean terminal-score differences (percentage points), raw $p$, and Holm $p$ for GT and Random. GT uses two paired seeds and Random eight. Correction covers these 24 tests and the eight GT/Random tests in Table~\ref{tab:stats-cell-xmodel}.}
\label{tab:stats-crossmodel}
\begin{tabular}{lrrrrrr}
\toprule
Pair & GT diff. & $p$ & $p_H$ & Random diff. & $p$ & $p_H$\\
\midrule
S2 $-$ S1 & 0.00 & 1 & 1 & +0.56 & 0.4375 & 1\\
S3 $-$ S1 & +1.25 & 0.5 & 1 & +1.88 & 0.0156 & 0.4219\\
S4 $-$ S1 & +65.25 & 0.5 & 1 & +4.12 & 0.0078 & 0.25\\
S3 $-$ S2 & +1.25 & 1 & 1 & +1.31 & 0.3594 & 1\\
S4 $-$ S2 & +65.25 & 0.5 & 1 & +3.56 & 0.0234 & 0.5859\\
S4 $-$ S3 & +64.00 & 0.5 & 1 & +2.25 & 0.0469 & 1\\
O2 $-$ O1 & +91.50 & 0.5 & 1 & +7.69 & 0.0156 & 0.4219\\
O3 $-$ O1 & +61.50 & 0.5 & 1 & +13.44 & 0.0078 & 0.25\\
O4 $-$ O1 & +99.50 & 0.5 & 1 & +47.06 & 0.0078 & 0.25\\
O3 $-$ O2 & $-$30.00 & 1 & 1 & +5.75 & 0.2109 & 1\\
O4 $-$ O2 & +8.00 & 0.5 & 1 & +39.38 & 0.0078 & 0.25\\
O4 $-$ O3 & +38.00 & 0.5 & 1 & +33.63 & 0.0078 & 0.25\\
\bottomrule
\end{tabular}
\end{table}

\subsubsection{Earlier controls and individual conditions}

None of the earlier group-size or steering comparisons survives the ten-test correction (Table~\ref{tab:stats-historical}). For steering, the 35.70-point mean-gain difference and the 3/5 versus 0/5 jackpot comparison both have unadjusted $p$ = 0.25. Averaging the group-size differences over three checkpoints leaves five seed blocks, rather than fifteen independent observations.

\begin{table}[!htbp]
\centering\small
\setlength{\tabcolsep}{3.2pt}
\caption{Group-size and steering comparisons with five seed blocks. The first four rows compare $G$ = 4 with $G$ = 8, including an average over checkpoints; the last compares GT and inverted-GT steering. Triplets report mean-gain and jackpot-rate differences (percentage points), raw $p$, and Holm $p$ over all ten tests.}
\label{tab:stats-historical}
\begin{tabular}{lrrrrrr}
\toprule
Comparison & Mean diff. & $p$ & $p_H$ & Rate diff. & $p$ & $p_H$\\
\midrule
P1 3,896B & +6.70 & 0.3125 & 1 & +20.00 & 1 & 1\\
P2 +5B & +12.90 & 0.1875 & 1 & +40.00 & 0.5 & 1\\
P2 +50B & +19.90 & 0.4375 & 1 & +40.00 & 0.5 & 1\\
Three-checkpoint mean & +13.17 & 0.3125 & 1 & +33.33 & 0.125 & 1\\
GT $-$ anti steering & +35.70 & 0.25 & 1 & +60.00 & 0.25 & 1\\
\bottomrule
\end{tabular}
\end{table}

Tables~\ref{tab:stats-cell-gsm}--\ref{tab:stats-other-cells} report changes within every condition and paired GT/Random differences. A within-condition sign test concerns whether a nonzero change is more often positive or negative. It does not test mean gain: a few large gains can coexist with many smaller losses. Tables~\ref{tab:stats-rates-1}--\ref{tab:stats-rates-3} give checkpoint-specific intervals for Random gain and high-score probabilities. SFT lacks per-seed results, so it enters neither these tests nor the interval estimates.

\begin{table}[!htbp]
\centering\small
\setlength{\tabcolsep}{3.2pt}
\caption{GSM changes within each checkpoint and reward condition. Tables~\ref{tab:stats-cell-code}--\ref{tab:stats-cell-xmodel} use the same layout. Up/down/tie counts use each run\textquotesingle s initial score. $p_{\rm dir}$ is a two-sided sign test; its $p_H$ corrects 150 within-condition tests. The final triplet gives the paired GT minus Random mean score (percentage points), permutation $p$, and correction over 53 scan GT/Random tests. One-seed comparisons have no $p$ value.}
\label{tab:stats-cell-gsm}
\begin{tabular}{lrrrrrrrrr}
\toprule
Checkpoint & GT $\uparrow/\downarrow/=$ & $p_{\rm dir}$ & $p_H$ & R $\uparrow/\downarrow/=$ & $p_{\rm dir}$ & $p_H$ & GT$-$R & $p$ & $p_H$\\
\midrule
P1 5B & 1/0/3 & 1 & 1 & 4/0/4 & 0.125 & 1 & +0.12 & 1 & 1\\
P1 34B & 2/1/1 & 1 & 1 & 5/1/2 & 0.2188 & 1 & $-$0.38 & 0.75 & 1\\
P1 462B & 3/0/1 & 0.25 & 1 & 3/4/1 & 1 & 1 & +1.75 & 0.125 & 1\\
P1 839B & 4/0/0 & 0.125 & 1 & 3/4/1 & 1 & 1 & +29.00 & 0.125 & 1\\
P1 1,259B & 3/1/0 & 0.625 & 1 & 6/1/1 & 0.125 & 1 & +47.12 & 0.25 & 1\\
P1 1,469B & 4/0/0 & 0.125 & 1 & 1/5/2 & 0.2188 & 1 & +15.62 & 0.125 & 1\\
P1 1,678B & 3/1/0 & 0.625 & 1 & 3/4/1 & 1 & 1 & +24.62 & 0.25 & 1\\
P1 1,888B & 4/0/0 & 0.125 & 1 & 3/2/3 & 1 & 1 & +26.38 & 0.125 & 1\\
P1 2,098B & 4/0/0 & 0.125 & 1 & 5/2/1 & 0.4531 & 1 & +61.75 & 0.125 & 1\\
P1 2,307B & 4/0/0 & 0.125 & 1 & 2/6/0 & 0.2891 & 1 & +72.62 & 0.125 & 1\\
P1 2,517B & 4/0/0 & 0.125 & 1 & 4/3/1 & 1 & 1 & +58.38 & 0.125 & 1\\
P1 2,727B & 4/0/0 & 0.125 & 1 & 6/2/0 & 0.2891 & 1 & +71.88 & 0.125 & 1\\
P1 2,937B & 4/0/0 & 0.125 & 1 & 3/5/0 & 0.7266 & 1 & +78.75 & 0.125 & 1\\
P1 3,020B & 4/0/0 & 0.125 & 1 & 5/3/0 & 0.7266 & 1 & +22.88 & 0.125 & 1\\
P1 3,272B & 4/0/0 & 0.125 & 1 & 4/2/2 & 0.6875 & 1 & +40.62 & 0.25 & 1\\
P1 3,440B & 4/0/0 & 0.125 & 1 & 4/3/1 & 1 & 1 & +24.38 & 0.125 & 1\\
P1 3,532B & 4/0/0 & 0.125 & 1 & 5/2/1 & 0.4531 & 1 & +80.50 & 0.125 & 1\\
P1 3,608B & 4/0/0 & 0.125 & 1 & 4/4/0 & 1 & 1 & +75.38 & 0.125 & 1\\
P1 3,691B & 4/0/0 & 0.125 & 1 & 4/4/0 & 1 & 1 & +82.75 & 0.125 & 1\\
P1 3,775B & 4/0/0 & 0.125 & 1 & 4/4/0 & 1 & 1 & +75.37 & 0.125 & 1\\
P1 3,817B & 4/0/0 & 0.125 & 1 & 4/4/0 & 1 & 1 & +80.88 & 0.125 & 1\\
P1 3,859B & 4/0/0 & 0.125 & 1 & 4/3/1 & 1 & 1 & +59.13 & 0.125 & 1\\
P1 3,880B & 4/0/0 & 0.125 & 1 & 3/3/2 & 1 & 1 & +79.00 & 0.125 & 1\\
P1 3,893B & 4/0/0 & 0.125 & 1 & 6/2/0 & 0.2891 & 1 & +77.62 & 0.125 & 1\\
P1 3,896B & 4/0/0 & 0.125 & 1 & 6/2/0 & 0.2891 & 1 & +73.62 & 0.125 & 1\\
P2 +5B & 4/0/0 & 0.125 & 1 & 6/2/0 & 0.2891 & 1 & +62.13 & 0.125 & 1\\
P2 +9B & 4/0/0 & 0.125 & 1 & 3/5/0 & 0.7266 & 1 & +14.75 & 0.625 & 1\\
P2 +13B & 4/0/0 & 0.125 & 1 & 1/7/0 & 0.0703 & 1 & +85.88 & 0.125 & 1\\
P2 +17B & 4/0/0 & 0.125 & 1 & 2/6/0 & 0.2891 & 1 & +84.88 & 0.125 & 1\\
P2 +21B & 4/0/0 & 0.125 & 1 & 3/5/0 & 0.7266 & 1 & +49.50 & 0.125 & 1\\
P2 +26B & 4/0/0 & 0.125 & 1 & 2/6/0 & 0.2891 & 1 & +77.75 & 0.125 & 1\\
P2 +30B & 4/0/0 & 0.125 & 1 & 3/5/0 & 0.7266 & 1 & +75.50 & 0.125 & 1\\
P2 +34B & 4/0/0 & 0.125 & 1 & 6/2/0 & 0.2891 & 1 & +71.75 & 0.125 & 1\\
P2 +38B & 4/0/0 & 0.125 & 1 & 3/5/0 & 0.7266 & 1 & +90.12 & 0.125 & 1\\
P2 +42B & 4/0/0 & 0.125 & 1 & 5/3/0 & 0.7266 & 1 & +59.75 & 0.125 & 1\\
P2 +47B & 4/0/0 & 0.125 & 1 & 3/5/0 & 0.7266 & 1 & +80.62 & 0.125 & 1\\
P2 +50B & 4/0/0 & 0.125 & 1 & 2/5/1 & 0.4531 & 1 & +86.38 & 0.125 & 1\\
\bottomrule
\end{tabular}
\end{table}

\begin{table}[!htbp]
\centering\small
\setlength{\tabcolsep}{3.2pt}
\caption{Code changes within each checkpoint and reward condition. Columns and tests follow Table~\ref{tab:stats-cell-gsm}.}
\label{tab:stats-cell-code}
\begin{tabular}{lrrrrrrrrr}
\toprule
Checkpoint & GT $\uparrow/\downarrow/=$ & $p_{\rm dir}$ & $p_H$ & R $\uparrow/\downarrow/=$ & $p_{\rm dir}$ & $p_H$ & GT$-$R & $p$ & $p_H$\\
\midrule
P1 5B & 0/0/1 & -- & -- & 0/0/2 & 1 & 1 & 0.00 & -- & --\\
P1 462B & 0/0/1 & -- & -- & 1/1/0 & 1 & 1 & +0.50 & -- & --\\
P1 839B & 0/0/1 & -- & -- & 0/0/2 & 1 & 1 & 0.00 & -- & --\\
P1 1,259B & 0/0/1 & -- & -- & 0/2/0 & 0.5 & 1 & +0.50 & -- & --\\
P1 2,098B & 1/0/0 & -- & -- & 1/1/0 & 1 & 1 & +9.50 & -- & --\\
P1 2,937B & 1/0/0 & -- & -- & 2/0/0 & 0.5 & 1 & 0.00 & -- & --\\
P1 3,532B & 1/0/0 & -- & -- & 0/0/2 & 1 & 1 & +0.50 & -- & --\\
P1 3,896B & 0/0/1 & -- & -- & 0/0/2 & 1 & 1 & 0.00 & -- & --\\
P2 +5B & 4/0/0 & 0.125 & 1 & 2/6/0 & 0.2891 & 1 & +15.88 & 0.125 & 1\\
P2 +13B & 4/0/0 & 0.125 & 1 & 4/0/4 & 0.125 & 1 & +19.75 & 0.125 & 1\\
P2 +26B & 4/0/0 & 0.125 & 1 & 2/0/6 & 0.5 & 1 & +22.62 & 0.125 & 1\\
P2 +50B & 4/0/0 & 0.125 & 1 & 2/0/6 & 0.5 & 1 & +23.25 & 0.125 & 1\\
\bottomrule
\end{tabular}
\end{table}

\begin{table}[!htbp]
\centering\small
\setlength{\tabcolsep}{3.2pt}
\caption{Lookup changes within each checkpoint and reward condition. Columns and tests follow Table~\ref{tab:stats-cell-gsm}.}
\label{tab:stats-cell-lookup}
\begin{tabular}{lrrrrrrrrr}
\toprule
Checkpoint & GT $\uparrow/\downarrow/=$ & $p_{\rm dir}$ & $p_H$ & R $\uparrow/\downarrow/=$ & $p_{\rm dir}$ & $p_H$ & GT$-$R & $p$ & $p_H$\\
\midrule
P1 5B & 0/0/2 & 1 & 1 & 0/0/8 & 1 & 1 & 0.00 & 1 & 1\\
P1 462B & 2/0/0 & 0.5 & 1 & 5/3/0 & 0.7266 & 1 & +83.25 & 0.5 & 1\\
P1 839B & 2/0/0 & 0.5 & 1 & 2/5/1 & 0.4531 & 1 & +78.50 & 0.5 & 1\\
P1 1,259B & 2/0/0 & 0.5 & 1 & 2/6/0 & 0.2891 & 1 & +77.75 & 0.5 & 1\\
P1 2,098B & 2/0/0 & 0.5 & 1 & 0/8/0 & 0.0078 & 1 & +87.50 & 0.5 & 1\\
P1 2,937B & 2/0/0 & 0.5 & 1 & 3/5/0 & 0.7266 & 1 & +97.75 & 0.5 & 1\\
P1 3,532B & 2/0/0 & 0.5 & 1 & 6/2/0 & 0.2891 & 1 & +90.50 & 0.5 & 1\\
P1 3,896B & 2/0/0 & 0.5 & 1 & 5/3/0 & 0.7266 & 1 & +78.75 & 0.5 & 1\\
P2 +5B & 2/0/0 & 0.5 & 1 & 1/7/0 & 0.0703 & 1 & +98.50 & 0.5 & 1\\
P2 +13B & 2/0/0 & 0.5 & 1 & 2/6/0 & 0.2891 & 1 & +29.00 & 0.5 & 1\\
P2 +26B & 2/0/0 & 0.5 & 1 & 0/8/0 & 0.0078 & 1 & +88.00 & 0.5 & 1\\
P2 +50B & 2/0/0 & 0.5 & 1 & 2/6/0 & 0.2891 & 1 & +83.00 & 0.5 & 1\\
\bottomrule
\end{tabular}
\end{table}

\begin{table}[!htbp]
\centering\small
\setlength{\tabcolsep}{3.2pt}
\caption{SmolLM2 and OLMo-3 changes within each checkpoint and reward condition. Columns follow Table~\ref{tab:stats-cell-gsm}, except that the GT minus Random correction covers 32 additional-model tests.}
\label{tab:stats-cell-xmodel}
\begin{tabular}{lrrrrrrrrr}
\toprule
Checkpoint & GT $\uparrow/\downarrow/=$ & $p_{\rm dir}$ & $p_H$ & R $\uparrow/\downarrow/=$ & $p_{\rm dir}$ & $p_H$ & GT$-$R & $p$ & $p_H$\\
\midrule
O1 & 0/2/0 & 0.5 & 1 & 0/5/3 & 0.0625 & 1 & 0.00 & 1 & 1\\
O2 & 2/0/0 & 0.5 & 1 & 0/8/0 & 0.0078 & 1 & +79.25 & 0.5 & 1\\
O3 & 2/0/0 & 0.5 & 1 & 3/5/0 & 0.7266 & 1 & +49.25 & 0.5 & 1\\
O4 & 2/0/0 & 0.5 & 1 & 4/4/0 & 1 & 1 & +44.50 & 0.5 & 1\\
S1 & 1/0/1 & 1 & 1 & 1/0/7 & 1 & 1 & +0.25 & 1 & 1\\
S2 & 2/0/0 & 0.5 & 1 & 6/0/2 & 0.0312 & 1 & +0.25 & 1 & 1\\
S3 & 1/0/1 & 1 & 1 & 4/1/3 & 0.375 & 1 & $-$3.25 & 0.5 & 1\\
S4 & 2/0/0 & 0.5 & 1 & 6/2/0 & 0.2891 & 1 & +60.75 & 0.5 & 1\\
\bottomrule
\end{tabular}
\end{table}

\begin{table}[!htbp]
\centering\small
\setlength{\tabcolsep}{3.2pt}
\caption{Changes within each input-control and earlier-study condition. Mean gains are percentage points. Sign tests compare upward with downward nonzero changes, with Holm correction over 150 tests. The final column gives the jackpot count and its marginal 95\% interval (\%).}
\label{tab:stats-other-cells}
\begin{tabular}{lrrrrr}
\toprule
Cell & Mean gain & $\uparrow/\downarrow/=$ & $p_{\rm dir}$ & $p_H$ & Gain: count [95\% CI]\\
\midrule
OLMo-2 B & $-$6.64 & 6/25/1 & $8.78\!\times\!10^{-4}$ & 0.1317 & 3/32 [2.0, 25.0]\\
OLMo-2 C & $-$4.94 & 6/9/1 & 0.6072 & 1 & 2/16 [1.6, 38.3]\\
OLMo-2 A & +2.78 & 7/9/0 & 0.8036 & 1 & 2/16 [1.6, 38.3]\\
OLMo-2 D & $-$3.00 & 13/19/0 & 0.3771 & 1 & 4/32 [3.5, 29.0]\\
OLMo-2 E & $-$0.14 & 13/13/6 & 1 & 1 & 0/32 [0.0, 10.9]\\
Qwen2.5 D & +6.56 & 4/4/0 & 1 & 1 & 4/8 [15.7, 84.3]\\
Qwen2.5 E & +2.00 & 6/2/0 & 0.2891 & 1 & 0/8 [0.0, 36.9]\\
Llama-3.1 D & $-$8.88 & 1/7/0 & 0.0703 & 1 & 1/8 [0.3, 52.7]\\
Llama-3.1 E & $-$5.94 & 2/6/0 & 0.2891 & 1 & 0/8 [0.0, 36.9]\\
P1 3,896B r\_g4 & +7.50 & 4/1/0 & 0.375 & 1 & 1/5 [0.5, 71.6]\\
P1 3,896B r\_g8 & +0.80 & 2/2/1 & 1 & 1 & 0/5 [0.0, 52.2]\\
P2 +5B r\_g4 & +8.90 & 3/2/0 & 1 & 1 & 2/5 [5.3, 85.3]\\
P2 +5B r\_g8 & $-$4.00 & 2/3/0 & 1 & 1 & 0/5 [0.0, 52.2]\\
P2 +50B r\_g4 & +13.80 & 2/3/0 & 1 & 1 & 2/5 [5.3, 85.3]\\
P2 +50B r\_g8 & $-$6.10 & 2/3/0 & 1 & 1 & 0/5 [0.0, 52.2]\\
P1 3,896B gt\_g4 & +83.10 & 5/0/0 & 0.0625 & 1 & 5/5 [47.8, 100.0]\\
P2 +5B gt\_g4 & +84.60 & 5/0/0 & 0.0625 & 1 & 5/5 [47.8, 100.0]\\
P2 +50B gt\_g4 & +63.30 & 5/0/0 & 0.0625 & 1 & 4/5 [28.4, 99.5]\\
P2 +5B gt\_K50 & +28.00 & 3/2/0 & 1 & 1 & 3/5 [14.7, 94.7]\\
P2 +5B anti\_K50 & $-$7.70 & 0/5/0 & 0.0625 & 1 & 0/5 [0.0, 52.2]\\
\bottomrule
\end{tabular}
\end{table}

\begin{table}[!htbp]
\centering\small
\setlength{\tabcolsep}{3.2pt}
\caption{Random-run success rates, part 1. Entries give success count / seed count and marginal 95\% Clopper--Pearson intervals (\%). Gain means $\Delta>10$ points and high score means terminal accuracy 50\% or more. Each checkpoint is estimated separately.}
\label{tab:stats-rates-1}
\begin{tabular}{lrr}
\toprule
Checkpoint / task & Gain: count [95\% CI] & High score: count [95\% CI]\\
\midrule
GSM P1 5B & 0/8 [0.0, 36.9] & 0/8 [0.0, 36.9]\\
GSM P1 34B & 0/8 [0.0, 36.9] & 0/8 [0.0, 36.9]\\
GSM P1 462B & 0/8 [0.0, 36.9] & 0/8 [0.0, 36.9]\\
GSM P1 839B & 0/8 [0.0, 36.9] & 0/8 [0.0, 36.9]\\
GSM P1 1,259B & 0/8 [0.0, 36.9] & 0/8 [0.0, 36.9]\\
GSM P1 1,469B & 0/8 [0.0, 36.9] & 0/8 [0.0, 36.9]\\
GSM P1 1,678B & 0/8 [0.0, 36.9] & 0/8 [0.0, 36.9]\\
GSM P1 1,888B & 0/8 [0.0, 36.9] & 0/8 [0.0, 36.9]\\
GSM P1 2,098B & 0/8 [0.0, 36.9] & 0/8 [0.0, 36.9]\\
GSM P1 2,307B & 0/8 [0.0, 36.9] & 0/8 [0.0, 36.9]\\
GSM P1 2,517B & 0/8 [0.0, 36.9] & 0/8 [0.0, 36.9]\\
GSM P1 2,727B & 0/8 [0.0, 36.9] & 0/8 [0.0, 36.9]\\
GSM P1 2,937B & 0/8 [0.0, 36.9] & 0/8 [0.0, 36.9]\\
GSM P1 3,020B & 0/8 [0.0, 36.9] & 0/8 [0.0, 36.9]\\
GSM P1 3,272B & 1/8 [0.3, 52.7] & 0/8 [0.0, 36.9]\\
GSM P1 3,440B & 0/8 [0.0, 36.9] & 0/8 [0.0, 36.9]\\
GSM P1 3,532B & 0/8 [0.0, 36.9] & 0/8 [0.0, 36.9]\\
GSM P1 3,608B & 0/8 [0.0, 36.9] & 0/8 [0.0, 36.9]\\
GSM P1 3,691B & 0/8 [0.0, 36.9] & 0/8 [0.0, 36.9]\\
GSM P1 3,775B & 1/8 [0.3, 52.7] & 0/8 [0.0, 36.9]\\
GSM P1 3,817B & 1/8 [0.3, 52.7] & 0/8 [0.0, 36.9]\\
GSM P1 3,859B & 2/8 [3.2, 65.1] & 0/8 [0.0, 36.9]\\
GSM P1 3,880B & 0/8 [0.0, 36.9] & 0/8 [0.0, 36.9]\\
GSM P1 3,893B & 1/8 [0.3, 52.7] & 0/8 [0.0, 36.9]\\
\bottomrule
\end{tabular}
\end{table}

\begin{table}[!htbp]
\centering\small
\setlength{\tabcolsep}{3.2pt}
\caption{Random-run success rates, part 2. Entries give success count / seed count and marginal 95\% Clopper--Pearson intervals (\%). Gain means $\Delta>10$ points and high score means terminal accuracy 50\% or more. Each checkpoint is estimated separately.}
\label{tab:stats-rates-2}
\begin{tabular}{lrr}
\toprule
Checkpoint / task & Gain: count [95\% CI] & High score: count [95\% CI]\\
\midrule
GSM P1 3,896B & 0/8 [0.0, 36.9] & 0/8 [0.0, 36.9]\\
GSM P2 +5B & 4/8 [15.7, 84.3] & 4/8 [15.7, 84.3]\\
GSM P2 +9B & 2/8 [3.2, 65.1] & 2/8 [3.2, 65.1]\\
GSM P2 +13B & 0/8 [0.0, 36.9] & 0/8 [0.0, 36.9]\\
GSM P2 +17B & 0/8 [0.0, 36.9] & 0/8 [0.0, 36.9]\\
GSM P2 +21B & 3/8 [8.5, 75.5] & 3/8 [8.5, 75.5]\\
GSM P2 +26B & 0/8 [0.0, 36.9] & 0/8 [0.0, 36.9]\\
GSM P2 +30B & 2/8 [3.2, 65.1] & 1/8 [0.3, 52.7]\\
GSM P2 +34B & 4/8 [15.7, 84.3] & 0/8 [0.0, 36.9]\\
GSM P2 +38B & 2/8 [3.2, 65.1] & 1/8 [0.3, 52.7]\\
GSM P2 +42B & 3/8 [8.5, 75.5] & 3/8 [8.5, 75.5]\\
GSM P2 +47B & 2/8 [3.2, 65.1] & 1/8 [0.3, 52.7]\\
GSM P2 +50B & 0/8 [0.0, 36.9] & 0/8 [0.0, 36.9]\\
CODE P1 5B & 0/2 [0.0, 84.2] & 0/2 [0.0, 84.2]\\
CODE P1 462B & 0/2 [0.0, 84.2] & 0/2 [0.0, 84.2]\\
CODE P1 839B & 0/2 [0.0, 84.2] & 0/2 [0.0, 84.2]\\
CODE P1 1,259B & 0/2 [0.0, 84.2] & 0/2 [0.0, 84.2]\\
CODE P1 2,098B & 0/2 [0.0, 84.2] & 0/2 [0.0, 84.2]\\
CODE P1 2,937B & 0/2 [0.0, 84.2] & 0/2 [0.0, 84.2]\\
CODE P1 3,532B & 0/2 [0.0, 84.2] & 0/2 [0.0, 84.2]\\
CODE P1 3,896B & 0/2 [0.0, 84.2] & 0/2 [0.0, 84.2]\\
CODE P2 +5B & 1/8 [0.3, 52.7] & 0/8 [0.0, 36.9]\\
CODE P2 +13B & 1/8 [0.3, 52.7] & 0/8 [0.0, 36.9]\\
CODE P2 +26B & 0/8 [0.0, 36.9] & 0/8 [0.0, 36.9]\\
\bottomrule
\end{tabular}
\end{table}

\begin{table}[!htbp]
\centering\small
\setlength{\tabcolsep}{3.2pt}
\caption{Random-run success rates, part 3. Entries give success count / seed count and marginal 95\% Clopper--Pearson intervals (\%). Gain means $\Delta>10$ points and high score means terminal accuracy 50\% or more. Each checkpoint is estimated separately.}
\label{tab:stats-rates-3}
\begin{tabular}{lrr}
\toprule
Checkpoint / task & Gain: count [95\% CI] & High score: count [95\% CI]\\
\midrule
CODE P2 +50B & 1/8 [0.3, 52.7] & 0/8 [0.0, 36.9]\\
Lookup P1 5B & 0/8 [0.0, 36.9] & 0/8 [0.0, 36.9]\\
Lookup P1 462B & 2/8 [3.2, 65.1] & 0/8 [0.0, 36.9]\\
Lookup P1 839B & 0/8 [0.0, 36.9] & 0/8 [0.0, 36.9]\\
Lookup P1 1,259B & 0/8 [0.0, 36.9] & 0/8 [0.0, 36.9]\\
Lookup P1 2,098B & 0/8 [0.0, 36.9] & 0/8 [0.0, 36.9]\\
Lookup P1 2,937B & 2/8 [3.2, 65.1] & 0/8 [0.0, 36.9]\\
Lookup P1 3,532B & 1/8 [0.3, 52.7] & 1/8 [0.3, 52.7]\\
Lookup P1 3,896B & 2/8 [3.2, 65.1] & 0/8 [0.0, 36.9]\\
Lookup P2 +5B & 1/8 [0.3, 52.7] & 1/8 [0.3, 52.7]\\
Lookup P2 +13B & 2/8 [3.2, 65.1] & 4/8 [15.7, 84.3]\\
Lookup P2 +26B & 0/8 [0.0, 36.9] & 0/8 [0.0, 36.9]\\
Lookup P2 +50B & 2/8 [3.2, 65.1] & 2/8 [3.2, 65.1]\\
O1 & 0/8 [0.0, 36.9] & 0/8 [0.0, 36.9]\\
O2 & 0/8 [0.0, 36.9] & 0/8 [0.0, 36.9]\\
O3 & 1/8 [0.3, 52.7] & 0/8 [0.0, 36.9]\\
O4 & 4/8 [15.7, 84.3] & 4/8 [15.7, 84.3]\\
S1 & 0/8 [0.0, 36.9] & 0/8 [0.0, 36.9]\\
S2 & 0/8 [0.0, 36.9] & 0/8 [0.0, 36.9]\\
S3 & 0/8 [0.0, 36.9] & 0/8 [0.0, 36.9]\\
S4 & 0/8 [0.0, 36.9] & 0/8 [0.0, 36.9]\\
\bottomrule
\end{tabular}
\end{table}

\end{document}